\documentclass[12pt]{article}
\usepackage[utf8]{inputenc}
\usepackage{authblk}
\usepackage[margin=1in]{geometry}
\usepackage[titletoc,title]{appendix}
\usepackage[dvipsnames,table,xcdraw]{xcolor}
\usepackage{color}
\usepackage{setspace}
\usepackage{amsmath}
\usepackage{bbm}

\usepackage{amsmath,amsfonts,amssymb,amsthm,mathtools,bm,cases}
\usepackage{algorithm2e}
\SetKwComment{Comment}{/* }{ */}
\SetKwProg{Init}{Initialize}{}{}
\RestyleAlgo{ruled}

\usepackage{graphicx,float}
\usepackage{caption,subcaption}  
\usepackage{booktabs}
\usepackage{booktabs,multirow}

\usepackage[colorlinks=true, pdfborder={0 0 0}, linkcolor=red]{hyperref}

\usepackage[style=authoryear,date=year,maxcitenames=2,maxbibnames=20,giveninits,url=false,backend=biber,uniquelist=false,uniquename=init,dashed=false]{biblatex}

\bibliography{revision_jrssb}

\usepackage{wasysym}
\usepackage{lipsum}

\allowdisplaybreaks
\def\One{\mathbf{1}}

\def\A{\mathcal A}

\def\D{\mathcal D}

\def\F{\mathcal F}

\def\I{\mathcal I}

\def\L{\mathcal L}

\def\R{\mathcal R}

\def\BB{\mathbb B}
\def\EE{\mathbb E}

\def\NN{\mathbb N}
\def\PP{\mathbb P}
\def\RR{\mathbb R}
\def\ZZ{\mathbb Z}

\renewcommand{\hat}{\widehat}
\renewcommand{\tilde}{\widetilde}

\newtheorem{definition}{Definition}
\newtheorem{assumption}{Assumption}
\newtheorem{theorem}{Theorem}

\newtheorem{corollary}{Corollary}
\newtheorem{lemma}{Lemma}
\newtheorem{remark}{Remark}

\title{Stochastic Gradient Descent: A Nonlinear Time Series Perspective}
\author[1]{Jiaqi Li\thanks{Corresponding author. Email: jiaqi.li@rice.edu}}
\author[2]{Zhipeng Lou}
\author[3]{Stefan Richter}
\author[4]{Wei Biao Wu}

\affil[1]{Department of Statistics, Rice University, USA}
\affil[2]{Department of Mathematics, University of California, San Diego, USA}
\affil[3]{Institute of Applied Mathematics, Heidelberg University, Germany}
\affil[4]{Department of Statistics, University of Chicago, USA}

\date{}

\begin{document}

\maketitle

\begin{abstract}
\setstretch{1.0}\fontsize{9pt}{14.5pt}\selectfont
This paper investigates the statistical behavior of Stochastic Gradient Descent (SGD) with constant step size under the framework of iterated random functions. Compared to previous studies establishing the convergence of SGD in probability measure, e.g., Wasserstein distance, our approach directly provides the convergence in Euclidean distance by showing the Geometric Moment Contraction (GMC) of SGD. This new convergence can address the non-stationarity of SGD due to fixed initial points and can provide a more refined asymptotic analysis of SGD. Specifically, we prove quenched central limit theorem, invariance principle and $\alpha$-stable convergence for averaged SGD (ASGD) regardless of initialization. Furthermore, we provide a novel perspective to understand the impact of step sizes in SGD by studying its derivative with respect to the step size. The existence of stationary solutions for the derivative processes is shown, enabling the use of interpolation to recover the full-trajectory of SGD. Subsequently, we utilize multiple step sizes and show an enhanced Richardson-Romberg extrapolation with improved bias representation, which brings ASGD estimates closer to the global optimum. Finally, we propose a new online inference method and a bias-reduced variant for the extrapolated ASGD. Empirical confidence intervals are constructed and the coverage probabilities approach the nominal level as $n$ grows.

\vspace{0.5cm}

\noindent\textbf{\textit{Keywords:}} stochastic gradient descent, iterated random function, geometric moment contraction, derivative process, quenched central limit theorem, online inference
\end{abstract}


\section{Introduction}

We consider the optimization problem posed by a strongly convex objective function $G:\, \RR^d\mapsto\RR$, defined as below,
\begin{equation}
    \label{eq_opt_goal}
    \theta^* = \arg\min_{\theta\in\RR^d}G(\theta), \quad \text{where } G(\theta) = \EE_{X\sim\Pi} [g(\theta,X)].
\end{equation}
In (\ref{eq_opt_goal}), $g(\theta,X)$ represents the noise-perturbed measurement of $G(\theta)$ and $X$ is a random variable sampled from the distribution $\Pi$. This optimization challenge has garnered tremendous attention across diverse fields, encompassing statistical learning, optimization and stochastic approximation among the others. 
In traditional M-estimation, the solution is obtained by minimizing the empirical risk function $\hat\theta=\arg\min_{\theta}\bar G_n(\theta)$, where $\bar G_n(\theta)=n^{-1}\sum_{i=1}^ng(\theta,X_i)$, which often comes with high computational costs, and is not feasible for large-scale problems or streaming data. To tackle these issues, various online methods have been proposed. The Robbins-Monro algorithm (\cite{robbins_stochastic_1951}), also known as Stochastic Gradient Descent (SGD), was among the first methods and has become the most widely used due to its simplicity and efficiency. 

Specifically, given a starting point $\theta_0\in\RR^d$, the $k$-th iteration of the SGD algorithm can be computed by the recursive procedure
\begin{equation}
    \label{eq_SGD}
    \theta_k(\gamma)=\theta_{k-1}(\gamma)-\gamma \nabla g(\theta_{k-1}(\gamma),X_k), \quad k = 1, 2, \ldots,
\end{equation}
where $X_1,X_2,\ldots$ are independent and identically distributed~(i.i.d.) observations sampled from the distribution $\Pi$, $\gamma > 0$ is a constant step size or learning rate, $\nabla g(\theta, X)$ is the stochastic gradient vector of the objective function $g(\theta,X)$ with respect to the first argument $\theta\in\RR^d$, and $\theta_0(\gamma)\equiv\theta_0$. We view $\theta_k(\gamma)$ as a random function of $\gamma$ and $\theta_0$ and it depends on $X_1, \ldots, X_k$. The primary goal of this paper is to provide a systematic asymptotic theory for $\{\theta_k(\gamma)\}_{k\in\NN}$ in the framework of nonlinear time series in terms of iterated random functions.

The convergence theory of SGD has been investigated for decades (\cite{kiefer_stochastic_1952,fabian_asymptotic_1968}). Additional references on stochastic approximation and iterated random functions are provided in Appendix~\ref{sec_additional_literature}. Despite the rich literature, the full distribution of iterates has been rarely studied in the optimization and machine learning communities. \textcite{bottou_optimization_2018} considered $\EE[G(\theta_k(\gamma))-G(\theta^*)]$ or $\EE|\theta_k(\gamma)-\theta^*|^2$, 
rather than for the full distribution of iterates. One approach to address this problem is by interpreting the SGD process $\{\theta_k(\gamma)\}_{k\in\NN}$ as a time-homogeneous Markov chain (\cite{pflug_stochastic_1986,bach_non-strongly-convex_2013,bach_bridging_2020,merad_convergence_2023}), which can be used to prove the convergence of $\{\theta_k(\gamma)\}_{k\in\NN}$ to the stationary distribution $\pi_{\gamma}$ in terms of probabilistic distances, such as the Wasserstein distance (\cite{bach_bridging_2020, merad_convergence_2023, gupta_convergence_2021}).
However, such type of convergence in a probabilistic measure is less effective in deriving refined convergence theory for the non-stationary SGD process $\{\theta_k(\gamma)\}_{k\in\NN}$ such as large-deviation inequalities and sharp invariance principles. Additionally, it can be nontrivial to generalize this type of Wasserstein convergence to the cases with heavy-tailed noises (\cite{krasulina_stochastic_1969, buraczewski_asymptotics_2012, cuny_martingale_2014, wang_convergence_2021}), since their applied tools require finite $p$-th central moments for $p\ge2$ (\cite{yu_analysis_2021,merad_convergence_2023}).

To bridge these gaps, this paper puts the SGD in a nonlinear time series framework, specifically by conceptualizing the SGD process as an iterated random function 
(\cite{diaconis_freedman_1999})
and establishing the geometric moment contraction (GMC) (\cite{wu_limit_2004}) for the SGD iterates. Different from the previous literature analyzing SGD-type online algorithms by iterated random functions 
(\cite{gupta_probabilistic_2024,gurbuzbalaban_heavy-tail_2021,hodgkinson_multiplicative_2021}),
our new insight based on GMC enables us to prove the convergence of SGD in the Euclidean norm by providing a direct $\L^p$-coupling between two SGD trajectories starting from different initial points. In particular, we show that for any two SGD sequences $\{\theta_k(\gamma)\}_{k\in\NN}$ and $\{\theta_k^{\dagger}(\gamma)\}_{k\in\NN}$ from recursion (\ref{eq_SGD}) with possibly different initial points $\theta_0(\gamma)=\theta_0,\theta_0^{\dagger}(\gamma)=\theta_0^{\dagger}\in\RR^d$, respectively, if $0<\gamma<\gamma(p)$, then
\begin{equation}
\label{eq_intro_gmc}
    \| \theta_{k}(\gamma) - \theta_{k}^{\dagger}(\gamma)\|_p := \big(\mathbb{E} |\theta_{k}(\gamma) - \theta_{k}^{\dagger}(\gamma)|^{p}\big)^{1/p} \leq \rho^{k} |\theta_{0} - \theta_{0}^{\dagger}|, \quad p > 1,
\end{equation}
where $\gamma(p) > 0$ and $\rho=\rho_{\gamma,p} \in(0, 1)$ are specified in Theorem~\ref{thm_GMC}, and $|\cdot|$ denotes the Euclidean norm on $\mathbb{R}^{d}$. Such convergence ensures the existence and uniqueness of the stationary distribution $\pi_{\gamma}$. In contrast to the SGD convergence in the commonly used Wasserstein-2 distance, GMC provides a convenient and easy-to-use framework for all $p$-th moments ($p>1$) on SGD solutions. More importantly, GMC can allow a systematic asymptotic and non-asymptotic theory, since it handles the non-stationarity of the SGD process due to fixed initializations. As such, this paper provides asymptotic results formulated for the complete observed SGD process starting from any arbitrary initial point. We also clarify the relation between our framework and SDE-based approaches to SGD. Stochastic modified equation methods (\cite{li_stochastic_2017,li_stochastic_2019}) and diffusion approximations (\cite{sirignano_stochastic_2020}) typically approximate SGD by a continuous-time diffusion in the small step-size regime as $\gamma\to0$. Our proposed framework is different as it works directly with discrete-time constant step-size SGD and investigates under which range of $\gamma$, the GMC condition holds, which further leads to the Brownian motion approximation with optimal rates in probability. In general, this perspective can shed new light on the SDE-type approaches by providing a route to derive approximation rates when studying the weak convergence of SGD iterates to SDE solutions.

Beyond moment convergence, it is also crucial to quantify the uncertainty of SGD estimators by providing confidence intervals (CI); see for example, \textcite{fang_online_2018,fang_scalable_2019,zhong_online_2023} employed bootstrap strategies to create confidence intervals; \textcite{liang_statistical_2019,su_higrad_2023} alternated iterations of the SGD algorithm, drawing inferences from a non-asymptotic perspective; \textcite{chao_generalization_2019} introduced a cutting-edge class of regularized dual averaging to quantify uncertainties for online sparse methodologies. In this paper, we propose an online inference method for the popular averaged SGD (ASGD), also referred to as Polyak-Ruppert averaging (\cite{ruppert_efficient_1988,polyak_acceleration_1992}). ASGD can effectively reduce noise impact and has an asymptotically normal limiting distribution. Specifically, given an initial point $\theta_0\in\RR^d$, define the $n$-th ASGD iterate by
\begin{equation}
    \label{eq_ASGD}
    \bar{\theta}_{n}(\gamma) = \frac{1}{n} \sum_{k = 1}^{n} \theta_{k}(\gamma), \quad n\ge1. 
\end{equation}
Under certain regularity conditions, $\{\bar\theta_n(\gamma)\}_{n\in\NN}$ converges at rate $O_{\PP}(n^{-1/2})$ to some point (\cite{gyorfi_averaged_1996,defossez_averaged_2015}). See also other variants of ASGD formulations in \textcite{moulines_non-asymptotic_2011,dieuleveut_nonparametric_2016,dieuleveut_harder_2017,jain_parallelizing_2018,lakshminarayanan_linear_2018}. However, since $\{\theta_k(\gamma)\}_{k\in\NN}$ is non-stationary due to the fixed initial point $\theta_0$, its asymptotics heavily rely on the convergence metric. For instance, to obtain $\L^p$-convergence of $\bar\theta_k(\gamma)$ from a Wasserstein-type convergence result for $p\ge1$, one needs that the $2p$-th moment of $\nabla g(\theta,X)$ exists (see Remark~\ref{rmk_moment}), which seems to be an unnecessarily strong condition. To close this gap, we establish the asymptotic stationarity of ASGD using the GMC property and provide a quenched central limit theorem (CLT). Roughly speaking, for an arbitrary initial point $\theta_0=v_0\in\RR^d$, we can achieve the following asymptotic normality 
\begin{equation}
    \label{eq_intro_clt}
    \big[\sqrt{n}\big(\bar\theta_n(\gamma) - \theta_{\infty}^*(\gamma)\big) {\, \mid \,} \theta_0=v_0\big] \Rightarrow N\big(0,\Sigma^{\circ}(\gamma)\big) \mbox{ as } 
    n\rightarrow\infty,
\end{equation}
where $\Rightarrow$ denotes weak convergence and $\Sigma^{\circ}(\gamma) = \sum_{k \in \mathbb{Z}} \mathrm{cov}(\theta_0^{\circ}(\gamma), \theta_k^{\circ}(\gamma))$ is the long-run covariance matrix
and $\theta_k^{\circ}(\gamma) \sim \pi_\gamma$ is the stationary sequence following recursion (\ref{eq_SGD}) with
\begin{equation}
    \label{eq_SGD_mean}
    \theta_{\infty}^*(\gamma)=\lim_{k\rightarrow\infty}\EE[\theta_k(\gamma)]
    = \EE[\theta_1^{\circ}(\gamma)] = \int y \pi_\gamma(d y).
\end{equation} 
This quenched CLT is particularly relevant and useful in practice because varying initial points are frequently employed in SGD, and a CLT formulated only for the stationary SGD sequence is not directly useful for statistical inference. We can further establish an invariance principle (also known as functional CLT) for the process convergence, that is,
\begin{equation}
    \label{eq_intro_invar}
    n^{-1/2}\Big\{\sum_{k=1}^{\lfloor nu\rfloor}\big(\theta_k(\gamma)-\theta_{\infty}^*(\gamma) \big), 0\le u\le 1\Big\} \Rightarrow \{\Sigma^{\circ}(\gamma)^{1/2}\BB(u), 0\le u\le 1\},
\end{equation}
as $n\rightarrow\infty$, where $\BB(u)$ is a standard $d$-dimensional Brownian motion. The weak convergence in (\ref{eq_intro_invar}) is also quenched, which means (\ref{eq_intro_invar}) is true for all initial points $\theta_0\in\RR^d$.

In this paper, we focus on constant step-size SGD which is a popular choice in practice (\cite{bach_non-strongly-convex_2013}) since it can benefit from faster convergence in over-parametrized situations (\cite{schmidt_fast_2013,needell_stochastic_2014, ma_power_2018, wu_how_2018,vaswani_fast_2019}). Note that the SGD process with a constant step size does not converge, but oscillates around the mean of the stationary distribution with an average magnitude of $\gamma^{1/2}$ (\cite{pflug_stochastic_1986}). We provide a range for $\gamma$ that ensures the moment contraction of SGD, offering valuable guidance for practical step-size selection. To gain a deeper insight into the variations in SGD as influenced by different step sizes, we use the GMC to show the existence of the first- and second-order derivative processes with respect to $\gamma$, i.e., $\{\partial_{\gamma} \theta_k(\gamma) \}_{k\in\NN}$ and $\{\partial^2_{\gamma}\theta_k(\gamma)\}_{k\in\NN}$, under some mild conditions. 
In fact, the parametric differentiation of gradient descent-type algorithms has been an important topic in the literature. For instance, \textcite{maclaurin_gradient-based_2015} considered differentiation through SGD dynamics for gradient-based hyperparameter optimization, while \textcite{pedregosa_hyperparameter_2016,franceschi_forward_2017} studied approximate, forward-mode, and reverse-mode hypergradient methods for iterative learning algorithms. More recently, \textcite{mehmood_automatic_2019} studied automatic differentiation of first-order methods in parametric optimization, and \textcite{ehrhardt_analyzing_2024} analyzed inexact hypergradients for bilevel learning. These works study differentiation of algorithmic outputs with respect to parameters, often for sensitivity analysis, bilevel optimization, or hyperparameter learning. Our use of derivative recursions is in a similar vein, while is distinct in the sense that we further study their stationary solutions and moment properties of constant step-size SGD. Notably, derivative processes are powerful tools in measuring the deviation of a non-stationary time series from stationarity. Inspired by this line of works such as \textcite{dahlhaus_rao_aos_2006,dahlhaus_bernoulli_2019} on locally stationary time series, we discuss the innovative usage of derivative processes. Based on the existence of derivative processes, we propose an interpolation method for SGD trajectories across a continuum of step sizes to help efficiently recover the entire orbit.

Before constructing confidence intervals for the ASGD, we shall note that the stationary ASGD with a constant step size $\gamma$ is biased. Therefore, we adopt a Richardson-Romberg extrapolation (\cite{stoer_intro_2002}), firstly introduced to SGD by \textcite{bach_bridging_2020} to reduce the bias. We provide an enhanced non-asymptotic bias expansion of $\theta^{*}_{\infty}(\gamma)$ in terms of $\gamma$, i.e., $\theta_{\infty}^*(\gamma)-\theta^*= \gamma \Delta_1+\gamma^2 \Delta_2 +\R_{\gamma}^{(3)}$, 
where $\Delta_1,\Delta_2$ are independent of $\gamma$, and the remaining term satisfies $|\R_{\gamma}^{(3)}| \leq C\gamma^{5/2}$ for some constant $C>0$ independent of $\gamma$. By this expansion, we can run three parallel SGD sequences $\theta_k(a_1\gamma)$, $\theta_k(a_2\gamma)$ and $\theta_k(a_3\gamma)$, $k\in\NN$, with three different step sizes $a_1\gamma$, $a_2\gamma$ and $a_3\gamma$, where the constants $a_1,a_2,a_3 > 0$ are explicitly provided. Then, we define an extrapolated ASGD estimator as $\hat{\theta}_n(\gamma)=b_1\bar\theta_n(a_1\gamma) + b_2\bar\theta_n(a_2\gamma) + b_3\bar\theta_n(a_3\gamma),$
with some weighting constants $b_1,b_2,b_3$. By choosing the constants $a_i, b_i$ ($i=1,2,3$) suitably, we can achieve a cancellation of the bias components $\gamma\Delta_1,\gamma^2\Delta_2$ (the detailed expressions are deferred to (\ref{eq_extrapo_b})). Subsequently, we propose a new online inference method for $\hat{\theta}_n(\gamma)$ by a blocking-based online estimator of the long-run covariance matrix $\Sigma(\gamma)$. The empirical CI is detailed in~(\ref{eq_CI}). Compared to other online inference methods (\cite{chen_statistical_2020,Zhu2023}), we show that our proposed online estimator can achieve better mean square error (MSE) rate in case of dependence (see Remark~\ref{remark_mse}). The empirical CI exhibits desirable coverage probabilities in numerical experiments.


\textit{Contributions.} 
This study introduces technical tools in nonlinear time series to the machine learning community by providing a new interpretation for the SGD process. Our primary contributions are three-fold: (i) conceptualize the evolution of SGD as an iterative random function and establish its convergence in the Euclidean norm with an explicit range of $\gamma$ (Section~\ref{sec_gmc}); (ii) provide the quenched limiting distributions for the (extrapolated) ASGD with arbitrary initial points, facilitating an online inference method and a debiased variant with optimal MSE among batch-means estimators (Sections~\ref{sec_asymp} \&~\ref{sec_inference}); (iii) reconstruct a continuous spectrum of SGD trajectories as function of the step size $\gamma$ by derivative processes and interpolation (Section~\ref{sec_derivative}).

\textit{Notation.} For a vector $v = (v_{1}, \ldots, v_{d})^{\top} \in\RR^{d}$ and $q > 0$, we denote $|v|_{q}=(\sum_{i = 1}^{d} |v_{i}|^{q})^{1/q}$ and $|v|=|v|_2$. For a matrix $A$, denote the Frobenius norm by $|A|_2$. Denote the identity matrix in $\RR^{d\times d}$ by $\bm{I}_d$ and the vector $(1,\ldots,1)^{\top}\in\RR^d$ by $\One_d$. For any $s > 0$ and a random vector $X$, we say $X\in\L^s$ if $\|X\|_{s} = (\mathbb{E} |X|_{2}^{s})^{1/s} < \infty$. For two positive number sequences $(a_n)$ and $(b_n)$, we say $a_n=O(b_n)$ or $a_n\lesssim b_n$ (resp. $a_n\asymp b_n$) if there exists $C>0$ such that $a_n/b_n\le C$ (resp. $1/C\le a_n/b_n\le C$) for all large $n$, and write $a_n=o(b_n)$ or $a_n\ll b_n$ if $a_n/b_n\rightarrow0$ as $n\rightarrow\infty$. Let $(X_n)$ and $(Y_n)$ be two sequences of random variables. Write $X_n=o_{\PP}(Y_n)$ if $X_n/Y_n\rightarrow 0$ in probability as $n\rightarrow\infty$. Let $\langle \cdot, \cdot \rangle$ denote the canonical inner product in the finite-dimensional Euclidean space $\RR^d$.

\section{Geometric Moment Contraction}\label{sec_gmc}
The SGD sequence $\{\theta_{k}(\gamma)\}_{k \in \mathbb{N}}$ generated from~\eqref{eq_SGD} is typically non-stationary since the initial point $\theta_0\in\RR^d$ is arbitrarily fixed. We are interested in the existence of stationary distribution and the related convergence rate of $\theta_{k}(\gamma)$ towards this stationary distribution. 
However, the recursive form in~\eqref{eq_SGD} imposes a complicated dependency structure on $\{\theta_{k}(\gamma)\}_{k\in \mathbb{N}}$, making it nontrivial to establish a systematic asymptotic theory. To address the dependency, we adapt the coupling idea from nonlinear time series to the SGD algorithm. In particular, we first introduce the definition of geometric-moment contraction (\cite{wu_limit_2004}) as follows.

\begin{definition}[Geometric-moment contraction]\label{def_gmc}
Let $X_i, X_j'$, $i, j \in\ZZ$, be i.i.d.~random variables. Consider the causal process $Y_k = H_k(X_k, \ldots,X_1,X_0,X_{-1},\ldots)$, 
where $H_k(\cdot)$ is a measurable function so that $Y_k$ is well-defined and $\EE |Y_k|^p < \infty$, $p \ge 1$. We say that $Y_k$ is \textit{geometric-moment contracting (GMC)} if there exist some constants $C>0$ and $0<r<1$ such that for all $m\in\NN$, 
\begin{equation}
    \label{eq_GMC}
    \sup_k\EE|Y_k - Y_{k,\{m\}}|^p \le Cr^m, \mbox{ where } Y_{k,\{m\}} = H_k(X_k,\ldots,X_{k-m+1},X_{k-m}',X_{k-m-1}',\ldots)
\end{equation}
is a coupled version of $ Y_k $ with $X_j, j \le k-m$, in the latter replaced by i.i.d. random variables $X'_j, j \le k-m$, which are also independent of $X_i$, $i\in\ZZ$.
\end{definition}

Since $(X'_{k-m},X'_{k-m-1},\ldots)$ is independent of $(X_k,\ldots,X_{k-m+1})$ and has the same distribution as $(X_{k-m},X_{k-m-1},\ldots)$,  the coupled version $Y_{k,\{m\}}$ satisfies $\EE[Y_{k,\{m\}} \mid X_k,\ldots,X_{k-m+1}] = \EE[Y_k\mid X_k,\ldots,X_{k-m+1}]$. 
In this section, we shall show that under suitable conditions, GMC as defined above holds for $Y_k=\theta_k(\gamma)$ with $\theta_k(\gamma)$ generated from the SGD recursion~(\ref{eq_SGD}). Following the framework of iterated random functions (\cite{diaconis_freedman_1999}), we rewrite the SGD recursion~\eqref{eq_SGD} as follows. Define a measurable function $F:\RR^d\times\RR\to\RR^d$ by
\begin{equation}
    \label{eq_F_recursive}
    (\theta,X) \mapsto F_X(\theta) = \theta - \gamma \nabla g(\theta,X).
\end{equation}
Let $\theta_0,\theta_0^{\dagger}\in\RR^d$ be some given initial points independent of $X_{1}, \ldots, X_{n}$. Then, the SGD iterates generated from~\eqref{eq_SGD} starting at $\theta_0$ and $\theta_0^{\dagger}$  can be respectively written into $\theta_k(\gamma) = F_{X_k}(\theta_{k-1}(\gamma)) = F_{X_k} \circ \cdots \circ F_{X_1}(\theta_0)$ and $\theta_k^{\dagger}(\gamma) = F_{X_k} \circ \cdots \circ F_{X_1}(\theta_0^{\dagger})$.
Intuitively, if the GMC property holds for SGD, then the initial conditions will be quickly forgotten at an exponential convergence rate, and thus the trajectories of the two SGD sequences $\{\theta_k(\gamma)\}_{k\in\NN}$ and $\{\theta_k^{\dagger}(\gamma)\}_{k\in\NN}$ 
will be exponentially close to each other as $k$ grows to infinity, regardless of their different initializations. As such, this property will enable us to establish the weak convergence of $\theta_k(\gamma)$ to a unique stationary distribution $\pi_{\gamma}\in\L^p(\RR^d)$, that is, $\theta_k(\gamma)\Rightarrow\pi_{\gamma}$ as $k\rightarrow\infty$ (see Theorem~\ref{thm_GMC}). Unlike the Wasserstein distance used in previous literature (\cite{bach_bridging_2020,merad_convergence_2023,gupta_probabilistic_2024}), the geometric moment contraction directly controls the effect of initialization through the stationary coupling $\|\theta_k(\gamma)-\theta_k^\circ(\gamma)\|_p\leq \rho^k\|\theta_0-\theta_0^\circ(\gamma)\|_p$. Consequently, the distribution of $\theta_k(\gamma)$ converges to the stationary distribution $\pi_\gamma$ in the $p$-Wasserstein distance as well.
This property substantially facilitates the construction of an asymptotic theory for SGD iterates and it offers a natural base from which limiting distributions and concentration inequalities related to $\theta_k(\gamma)$ can be derived. In addition, it allows the heavy-tailed cases where the stochastic gradients may have infinite variances.

\begin{figure}[htbp!]
     \centering
     \begin{subfigure}{0.48\textwidth}
         \centering
         \includegraphics[width=\textwidth]{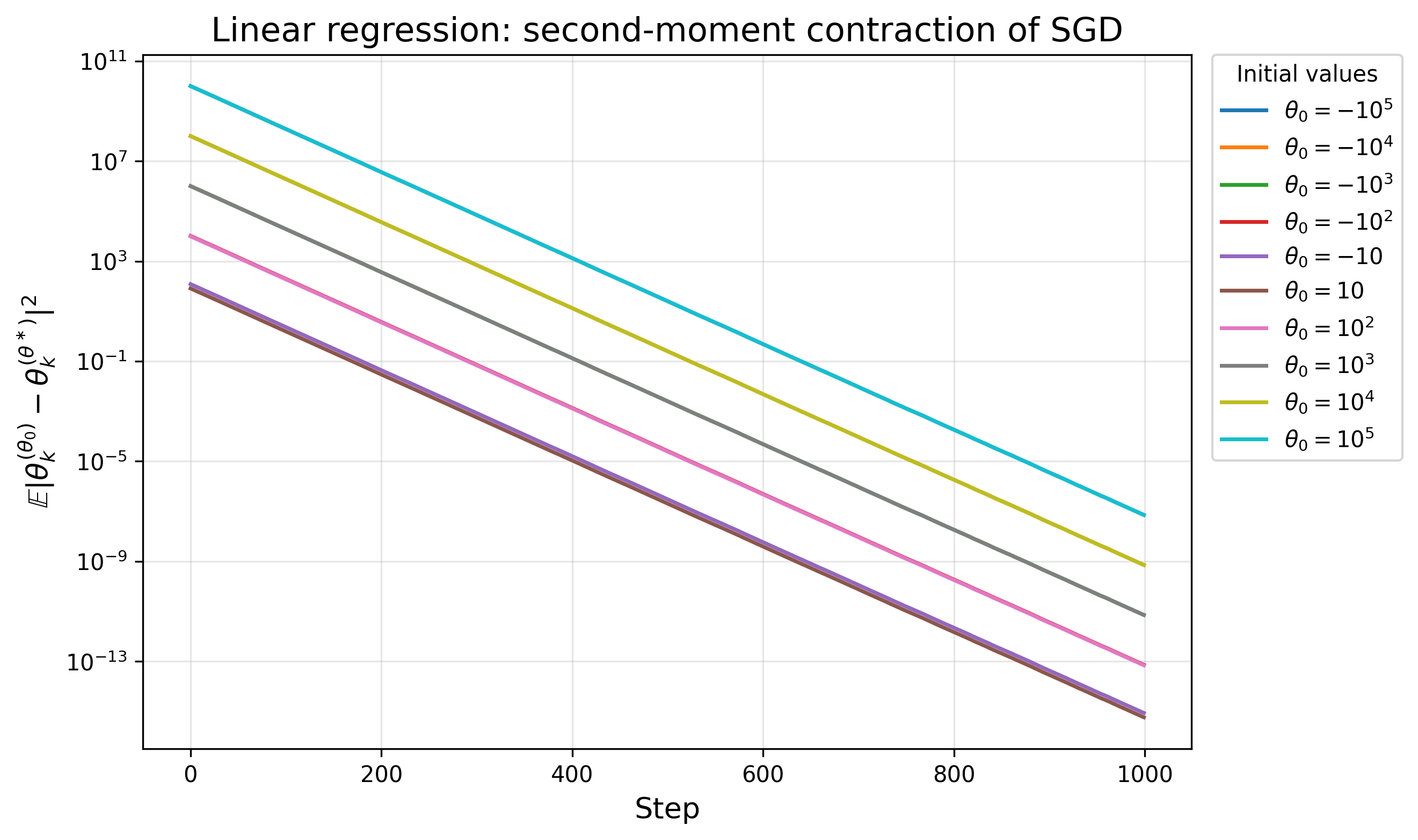}
         \caption{Linear regression.}
     \end{subfigure}
     \hfill
     \begin{subfigure}{0.5\textwidth}
         \centering
         \includegraphics[width=\textwidth]{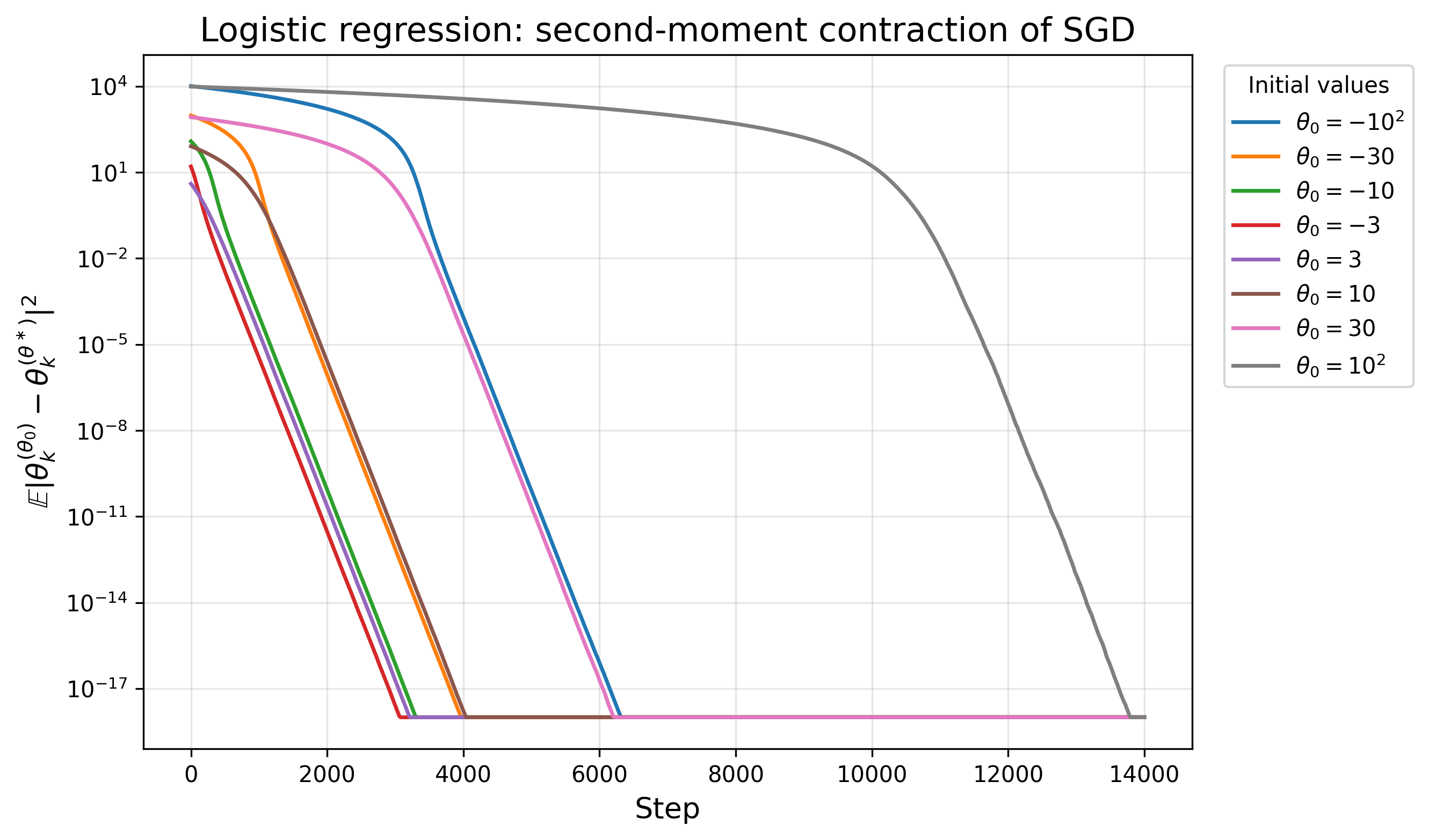}
         \caption{Logistic regression.}
     \end{subfigure}
    \caption{A simulation example illustrating GMC of SGD from initial values over multiple orders of magnitude. The optimum is $\theta^\ast=1$, the constant step size is $\gamma=0.02$, and we report $\EE|\theta_k^{(\theta_0)}-\theta_k^{(\theta^\ast)}|^2$ on a logarithmic scale. The initializations vary from $\pm 10$ to $\pm 10^5$ for linear regression and from $\pm 3$ to $\pm 10^2$ for logistic regression.}
        \label{fig_trace}
\end{figure}

\begin{remark}[Exponential convergence of SGD iterates]\label{rmk_sgd_converge}
To motivate the geometric-moment contraction for the SGD iterates, we present a simulation example to show that for varied initial points $\theta_0\in\RR$, the SGD iterates $\{\theta_k(\gamma)\}_{k\in\NN}$ forget the initial point exponentially fast for well-conditioned problems such as strongly convex objectives. In Figure \ref{fig_trace}, we show the convergence trace of SGD estimates for the least-square loss of the linear regression and the negative log-likelihood loss of the penalized logistic regression. We defer the details of the synthetic data to Section \ref{sec_numerical}. As shown in Figure \ref{fig_trace}, both models demonstrate that SGD trajectories initialized at different $\theta_0$ approach the reference trajectory initialized at $\theta^*$ at an exponential rate. This motivates us to directly study the contraction between SGD sequences with different initial values, beyond distributional metrics such as the Wasserstein distance, and to establish geometric moment contraction in Euclidean distance.
\end{remark}

Before stating the formal result of GMC, we first introduce some conditions on the convexity and smoothness of the loss functions and stochastic gradients.
\begin{assumption}[$\mu$-strong convexity]
\label{asm_conv}
Assume that for any $x$, $\theta \mapsto g(\theta,x)$ is a continuously differentiable function. Let $m(\theta)=\EE[\nabla g(\theta,X)]$. Suppose that there exists a constant $\mu>0$ such that for all $\theta,\theta' \in \mathbb{R}^d$, $\langle m(\theta)-m(\theta'), \theta-\theta' \rangle \ge \mu|\theta-\theta'|^2$.
\end{assumption}

\begin{assumption}[Stochastic Lipschitz continuity]
    \label{asm_Lip} 
    Let $p>1$. (i) For $\theta^*$ defined in (\ref{eq_opt_goal}), assume that $\|\nabla g(\theta^*,X)\|_p<\infty$; 
    (ii) Assume that there exists some constant $L_p>0$ such that for all $\theta,\,\theta'\in\RR^d$, $\|\nabla g(\theta, X) - \nabla g(\theta', X)\|_{p} \leq L_p|\theta-\theta'|$.
\end{assumption}

Assumption~\ref{asm_conv} on strong convexity is commonly seen in literature (\cite{bach_bridging_2020,Huo2023_linear_stochastic_approx}). Assumption \ref{asm_Lip}(i) is a mild condition on the $p$-th moment of the stochastic gradient. Assumption~\ref{asm_Lip}(ii) requires stochastic Lipschitz continuity of the gradient, which also indicates that the function $m(\theta)$ is $L_1$-smooth, by noting that $\|\cdot\|_p\ge\|\cdot\|_1=\EE|\cdot|$. Similar assumptions have been adopted in the literature on SGD asymptotics; see for example \textcite{chen_statistical_2020} on parameter estimation and inference with the SGD. It is noteworthy that previous studies usually impose $L$-cocoercive assumptions which only account for the cases with $p\ge2$ and require $p$ to be an integer  (\cite{bach_bridging_2020,merad_convergence_2023}). However, our Assumption \ref{asm_Lip} only requires $p>1$ and $p$ can be fractional. This is a nontrivial extension that facilitates general $\L^p$ convergence and also allows for infinite variances.

Next, we show that for sufficiently small $\gamma>0$, we can obtain the weak convergence $\theta_k(\gamma) \Rightarrow \pi_\gamma$, as $k \to \infty$, where $\pi_\gamma$ is the stationary distribution. We define the boundary $\gamma(p) > 0$ as the root to the equation
\begin{numcases}{}
(1 + \gamma L_p)^p
    = 1 + p \gamma L_p + p \mu \gamma
    & if $p \ge 2$
    \label{eq_gammap_gtr2}\\
\gamma(p)
    = \big(p \mu/(2^{2-p} L_p^p)\big)^{1/(p-1)}
    & if $1 < p < 2$.
    \label{eq_gammap}
\end{numcases}
When $p \ge 2$, since the function $h_p(u) = ((1 + u)^p - 1 - p u)/ u = p \int_0^1 ((1+u t)^{p-1} - 1) d t$ is strictly increasing in $u \in (0, \infty)$, $\lim_{u\downarrow0}h_p(u) = 0$, and $\lim_{u\uparrow\infty}h_p(u) = \infty$, there is a unique root $u = u(p)$ to the equation $h_p(u) = p \mu / L_p$, which implies that the root $\gamma(p)$ to equation (\ref{eq_gammap_gtr2}) is $u(p) / L_p$. A particularly interesting special case is $p = 2$. In this case both (\ref{eq_gammap_gtr2}) and (\ref{eq_gammap}) with $p \uparrow 2$ yield $\gamma(2) = 2 \mu / L_2^2$, suggesting a smooth transition at the borderline $p = 2$, and a central limit theorem is given in Theorem \ref{thm2_quenched_CLT}. 
Although the general GMC framework for iterated random functions was developed in \textcite{wu_limit_2004}, verifying this property for constant step-size SGD is far from automatic. In particular, one needs to control the random update map $F_X(\theta)=\theta-\gamma\nabla g(\theta,X)$ in $\L^p$ norm and to identify an explicit range of step sizes $\gamma$ under which it is contractive. This is especially delicate when $1<p<2$, since the stochastic gradients may have infinite variance and the usual second-moment arguments are no longer available. To address this difficulty, Lemma~\ref{lemma_power_ineq} establishes $\L^p$ moment inequalities that allow us to prove contraction for all $p>1$. These inequalities are a key technical ingredient of Theorem~\ref{thm_GMC} and may also be useful for analyzing other online learning recursions with heavy-tailed stochastic updates.

\begin{theorem}[Geometric moment contraction]
\label{thm_GMC}
Let $p > 1$. Recall (\ref{eq_gammap_gtr2}) and (\ref{eq_gammap}) for $\gamma(p)$. Suppose that Assumptions~\ref{asm_conv} and \ref{asm_Lip} hold and $0 < \gamma < \gamma(p)$. Then, for all $\theta, \theta'\in\RR^d$, 
\begin{equation}
    \label{eq_contraction}
    \|F_X(\theta) - F_X(\theta')\|_{p} \leq \rho |\theta - \theta'|,
\end{equation}
where $F_X(\theta)$ is defined in~\eqref{eq_F_recursive},
\begin{align}
\label{eq_rho}
    \rho=\rho_{\gamma,p} \quad \text{and}\quad \rho_{\gamma,p}^{p} = \left\{
    \begin{array}{cc}
    1 - p \mu \gamma + \gamma^p 2^{2-p} L_p^p < 1, & \quad \mathrm{if} \enspace 1 < p < 2;\\
    (1 + \gamma L_p)^p  - p \gamma L_p - p \mu \gamma < 1, & \quad \mathrm{if} \enspace p \geq 2. 
    \end{array}
    \right.
\end{align}
Consequently, there exists a unique stationary distribution $\pi_{\gamma}$ such that for all $\theta_0\in\RR^d$, $\theta_k(\gamma) \Rightarrow \pi_{\gamma}$ as $k\rightarrow\infty$, and the distribution $\pi_{\gamma}$ has a finite $p$-th moment $\int | y |^p \pi_{\gamma}(d y) < \infty$. Additionally, let $\theta_0^{\circ}(\gamma) \sim \pi_{\gamma}$ follow the stationary distribution and define $\theta_k^{\circ}(\gamma)$, $k \ge 1$, according to the iteration (\ref{eq_SGD}). Then the following geometric moment contraction holds:
\begin{equation}\label{21}
    \| \theta_k(\gamma) - \theta_k^{\circ}(\gamma)\|_p \le \rho^k \| \theta_0 - \theta_0^{\circ}(\gamma) \|_p
    = \rho^k \left[ \int |\theta_0 - y |^p \pi_{\gamma}(d y) \right ]^{1/p}.
\end{equation}
Moreover, let $p\ge2$, we have $\|\theta_k^{\circ}(\gamma)-\theta^*\|_p^{2} = O(\gamma)$.


\end{theorem}

Theorem \ref{thm_GMC} provides an explicit sufficient condition for the existence of the $p$-th moment of the stationary distribution based on the step size $\gamma$. In general, the existence of higher-order moments asks for smaller values of $\gamma$. Standard analysis of SGD shows linear (geometric) decay of the dependence on initialization in the expected distance to optimality such as $\EE[G(\theta_k)-G(\theta^*)]$ (see, for example, Theorem 4.6 of \textcite{bottou_optimization_2018}). The geometric contraction in~\eqref{21} additionally provides an explicit coupling bound between $\theta_k(\gamma)$ and $\theta_k^{\circ}(\gamma)$. This bound quantifies the effect of initialization through the distance between coupled iterates, with a linear (geometric) decay rate.
Besides, the GMC property in Theorem~\ref{thm_GMC} holds for any $p>1$ as long as Assumption~\ref{asm_Lip} is satisfied for this $p$. This allows infinite variances for $\theta_k(\gamma)$ (see Remark~\ref{rmk_moment}) and is fundamental to the derivation of non-Gaussian limiting distributions for the heavy-tailed settings in   Appendix~\ref{subsec_heavy_tail}.

In general, the strong convexity in Assumption~\ref{asm_conv} cannot be relaxed for the moment contraction (\ref{eq_contraction}). Assume otherwise there exists a pair $\theta \neq \theta'$ such that $K := \langle \theta-\theta',m(\theta) - m(\theta')\rangle/|\theta-\theta'|^2 \le 0$. 
For this particular pair $\theta \neq \theta'$, letting $p=2$, we have
\begin{align*}
    { {\EE|F_X(\theta) - F_X(\theta')|^2} \over |\theta-\theta'|^2}  
    = 1 - 2 \gamma K + \gamma^2 { {\|\nabla g(\theta, X) - \nabla g(\theta', X)\|^2 \over |\theta-\theta'|^2} },
\end{align*}
which is strictly larger than $1$ for all $\gamma > 0$, violating (\ref{eq_contraction}). 

Note that to establish the convergence of the SGD to the stationary distribution $\pi_{\gamma}$ in terms of the Wasserstein-2 distance, it is usually assumed that the gradient $\nabla g(\theta,X)$ is almost surely $L$-co-coercive (see for example Assumption A4($p$) in \textcite{bach_bridging_2020}). However, this assumption is stronger than our Assumption \ref{asm_Lip} since it poses constraints beyond Lipschitz continuity directly on $\nabla g$. For example, it requires the observed data to be a.s. bounded or with bounded kurtosis. We shall consider the following linear model as an example to show that our conditions are less restrictive, equivalent to a relaxed $L$-co-coercive condition such as Assumption A7($p$) in \textcite{bach_bridging_2020}.

\begin{remark}[Range of $\gamma$ in linear regression]
    Let $X_{k} = (Z_{k}, Y_{k}) \in \mathbb{R}^{d} \times \mathbb{R}$, $k \geq 1$, be i.i.d.~random samples following the linear regression model $Y_{k} = Z_{k}^{\top}\theta^{*} + \epsilon_{k}$, where $\theta^{*} \in \mathbb{R}^{d}$ is the population parameter vector of interest and $\epsilon_{k} \in \mathbb{R}$, $k \geq 1$, are i.i.d.~random noise independent of $\{Z_{k}\}_{k \geq 1}$. Assume $\EE[\epsilon_k]=0$ and $\EE[\epsilon_k^2]>0$.
    Our Assumption~\ref{asm_conv} and Assumption~\ref{asm_Lip} hold with $\mu = \lambda_{\min}\{\mathbb{E}(Z_{1} Z_{1}^{\top})\}$ and $L_{2} = \sup_{\delta \in \mathbb{R}^{d} : |\delta| = 1} \|Z_{1} Z_{1}^{\top} \delta\|_{2}$, respectively, where $\lambda_{\min}\{\cdot\}$ denotes the smallest eigenvalue. Consequently, Theorem~\ref{thm_GMC} ensures the GMC of the sequence $\{\theta_{k}(\gamma)\}_{k \geq 1}$ as long as 
    \begin{align}
    \label{eq_range_gamma_linear_regression}
        0 < \gamma < \frac{2 \lambda_{\min}\{\mathbb{E}(Z_{1} Z_{1}^{\top})\}}{\sup_{\delta \in \mathbb{R}^{d} : |\delta| = 1} \|Z_{1} Z_{1}^{\top} \delta\|_{2}^{2}}. 
    \end{align}
    It is worth mentioning that the range above for $\gamma$ can not be improved in general for $p=2$. To see this, suppose now $d = 1$, $\theta^*=0$, and $\mathbb{E} (Z_{1}^{4}) = \varpi \mathbb{E} (Z_{1}^{2})$ for some $0 < \varpi < \infty$. Then~\eqref{eq_range_gamma_linear_regression} reduces to $0 < \gamma < 2/\varpi$. In particular, for any $\gamma \ge 2/\varpi$, the second moment of the stationary distribution $\pi_{\gamma}$ is no longer finite. Assume otherwise, the second moment of the asymptotic distribution $\pi_{\gamma}$ is finite for $\gamma\ge2/\varpi$, that is, $\varsigma^2:=\int y^2\pi_{\gamma}(d y)<\infty$. Then, in view of~\eqref{eq_SGD}, for any $\gamma \ge 2/\varpi$, we have $\varsigma^{2} = \varsigma^{2} \mathbb{E} (1 - 2 \gamma Z_{1}^{2} + \gamma^{2} Z_{1}^{4}) + \gamma^{2} \mathbb{E} (Z_{1}^{2}) \mathbb{E} (\epsilon_{1}^{2}) \ge \varsigma^{2} + \gamma^{2} \mathbb{E} (Z_{1}^{2}) \mathbb{E} (\epsilon_{1}^{2}) > \varsigma^{2}$,
    which leads to a contradiction. 
\end{remark}

\begin{remark}[Extension to heavy-tailed noises]
\label{rmk_moment}
Note that by providing the convergence of SGD in terms of the Wasserstein-2 distance, one requires at least the $2p$-th finite moment to bound the distance $\EE|\bar\theta_k(\gamma) - \theta_{\infty}^*(\gamma)|^p$, 
and the moment contraction can only be established for $p\ge2$ (\cite{bach_bridging_2020}). The reason is that the martingale decomposition is adopted in the previous literature, that is, $F_X(\theta) = \theta - \gamma m(\theta) + \gamma D(\theta, X)$, where $m(\theta)=\EE [\nabla g(\theta, X)]$ and $D(\theta,X) = m(\theta) - \nabla g(\theta, X)$. 
This decomposition may lead to a loose upper bound of $\|F_X(\theta)-F_X(\theta')\|_p$ when applying the Cauchy-Schwarz inequality twice. This type of condition on higher-order finite moments (at least fourth moment) can be too restrictive, especially if the input data have heavy tails or are influenced by outliers. Differently, we address this issue by directly providing the bound for $\|F_X(\theta)-F_X(\theta')\|_p$ without this martingale decomposition. As such, we only require the existence of $p$-th moment instead of $2p$ for the GMC property of the SGD, and the moment contraction holds for all $p>1$. This improvement can accommodate a wider variety of distributions of the observations, even those with an infinite variance. We provide the quenched $\alpha$-stable convergence for the ASGD sequence $\{\bar\theta_k(\gamma)\}_{k\ge1}$ with heavy-tailed noises in Appendix~\ref{subsec_heavy_tail}.
\end{remark}

The GMC property in Theorem \ref{thm_GMC} characterizes the dependencies of the SGD process $\{\theta_k(\gamma)\}_{k\in\NN}$ on the initial conditions $\theta_0\in\RR^d$, which decays exponentially fast as the step number $k\rightarrow\infty$. This theoretical guarantee can be especially useful in practical applications, since SGD is often implemented with a variety of starting points $\theta_0$ to enhance the robustness and effectiveness of the optimization process. Based on GMC, we shall further introduce the convergence for the ASGD sequence $\{\bar\theta_n(\gamma)\}_{n\in\NN}$ in Section \ref{sec_asymp}. These convergence results shall facilitate more refined limiting behavior of ASGD, such as the quenched central limit theorems, which provides a fundamental base for the statistical inference of ASGD estimates with any arbitrary initial conditions (see Section \ref{sec_inference}).

\section{Asymptotic Theory}
\label{sec_asymp}

In this section, we first introduce the asymptotic properties of the ASGD estimate $\bar\theta_n(\gamma)$ defined in (\ref{eq_ASGD}), for all $p>1$, by using the geometric-moment contraction provided in Theorem \ref{thm_GMC}. In particular, we give a bound on the moment $\|\bar\theta_n(\gamma) - \theta_{\infty}^*(\gamma)\|_p$ and provide an almost sure convergence result, for all $p>1$. Then, for $p\ge2$, we will provide the quenched central limit theorem for $\bar\theta_n(\gamma)$ and establish the quenched invariance principle with $p>2$.

\subsection{Quenched Central Limit Theorem}
\label{subsec_clt}

We start with the moment convergence and the almost sure convergence of ASGD $\bar\theta_n(\gamma)$.

\begin{theorem}[Convergence of the ASGD]
\label{thm_asgd_converge}
Let $p > 1$. Recall $\theta_{\infty}^*(\gamma)$ defined in (\ref{eq_SGD_mean}). Under the conditions of Theorem~\ref{thm_GMC}, we have the $p$-th order moment convergence 
$\|\bar{\theta}_{n}(\gamma) - \theta_{\infty}^{*}(\gamma)\|_{p} = O(n^{1/(p \wedge 2) - 1}),$
and the following almost sure convergence: 
\begin{itemize}
    \item [(i)] for $1<p<2$, $|\bar{\theta}_{n}(\gamma) - \theta_{\infty}^{*}(\gamma)| = o_{a.s.}(n^{1/p - 1})$;
    \item [(ii)] for $p\ge2$, $|\bar{\theta}_{n}(\gamma) - \theta_{\infty}^{*}(\gamma)| = O_{a.s.}\big((n^{-1} \log\log(n) )^{1/2}\big).$
\end{itemize} 

\end{theorem}

Next, we shall show the (quenched) central limit theorem ($p = 2$) for this Markov chain, that is, for any initial point $\theta_0\in \RR^d$, the central limit theorem holds for $\bar\theta_n(\gamma)$. Since the SGD sequence with an arbitrary starting point is a non-stationary Markov chain, we need to derive an approximation by a stationary process by following (\ref{21}) of Theorem \ref{thm_GMC}, which is the key to the proof of the quenched limit theorems. Theorem \ref{thm_invar_prin} concerns the refined Gaussian approximation of strong invariance principle under $p > 2$. The heavy-tailed case with $1<p<2$ is studied in Appendix~\ref{subsec_heavy_tail}.

\begin{theorem}[Quenched central limit theorem]
\label{thm2_quenched_CLT}
Suppose that Assumptions~\ref{asm_conv} and \ref{asm_Lip} hold with $p = 2$. Let $\gamma_1, \ldots, \gamma_\ell \in (0, \,\gamma(2))$, $\ell \geq 1$, be different step sizes and
\begin{equation}
\label{eq_clt_partial_sum}
 S_n^\circ = \sum_{k=1}^n \varepsilon^\circ_k, \mbox{ where } \varepsilon^\circ_k = \mathrm{vec} (\theta_k^{\circ}(\gamma_1) - \theta_\infty^*(\gamma_1), \ldots, \theta_k^{\circ}(\gamma_\ell) - \theta_\infty^*(\gamma_\ell)),
\end{equation}
with $\mathrm{vec}(v_1, \ldots, v_\ell)=(v_1^{\top},\ldots,v_\ell^{\top})^{\top}$. All $\ell$ chains use the same random samples $X_1,\ldots X_n$. Let $\Sigma^{\circ} = \sum_{k=-\infty}^\infty {\rm cov}(\varepsilon^\circ_0, \varepsilon^\circ_k)$ be the long-run covariance matrix of the stationary $(d \ell)$-dimensional process $(\varepsilon^\circ_k)_{k \in \mathbb{Z}}$. Assume $\lambda_{\min}(\Sigma^{\circ})\ge c_0$ for some positive constant $c_0<\infty$. Similarly we define $S_n$ as $S_n^\circ$ with $\theta_k^{\circ}(\gamma)$ in the latter replaced by $\theta_k(\gamma)$. Then $n^{-1/2} S_n^\circ \Rightarrow N(0, \Sigma^{\circ})$, and for an arbitrary initial point $\mathrm{vec}(\theta_0(\gamma_1),\ldots,\theta_0(\gamma_{\ell}))\in\RR^{d\ell}$, $n^{-1/2} S_n \Rightarrow N(0, \Sigma^{\circ})$. 
\end{theorem}

Based on the asymptotic normality results of the ASGD provided above, we can construct asymptotically valid confidence intervals for $\theta_\infty^*(\gamma)$. To facilitate the use of the result in practice, we introduce an online inference method in Section \ref{sec_inference} for the estimation of the long-run covariance matrix $\Sigma(\gamma)$ of the bias-reduced ASGD (see (\ref{eq_long_run_variance})). The long-run covariance matrix $\Sigma^{\circ}$ of the stationary SGD in Theorem \ref{thm2_quenched_CLT} can be similarly estimated. Two concrete examples for constructing confidence intervals are provided on a linear regression model and a logistic regression model; see Section \ref{sec_numerical} for more details.

\subsection{Quenched Invariance Principles}
\label{subsec_invar_prin}

The seminal work of Komlós, Major, and Tusnády (\cite{komlos_approximation_1975,komlos_approximation_1976}), presents an optimal Gaussian approximation (KMT embedding) for the partial sums of i.i.d.~random variables. Their findings have become indispensable tools in both probability and statistics; see for example, \textcite{berkes_liu_wu_2014}  extended this result to the dependent case; \textcite{karmakar_optimal_2020} generalized the KMT approximation to the multiple time series. Building on these previous works, in this section, we adapt the KMT approximation towards the recursive procedures, specifically by providing an invariance principle and its quenched version for the vector-valued ASGD processes $\{\bar\theta_k(\gamma)\}_{k\in\NN}$.

\begin{theorem}[Quenched invariance principle]
\label{thm_invar_prin}
Suppose that Assumptions~\ref{asm_conv} and \ref{asm_Lip} hold with some $p>2$. Recall Theorem \ref{thm2_quenched_CLT} for $S_n$ and $S_n^\circ$ with step sizes $\gamma_1, \ldots, \gamma_\ell \in (0, \, \gamma(p))$. All $\ell$ chains use the same random samples $X_1,\ldots X_n$. Then there exists a (richer) probability space $(\Omega_c,\A_c,\PP_c)$ on which we can define random vectors $\varepsilon_k^c \in\RR^{d\ell}$ with the partial sum process $S_i^c=\sum_{k=1}^i \varepsilon_k^c$, and a Gaussian process $G_i^c=\sum_{k=1}^i Z_k^c$, where $\{Z_k^c\}_{k \geq 1}$ are independent $N(0, \bm{I}_{d\ell})$ random vectors, such that 
$(S_i^c)_{1\le i\le n}\stackrel{\D}{=}(S_i^\circ)_{1\le i\le n}$ and
    \begin{equation}
        \label{eq_invar_prin}
        \max_{i\le n}\big|S_i^c - \Sigma^{\circ1/2} G_i^c\big|=o_{\PP}(n^{1/p}), \quad \text{in }(\Omega_c,\A_c,\PP_c),
    \end{equation} 
with $\Sigma^{\circ}$ defined in Theorem~\ref{thm2_quenched_CLT}. Additionally, the above approximation also holds for $S_n$ with an arbitrary initial point $\mathrm{vec}(\theta_0(\gamma_1),\ldots,\theta_0(\gamma_{\ell}))\in\RR^{d\ell}$.    
\end{theorem}

As a direct consequence of Theorem~\ref{thm_invar_prin}, the quenched functional CLT in~\eqref{eq_intro_invar} can be achieved. Such a result provides the asymptotic distribution of the entire path of the SGD estimates over iterations, not just their final or averaged values. Compared to the diffusion-based functional limit theorems that usually rescale the SGD sequence itself and prove weak convergence to a limiting diffusion process as $\gamma\rightarrow0$; see for example, \textcite{sirignano_stochastic_2020,flamand_functional_2026}, our invariance principle (functional central limit theorem) does not require $\gamma\rightarrow0$ and provides the optimal convergence rate $o_{\PP}(n^{1/p})$ in probability. The techniques for our optimal Brownian motion approximation can also be useful for deriving approximation rates of SGD by diffusion processes.

\section{Derivative SGD Processes}
\label{sec_derivative}
As shown in Theorem~\ref{thm_GMC}, smaller step sizes $\gamma$ result in slower convergence, while larger step sizes may cause instability. It is important to understand the behavior of SGD across a continuous range of step sizes. In this paper, we make use of the derivative processes of SGD with respect to $\gamma$ for interpolation across multiple step sizes. This strategy is particularly useful for practitioners since it avoids independently running SGD for each step size, enabling efficient reconstruction of the continuous spectrum of SGD orbits.

We begin by establishing the existence of stationary solutions for the first- and the second-order derivative processes, which forms the foundation for our interpolation of SGD orbits.

\subsection{Existence of the First Derivative Processes}

At the outset, we consider the first-order derivative process. Assume that $g(\theta, X)$ is second-differentiable with respect to the first argument. Let the starting function $\theta_0(\gamma)$ in (\ref{eq_SGD}) be a bounded differentiable function with bounded derivative (for example we can choose the constant function $\theta_0(\gamma) \equiv c$). Taking derivative of (\ref{eq_SGD}) with respect to $\gamma$, we obtain the recursion
\begin{align}
    \label{eq_der_midstep}
    \partial_{\gamma} \theta_k(\gamma) &= \bm{A}_\gamma(\theta_{k-1}(\gamma),X_k)
    \partial_{\gamma} \theta_{k - 1}(\gamma) - \nabla g(\theta_{k-1}(\gamma),X_k), \,\, k=1, 2, \ldots.
\end{align}
where $\nabla^2 g(\theta, X)$ is the Hessian matrix in $\RR^{d\times d}$ with respect to $\theta$ and 
\begin{equation}
    \label{eq_A_k}
    \bm{A}_\gamma(\theta,X) = \bm{I}_d - \gamma \nabla^2g(\theta,X).
\end{equation}
Consider the two-sided stationary SGD process $\{\theta_k^{\circ}(\gamma)\}_{k\in\ZZ}$ 
\begin{equation}
    \label{eq_der_stationary}
    \theta_k^{\circ}(\gamma) =\theta_{k-1}^{\circ}(\gamma) - \gamma \nabla g(\theta_{k-1}^{\circ}(\gamma),X_k), \,\,  k = \ldots, -1, 0, 1, \ldots
\end{equation}
and define the process $\{\dot{\theta}_k(\gamma)\}_{k\in\ZZ}$ by
\begin{equation}
    \label{eq_derivative}
    \dot{\theta}_k(\gamma) = \bm{A}_\gamma(\theta_{k-1}^{\circ}(\gamma),X_k)\dot{\theta}_{k-1}(\gamma)
    -\nabla g(\theta_{k-1}^{\circ}(\gamma),X_k), 
\end{equation}

A subtle point is that the stationary derivative process cannot be defined simply as
$\partial_{\gamma}\theta_k^{\circ}(\gamma)$ by differentiating the two-sided stationary
recursion~\eqref{eq_der_stationary}. Although the nonstationary iterates
$\theta_k(\gamma)$ are differentiable with respect to $\gamma$ under the above
smoothness conditions, the two-sided stationary solution
$\{\theta_k^\circ(\gamma)\}_{k\in\ZZ}$ is defined implicitly through its infinite past.
Therefore, differentiating~\eqref{eq_der_stationary} would require justifying the
differentiability, with respect to $\gamma$, of the whole stationary solution
$\theta_k^\circ(\gamma),\theta_{k-1}^\circ(\gamma),\ldots$, which is not immediate.

To circumvent this issue, we define the stationary derivative process directly through
the linear stochastic recursion~\eqref{eq_derivative}, which is the stationary analogue
of the nonstationary derivative recursion~\eqref{eq_der_midstep}. Theorem~\ref{Theorem_first_order_derivative}
below shows that~\eqref{eq_derivative} indeed admits a stationary solution
$\{\dot{\theta}_k(\gamma)\}_{k\in\ZZ}$ with finite moment, and that the derivative process
$\{\partial_\gamma\theta_k(\gamma)\}_{k\in\NN}$ computed from arbitrary initialization
converges to this stationary solution. In this sense,
$\{\dot{\theta}_k(\gamma)\}_{k\in\ZZ}$ is the stationary version of the derivative
process.


\begin{theorem}[Existence of stationary first derivative process]
\label{Theorem_first_order_derivative}
    Assume that $g(\theta,X)$ in~\eqref{eq_opt_goal} is a twice continuously differentiable function with respect to $\theta \in \mathbb{R}^{d}$. Let $p > 1$. Under conditions of Theorem~\ref{thm_GMC}, \eqref{eq_derivative} admits a stationary solution with finite $p$-th moment. Moreover, assume that there exists a positive constant $L_p^{\circ} < \infty$ such that for all $\theta, \theta' \in \mathbb{R}^{d}$, $\|\nabla^{2} g(\theta, X) - \nabla^2 g(\theta', X)\|_{p} \leq L_{p}^{\circ} |\theta - \theta'|.$ 
    Then, we have $\mathbb{E} |\partial_{\gamma} \theta_{k}(\gamma) - \dot{\theta}_{k}(\gamma)| = O(\rho_{\circ}^{k})$, where $\rho_{\circ} \in (0, 1)$. 
\end{theorem}
Theorem~\ref{Theorem_first_order_derivative} asserts that the stationary derivative process $\{\dot{\theta}_k(\gamma)\}_{k\in\ZZ}$ exists with $p$-th finite moment for some $p>1$, and the derivative process $\{\partial_\gamma\theta_k(\gamma)\}_{k\in\NN}$ computed from an arbitrary initialization converges to it. This provides the derivative-process counterpart of the asymptotic stationarity result for the original SGD iterates.

\subsection{Existence of the Second Derivative Processes}
In the bias expansion for the ASGD estimate in Theorem~\ref{thm_bias_expansion}, we will provide an explicit form for the coefficient of $\gamma^2$ term and bound the remaining term by $O(\gamma^{5/2})$ under some additional smoothness conditions in Assumption~\ref{asm_g_smooth}. Here, beyond the first derivative process with respect to $\gamma$, we would like to further establish the existence of a stationary solution for the second-order derivative process, which can provide a more refined bias expansion for the SGD estimates.

Recall the definition of $\bm{A}(\theta,X)$ in~\eqref{eq_A_k}. We take the derivative on the both sides of the first derivative process $\{\partial_{\gamma} \theta_{k}(\gamma)\}_{k \geq 1}$ in~\eqref{eq_der_midstep} and obtain
\begin{align*}
    \partial_{\gamma}^{2} \theta_{k}(\gamma) =& \bm{A}_{\gamma} (\theta_{k - 1}(\gamma), X_{k}) \partial_{\gamma}^{2} \theta_{k - 1}(\gamma) - 2 \nabla^{2} g(\theta_{k - 1}(\gamma), X_{k}) \partial_{\gamma} \theta_{k - 1}(\gamma) \cr 
    &-  \gamma \mathrm{vec}(\partial_{\gamma}^{\top} \theta_{k - 1}(\gamma) \nabla^{3} g(\theta_{k - 1}(\gamma), X_{k}) \partial_{\gamma} \theta_{k - 1}(\gamma)).
\end{align*}
Following~\eqref{eq_derivative}, we define the process $\{\ddot{\theta}_{k}(\gamma)\}_{k \in \mathbb{Z}}$ as follows,  
\begin{align}
\label{eq_derivative_second}
    \ddot{\theta}_{k}(\gamma) =& \bm{A}_{\gamma} (\theta_{k - 1}^{\circ}(\gamma), X_{k}) \ddot{\theta}_{k - 1}(\gamma) - 2 \nabla^{2} g(\theta_{k - 1}^{\circ} (\gamma), X_{k}) \dot{\theta}_{k - 1}(\gamma) \cr 
    &- \gamma \mathrm{vec}(\dot{\theta}_{k - 1}^{\top}(\gamma) \nabla^{3} g(\theta_{k - 1}^{\circ}(\gamma), X_{k}) \dot{\theta}_{k - 1}(\gamma)).
\end{align}

\begin{theorem}
\label{Theorem_Second_Derivative_Bound}
    Let $p > 2$. Assume that $\|\nabla^{3} g(\theta^{*}, X)\|_{p} < \infty$. Then, under the conditions of Theorem~\ref{Theorem_first_order_derivative}, \eqref{eq_derivative_second} admits a stationary solution with finite $(p/2)$th moment. Moreover, assume that there exists a positive constant $L_p^{\diamond} < \infty$ such that for all $\theta, \theta' \in \mathbb{R}^{d}$, $\mathbb{E} |\nabla^{3} g(\theta, X) - \nabla^{3} g(\theta', X)| \leq L_p^{\diamond} |\theta - \theta'|.$ 
    Then, we have $\mathbb{E} |\partial_{\gamma}^{2} \theta_{k}(\gamma) - \ddot{\theta}_{k}(\gamma)| = O(\rho_{\diamond}^{k})$, where $\rho_{\diamond} \in (0, 1)$. 
\end{theorem}

\begin{figure}[!htbp]
    \centering
    \begin{subfigure}{0.49\textwidth}
        \centering
        \includegraphics[width=1\linewidth]{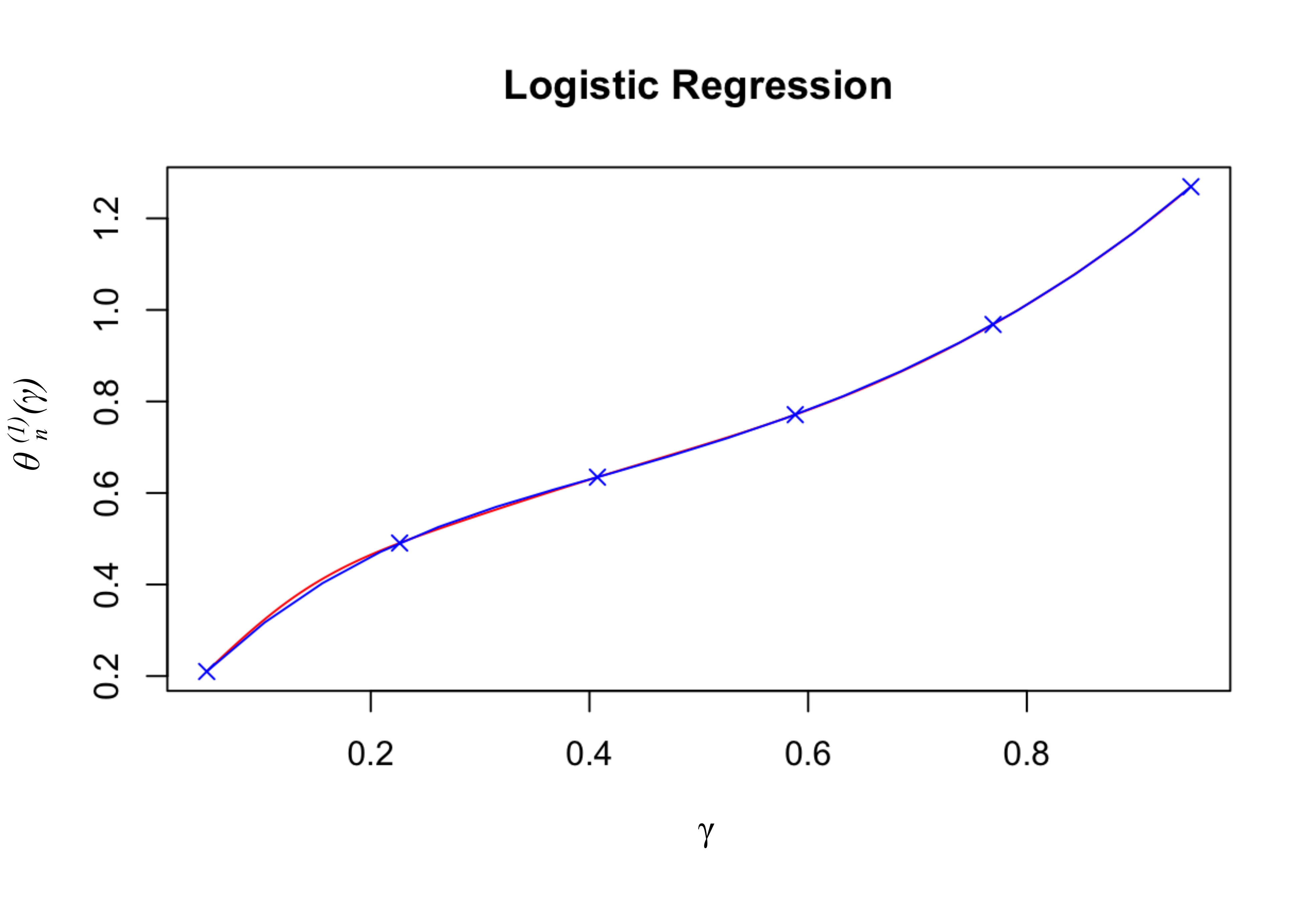}
        \caption{$\theta_n^{(1)}(\gamma)$.}
        \label{fig_logi_d1}
    \end{subfigure}   
    \begin{subfigure}{0.49\textwidth}
        \centering
        \includegraphics[width=1\linewidth]{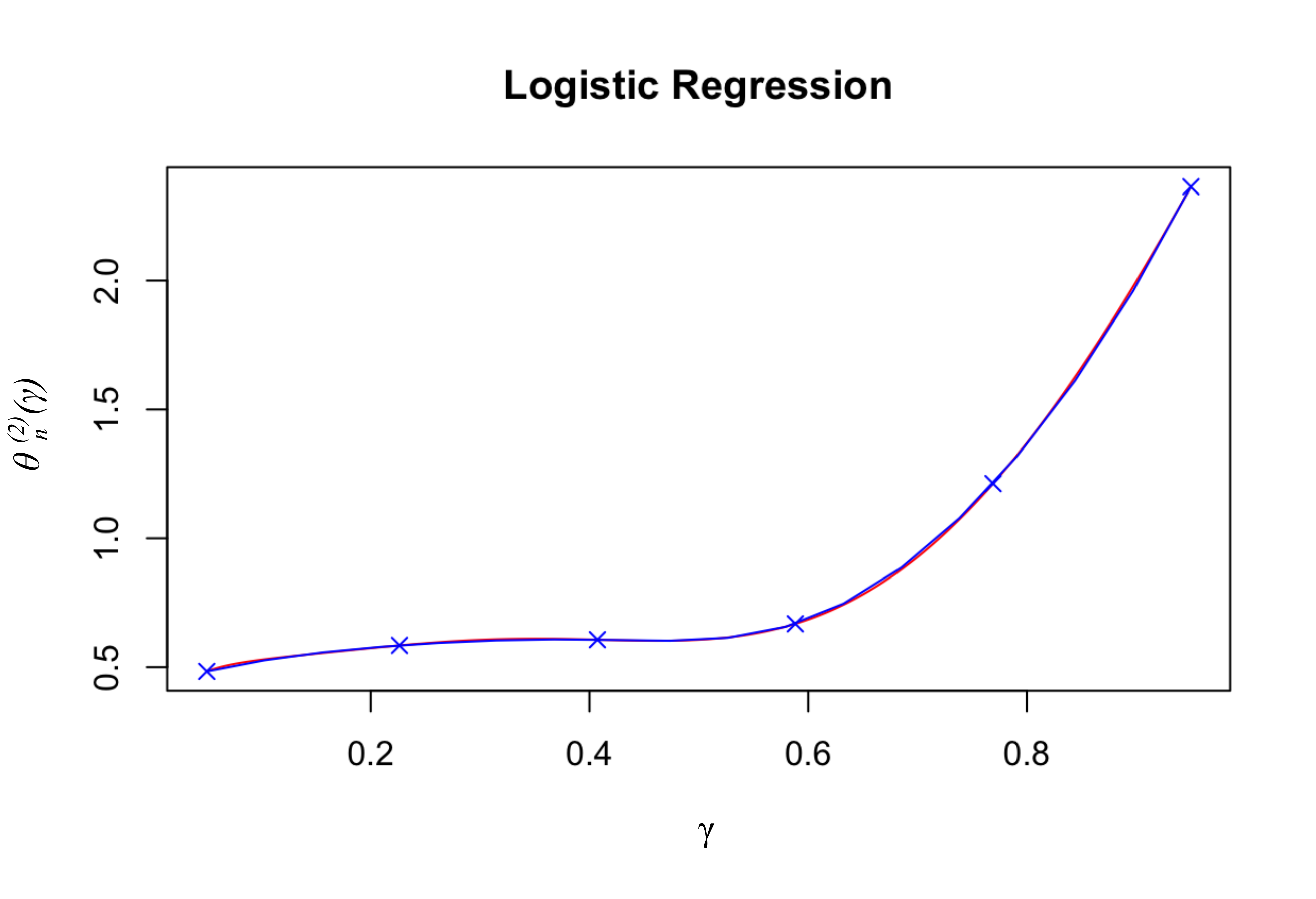}
        \caption{$\theta_n^{(2)}(\gamma)$.}
        \label{fig_orbit_lig_d2}
    \end{subfigure}   
    \caption{Spline interpolation on the last iterate of SGD, denoted by $\theta_n(\gamma)=(\theta_n^{(1)}(\gamma),\theta_n^{(2)}(\gamma))^{\top}$, for the logistic regression ($d=2$) with 6 internal knots shown as crosses. Fitted line by interpolation is shown in blue and the line of SGD estimates is shown in red.}
    \label{fig_orbit_logi_two_dim}
\end{figure}

The existence of derivative processes provides theoretical justification to reconstruct a continuous spectrum of SGD trajectories as a function of $\gamma$; say $\gamma \in [a, b]$, by running the SGD on a discrete set of step sizes, say $a =: \gamma_1 < \ldots < \gamma_{N-1} < \gamma_N := b$, with the help of a spline interpolation procedure. The latter approach avoids executing the SGD algorithm with each step size $\gamma$ which is computationally infeasible. Here, we consider the logistic regression model as an example. We generate i.i.d. $\{(\bm{z}_k,y_k)\}_{k\in\NN}$ with $\bm{z}_k\in\RR^2$ from the standard $N(0,\bm{I}_2)$ and $y_k=\mathrm{Bernoulli}\{1/[1+\exp(-\bm{z}_k^{\top}\theta^*)]\}$. The loss function is the negative log-likelihood defined by $g(\theta,(\bm{z},y))=(1-y)\bm{z}^{\top}\theta + \log(1+\exp(-\bm{z}^{\top}\theta))$. In Figure~\ref{fig_orbit_logi_two_dim}, we use the built-in R function {\tt ‘spline()’} and show the spline interpolation \textit{(blue)} on the last iterate of SGD \textit{(red)} with $6$ internal knots shown as crosses. It agrees very well with the red curves which show $\theta_n(\gamma)=(\theta_n^{(1)}(\gamma),\theta_n^{(2)}(\gamma))^{\top}$ for $\gamma$ ranging from $0.05$ to $0.95$. 

In the experiment for Figure~\ref{fig_orbit_logi_two_dim}, given the same $\{X_k\}_{k\ge1}$, the SGD iterates at different spline knots are computed. Moreover, we have regenerated the data stream $\{X_k\}_{k\ge1}$ using different random seeds and rerun SGD accordingly for multiple times, the similar realized paths and splines were observed, and the agreement between the interpolated and the original sequences is quite stable.

\section{Online Inference}\label{sec_inference}
In this section, we propose an online statistical inference method based on the SGD algorithm. To this end, we shall first provide an explicit non-asymptotic bias expansion for the ASGD estimates in terms of the step size $\gamma$ (see Theorem \ref{thm_bias_expansion}) and introduce a Richardson-Romberg extrapolated ASGD estimator utilizing three different step sizes, which effectively cancels out the bias term on the order of $\gamma$ and $\gamma^2$.

\subsection{Richardson-Romberg Extrapolation}

In the context of the SGD, the iterates can exhibit a bias due to the constant step size when approaching the optimal solution. This bias arises because the noisy gradient estimates do not average out entirely, even if the iterations are performed infinitely often. To resolve this issue, \textcite{bach_bridging_2020} first time adopted the Richardson-Romberg extrapolation, which is a technique primarily used in numerical integration to improve the accuracy of a solution by exploiting the knowledge of its error's rate of convergence. For example, \textcite{Huo2023_linear_stochastic_approx} provides a general bias expansion of Richardson-Romberg extrapolated estimators under mixing  conditions. In this section, we establish a bias expansion to cancel out both $\gamma$ and $\gamma^2$ terms in the extrapolated ASGD estimator expansion.

First, we provide a theoretical foundation, specifically by giving a non-asymptotic expansion of the bias between the ASGD solution in (\ref{eq_ASGD}) and the global minimum $\theta^*$, that is,  $\EE[\bar\theta_k(\gamma)-\theta^*]$. This bias can be further decomposed into two parts, that is, $\EE[\bar\theta_k(\gamma)-\theta^*] = \EE[\bar\theta_k(\gamma)-\theta_k^{\circ}(\gamma)] + \EE[\theta_k^{\circ}(\gamma)-\theta^*].$
Here $\theta_k^{\circ}(\gamma)$ has the stationary distribution $\pi_{\gamma}$ for all $k\in\NN$. The first part $\EE[\bar\theta_k(\gamma)-\theta_k^{\circ}(\gamma)]$ converges to zero at a rate of $O(1/k)$ as $k\rightarrow\infty$ by applying the GMC property shown in Theorem~\ref{thm_GMC}. Moreover, we provide an expansion of the second part $\EE[\theta_k^{\circ}(\gamma)-\theta^*]$ for small step sizes $\gamma > 0$, with an error of the order of $O(\gamma^{5/2})$. In addition, we shall show an explicit formula for the coefficients of $\gamma$ and $\gamma^2$ in the expansion.

To provide bias expansion, we introduce the necessary multi-index notation and definitions here. For two vectors $v=(v_1,\ldots,v_d)^{\top}$ and $\nu=(\nu_1,\ldots,\nu_d)^{\top}$, we write $v^{\nu}=v_1^{\nu_1}\cdots v_d^{\nu_d}$. For $\bm{a}=(a_1,\ldots,a_{\ell})^{\top}$,  $\ell\in\{1,2,\ldots\}$, where $a_j\in\{1,\ldots,d\}$, define the expectation of the product of partial derivatives of $g(\theta,X)$ by
\begin{equation*}
    m_{\bm{a}}(\theta) = \EE\Bigg(\frac{\partial g(\theta,X)}{\partial\theta_{a_1}}\cdots\frac{\partial g(\theta,X)}{\partial\theta_{a_{\ell}}}\Bigg).
\end{equation*}
For a vector $\bm{b}=(b_1,\ldots,b_d)^{\top}$ where $b_i\in\{0,1,2,3,4,5\}$, define $|\bm{b}|:=\sum_{i=1}^db_i$. Consider the stationary SGD process $\{\theta_k^{\circ}(\gamma)\}_{k\in\NN}$ and let $\mu_{\bm{b}}=\EE[\theta_1^{\circ}(\gamma)^{\bm{b}}]. $
For a function $f:\RR^d\mapsto\RR$, define 
\begin{equation*}
    f^{\bm{b}}(v) = \Big(\frac{1}{b_1!}\cdots\frac{1}{b_d!}\Big)\frac{\partial^{|\bm{b}|}f(v)}{\partial v_1^{b_1}\cdots \partial v_d^{b_d}}.
\end{equation*}
We impose the following condition on the smoothness of the objective function and the bounded variation of the derivatives. 
\begin{assumption}[Bounded variation of derivatives]
    \label{asm_g_smooth}
    For every $\bm{a}=(a_1,\ldots,a_{\ell})$ with $\ell\in\{1,2,3,4\}$ and $a_j\in\{1,\ldots,d\}$, assume that $m_{\bm{a}}(\theta)$ is five times continuously differentiable and $\sup_{\theta\in\RR^d}|m_{\bm{a}}^{\bm{b}}(\theta)|<\infty$ for every $\bm{b}=(b_1,\ldots,b_d)^{\top}$ satisfying $\ell\le|\bm{b}|=\sum_{i=1}^db_i\le 5$, where $b_i\in\{0,1,2,3,4,5\}$.
\end{assumption}

\begin{theorem}[Bias expansion]
    \label{thm_bias_expansion}
    Suppose that Assumptions \ref{asm_conv}--\ref{asm_g_smooth} hold for some $p\ge5$ and the constant step size $\gamma\in(0,\gamma(p))$. Then, there exist vectors $\Delta_1,\Delta_2\in\RR^d$ that are independent of $\gamma$ such that, as $\gamma\rightarrow0$,
\begin{align}
    \label{eq_bias}
    \theta_{\infty}^*(\gamma) - \theta^{*} = \gamma \Delta_1 + \gamma^2\Delta_2 + \R_{\gamma}^{(3)},
\end{align}
where $\R_{\gamma}^{(3)}\in\RR^d$ with $|\R_{\gamma}^{(3)}| \leq C\gamma^{5/2}$ for some constant $C>0$ independent of $\gamma$. 
\end{theorem}
Since the detailed definitions of $\Delta_1$ and $\Delta_2$ are quite involved, we defer them to the proof of Theorem~\ref{thm_bias_expansion} in Appendix~\ref{sec_proof}. 
Compared to the bias expansion in the SDE literature, including the uniform-in-time weak-error analysis (\cite{feng_uniform--time_2020}) and the invariant-measure expansions for numerical discretizations of ergodic SDEs, e.g., Langevin-type dynamics (\cite{abdulle_high_2014}), our bias expansion in Theorem~\ref{thm_bias_expansion} shares a similar flavor, but with a main distinction that \eqref{eq_bias} is derived directly for the stationary law of the discrete-time SGD, rather than for the invariant law of a prescribed limiting SDE or for a numerical discretization of that SDE.

Now we proceed with the basic idea of the Richardson-Romberg extrapolation in SGD. In particular, we compute the trajectory of the SGD at multiple step-sizes and then combine these trajectories in a specific manner to produce a new, bias-reduced trajectory. This approach allows for a more accurate convergence to the solution, leveraging the information from multiple step-sizes to correct the inherent bias in constant step-size SGD. Specifically, we consider three different constant step sizes $a_1\gamma$, $a_2\gamma$ and $a_3\gamma$ with $\gamma>0$ and the constants $a_1,a_2,a_3>0$. Then, we have three sequences of stochastic gradient descent iterates $\{\theta_{k}(a_1\gamma)\}_{k\in\NN}$, $\{\theta_{k}(a_2\gamma)\}_{k\in\NN}$ and $\{\theta_{k}(a_3\gamma)\}_{k\in\NN}$ respectively, with the initial point $\theta_{0}\in\RR^d$. Motivated by Theorem \ref{thm_bias_expansion}, we propose to estimate $\theta^{*}$ via the Richardson-Romberg extrapolation as follows (\cite{Hardle1986,bach_bridging_2020}),
\begin{align}
    \label{eq_extra_asgd}
    \hat{\theta}_{n} (\gamma) = b_1\bar{\theta}_{n}(a_1\gamma) + b_2\bar{\theta}_{n}(a_2\gamma) + b_3\bar{\theta}_{n}(a_3\gamma) := \frac{1}{n} \sum_{k = 1}^{n} \vartheta_{k}(\gamma) ,
 \end{align}
where $\vartheta_{k}(\gamma) = b_{1} \theta_{k}(a_{1} \gamma) + b_{2} \theta_{k}(a_{2} \gamma) + b_{3} \theta_{k}(a_{3} \gamma)$. 
One can see that the constants $b_1,b_2,b_3$ defined above satisfy $b_1+b_2+b_3=1$, $a_1b_1+a_2b_2+a_3b_3=0$ and $a_1^2b_1+a_2^2b_2+a_3^2b_3=0$, which leads to a cancellation of the terms $\gamma\Delta_1$ and $\gamma^2\Delta_2$ in (\ref{eq_bias}). Detailed solution is deferred to~\eqref{eq_extrapo_b} in Appendix~\ref{sec_longrun_appendix}.
Notably, this Richardson-Romberg estimator $\hat{\theta}_{n} (\gamma)$ has the same stochastic convergence rate as the original ASGD $\bar{\theta}_{n} (\gamma)$ because it is still an averaged estimator based on dependent SGD iterates. However, it can effectively reduce the bias introduced by the constant step size. In Appendix~\ref{sec_more_simu}, we have provided simulation results to demonstrate this benefit from the Richardson-Romberg extrapolation.

Before presenting the asymptotic distribution of $\hat{\theta}_{n}(\gamma)$, we shall introduce some basic definitions and notation. Denote the stationary SGD sequence with the step size $a\gamma$ by $\{\theta_{k}^{\circ}(a\gamma)\}_{k \in \mathbb{Z}}$, for some constant $a>0$, which follows the stationary distribution $\pi_{a\gamma}$. Moreover, for the same constants $a_i,b_i$, $i=1,2,3$, specified in (\ref{eq_extra_asgd}), we define the stationary process $\{\vartheta_{k}^{\circ}(\gamma)\}_{k \in \mathbb{Z}}$, where  
\begin{align}
\label{eq_stationary_difference}
    \vartheta_{k}^{\circ}(\gamma) = b_1 \theta_{k}^{\circ}(a_1\gamma) + b_2 \theta_{k}^{\circ}(a_2\gamma) + b_3 \theta_{k}^{\circ}(a_3\gamma).
\end{align}
Consequently, the long run covariance matrix of the stationary sequence $\{\vartheta_{k}^{\circ}(\gamma)\}_{k \in \mathbb{Z}}$ is 
\begin{align}
    \label{eq_long_run_variance}
    \Sigma(\gamma) = \sum_{k \in \mathbb{Z}} \mathrm{cov}\{\vartheta_{0}^{\circ}(\gamma), \vartheta_{k}^{\circ}(\gamma)\}. 
\end{align}
Throughout this paper, assume that $\lambda_{\min}(\Sigma(\gamma)) \geq \tilde{c}_{0} > 0$ for some constant $\tilde{c}_{0} < \infty$.
One can show that the process $\{\vartheta_{k}^{\circ}(\gamma)\}_{k \in \mathbb{Z}}$ satisfies the geometric moment contraction as in~\eqref{21} with $\|\vartheta_{k}(\gamma) - \vartheta_{k}^{\circ}(\gamma) \|_{p} \leq C(a, \theta_{0}) \bar{\rho}^{k}$,
where $\bar{\rho} = \max\{\rho_{a_{1} \gamma, p}, \rho_{a_{2} \gamma, p}, \rho_{a_{3} \gamma, p}\}$ and $C(a, \theta_{0})$ is a constant depending only on $(a_{1}, a_{2}, a_{3})$ and $\theta_0$. We defer the detailed derivation in Appendix~\ref{sec_longrun_appendix}.

Corollary \ref{cor_clt_RR} below follows from Theorems \ref{thm2_quenched_CLT} and \ref{thm_invar_prin}, whose target is the stationary mean $\vartheta^*_{\infty}(\gamma)$, and the distance $\vartheta^*_{\infty}(\gamma)-\theta^*$ can be characterized following Theorem~\ref{thm_bias_expansion}. As a direct application of Theorem~\ref{thm2_quenched_CLT}, by the Cr\'amer-Wold device, the CLT also holds for the Richardson-Romberg bias corrected estimate~\eqref{eq_extra_asgd} as stated in Corollary~\ref{cor_clt_RR}, which is a linear combination of $\bar{\theta}_n(\gamma_s)$, $s=1,\ldots,\ell$. Furthermore, the quenched invariance principle in Theorem~\ref{thm_invar_prin} similarly holds for the extrapolated ASGD with the same approximation rate $o_{\PP}(n^{1/p})$.

\begin{corollary}[Quenched CLT and invariance principle for the extrapolated ASGD]
\label{cor_clt_RR}
Consider the extrapolated ASGD estimator $\hat{\theta}_n(\gamma)$ in (\ref{eq_extra_asgd}). When $p=2$, under the conditions of Theorem~\ref{thm2_quenched_CLT}, for any initial point $\theta_0\in\RR^d$, 
\begin{align*}
    \sqrt{n}[\hat{\theta}_{n}(\gamma) - \vartheta_{\infty}^{*}(\gamma)] \Rightarrow N(0, \Sigma(\gamma)), \quad \mathrm{as} \enspace n \rightarrow \infty,
\end{align*}
where the center $\vartheta_{\infty}^*(\gamma)= b_1 \theta_{\infty}^*(a_1\gamma) + b_2 \theta_{\infty}^*(a_2\gamma) + b_3 \theta_{\infty}^*(a_3\gamma)$ with $\theta_{\infty}^*(\gamma)$ defined in (\ref{eq_SGD_mean}) and the long-run covariance matrix $\Sigma(\gamma)$ is defined in~\eqref{eq_long_run_variance}. Furthermore, when $p > 2$, under the conditions of Theorem~\ref{thm_invar_prin}, (\ref{eq_invar_prin}) holds with $\Sigma^{\circ}$ therein replaced by $\Sigma(\gamma)$ and sums $\hat S_i^{\circ}=\sum_{k=1}^i[\vartheta_k^{\circ}(\gamma)-\vartheta_{\infty}^{*}(\gamma)]$ and $\hat S_i = \sum_{k=1}^i [\vartheta_k(\gamma) - \vartheta_{\infty}^{*}(\gamma)]$ with any initial point $\theta_0\in\RR^d$.
\end{corollary}
We note that $\theta^*$ is the population minimizer, $\theta_\infty^*(\gamma)=\int y\,\pi_\gamma(dy)$ is the stationary mean of SGD with step size $\gamma$, and here $\vartheta_\infty^*(\gamma)=\sum_{j=1}^{\ell}b_j\theta_\infty^*(a_j\gamma)$ is the stationary mean of the extrapolated process $\{\vartheta_k(\gamma)\}_{k\in\ZZ}$.

\begin{remark}[Refined bias expansion]
We derive the theoretical properties of the SGD estimator with three different step sizes utilized in the Richardson-Romberg extrapolation in Corollary \ref{cor_clt_RR}, which can be extended to even more step sizes. However, these results can be hardly achieved by using the second-moment Wasserstein distance applied in \textcite{bach_bridging_2020}, since the SGD sequence can be non-stationary and the bound in the $\L^2$-norm may not be sufficient and is nontrivial to be generalized to higher orders. Therefore, we can see the advantage of using the Euclidean distance (i.e., the GMC property) to characterize how the SGD iterates depend on the initial conditions. Here again, we shall emphasize the benefits from leveraging the technical tools in nonlinear time series to study the SGD processes.
\end{remark}

\subsection{Online Estimators for Long-Run Covariance Matrices}

The long-run covariance matrix $\Sigma(\gamma)$ defined in~\eqref{eq_long_run_variance} appears in the limiting normal distribution in the quenched CLT (see Corollary~\ref{cor_clt_RR}). Online estimation of $\Sigma(\gamma)$ has been studied in the time series and SGD literature (\cite{Wu2009, Xiao2011, Zhu2023}). Building on the recursive estimation methods of \textcite{Wu2009, Xiao2011}, we establish convergence rates for a recursive estimator and its bias-reduced variant for constant step-size SGD with arbitrary initialization. These estimators retain the rates available in the stationary setting, providing theoretical guarantees for their use in online inference for SGD.

Denote the online long-run covariance estimator at the $n$-th iteration by $\hat\Sigma_n(\gamma)$ and denote the bias-reduced variant by $\tilde\Sigma_n(\gamma)$. Since the notation is quite involved, we defer the definitions of these two estimators to Sections~\ref{subsec_online_inference}~and~\ref{subsec_online_inference_bias_reduce} in Appendix~\ref{sec_longrun_appendix}. The online inference algorithms with these two estimators implemented can be found in Algorithms~\ref{algo1}~and~\ref{algo2}, respectively. In this section, we provide the theoretical guarantees for the long-run covariance estimators. Let $\{\eta_{m}\}_{m \in \mathbb{N}}$ be a strictly increasing positive integer-valued sequence. For each $m \in \mathbb{N}$, we define the consecutive blocks $\{B_{m}\}_{m \in \mathbb{N}}$ as $B_{m} = \{\eta_{m}, \eta_{m} + 1, \ldots, \eta_{m + 1} - 1\}$.
Throughout, assume that $\eta_{m + 1} - \eta_{m} \to \infty$ as $m \to \infty$. Then one can construct the non-overlapping recursive estimators $\hat\Sigma_{n}(\gamma)$ and $\tilde\Sigma_{n}(\gamma)$ (see~\eqref{eq_recursive_estimator_sigma_hat}~and~\eqref{eq_recursive_longrun_est_debias}).

\begin{theorem}[Rate of convergence of $\hat\Sigma_n(\gamma)$]
\label{Theorem_Consistency_recursive_sigma}
Let $\eta_{m} = \lfloor c_{1} m^{\beta} \rfloor$ for some $c_{1} > 0$ and $\beta > 1$. Suppose that Assumption~\ref{asm_conv} and \ref{asm_Lip} hold with $p = 4$ and $a_{1} \gamma, a_{2} \gamma, a_{3} \gamma \in (0, \gamma(4))$. Then, $\mathbb{E} |\hat{\Sigma}_{n}(\gamma) - \Sigma(\gamma)|_{2}^{2} \lesssim n^{(2/\beta - 2) \vee (-1/\beta)}.$ 
\end{theorem}

\begin{remark}
\label{remark_mse}
Theorem~\ref{Theorem_Consistency_recursive_sigma} establishes the mean squared error (MSE) of the recursive estimator for the long-run covariance matrix $\Sigma (\gamma)$. In particular, we have $\mathbb{E} |\hat{\Sigma}_{n}(\gamma) - \Sigma(\gamma)|_{2}^{2} \lesssim n^{-2/3}$ when $\beta = 3/2$. Recently, \textcite{Zhu2023} proposed a similar recursive estimator for $\Sigma(\gamma)$ in the context of SGD estimation with decaying step size $\gamma_{t} = c_{0} t^{-\alpha}$ for some $\alpha \in (1/2, 1)$. Under some regularity conditions, the mean squared error of their recursive estimator is $n^{-(1 - \alpha)/2}\gg n^{-1/4}$.  
\qed
\end{remark}


\begin{theorem}[Rate of convergence of $\tilde\Sigma_n(\gamma)$]
\label{Theorem_Consistency_recursive_sigma_new}
Suppose that Assumption~\ref{asm_conv} and \ref{asm_Lip} hold with $p = 4$ and $a_{1} \gamma, a_{2} \gamma, a_{3} \gamma \in (0, \gamma(4))$. Define a sequence of positive integers $\{\tau(m)\}_{m \in \mathbb{N}}$ such that $\tau(m) \leq (\eta_{m + 1} - \eta_{m})$ for all $m \in \mathbb{N}$. 
\begin{itemize}
    \item[(1)] Let $\eta_{m} = \lfloor c_{1} m^{\beta} \rfloor$ and $\tau(m) = \lfloor c_{2} \log m \rfloor$ for some $c_{1} > 0$, $\beta > 1$ and $c_{2} > 0$ with $2 c_{2} \log (\bar{\rho}) \leq 2 \beta - 3$. We have $\mathbb{E} |\tilde{\Sigma}_{n}(\gamma) - \Sigma(\gamma)|_{2}^{2} \lesssim n^{-1/\beta}. $
    \item[(2)] Let $\eta_{m} = \lfloor \bar{c}_{1} m \log m \rfloor$ and $\tau(m) = \lfloor \bar{c}_{2} \log m \rfloor$ for some $\bar{c}_{1} > \bar{c}_{2} > 0$ with $2 \bar{c}_{2} \log (\bar{\rho}) \leq -1$. Then, $\mathbb{E} |\tilde{\Sigma}_{n}(\gamma) - \Sigma(\gamma)|_{2}^{2} \lesssim (\log n)/n. $
\end{itemize}
\end{theorem}
We remark that the convergence rate in Theorem~\ref{Theorem_Consistency_recursive_sigma_new}~(1) is quite sharp, by noting that the mean squared error for a parametric model is typically of order $O(1/n)$.

\section{Numerical Experiments}\label{sec_numerical}
This section is devoted to the experiments on simulated data to demonstrate the validity of our proposed online inference methods. 
Specifically, we consider two classes of examples: linear regression and logistic regression.

\subsection{Linear regression and logistic regression}

We generate two sequences of i.i.d. observation pairs $\{(\bm{z}_{1,t},y_{1,t})\}_{t\ge1}$ and $\{(\bm{z}_{2,t},y_{2,t})\}_{t\ge1}$ for the two regression models, respectively. Let $\bm{x}_{1,t}=(\bm{z}_{1,t},y_{1,t})$ and $\bm{x}_{2,t}=(\bm{z}_{2,t},y_{2,t})$. We denote the true unknown parameter in the models by $\theta_1^*$ and $\theta_2^*$.

First, we generate $\bm{z}_{1,t},\,\bm{z}_{2,t}\in\RR^d$ from the standard normal distribution, i.e., $\bm{z}_{1,t} \sim N(0,\bm{I}_d)$, $\bm{z}_{2,t} \sim N(0,\bm{I}_d)$. 
Then, we generate $y_{1,t}=\bm{z}_{1,t}^{\top}\theta_1^*+\epsilon_t$, where $\epsilon_t\sim N(0,1)$, and $y_{2,t}\sim\mathrm{Bernoulli}\{1/[1+\exp(-\bm{z}_{2,t}^{\top}\theta_2^*)]\}$. The corresponding loss functions are $g_1(\theta,\bm{z}_{1,t},y_{1,t})=(\bm{z}_{1,t}^{\top}\theta-y_{1,t})^2/2$ and $g_2(\theta,\bm{z}_{2,t},y_{2,t}) =(1-y_{2,t})\bm{z}_{2,t}^{\top}\theta +\log\{1+\exp(-\bm{z}_{2,t}^{\top}\theta)\}$. The true parameter $\theta^*\in\RR^d$ is linearly spaced between 0 and 1. Since the logistic loss $g_2$ is convex but not strongly convex (\cite{bach_adaptivity_2014}), we add the regularization term $0.005|\theta|^2$ to make it strongly convex.

\subsection{Empirical Performance of the Online Estimators}
We first evaluate the empirical performance of our proposed online inference method. Note that in most general cases, the true long-run covariance matrix $\Sigma(\gamma)$ does not have a closed form solution. Therefore, throughout this section, we use the long-run covariance estimate computed at iteration $n=10^8$ as a numerical benchmark.
To compare our two online estimators, without and with bias reduction, we consider the one-dimensional case to report the bias of the two estimators relative to the iteration steps. All of our measurements are averaged over 500 independent runs. As shown in Figure~\ref{fig_var_compare}, the results suggest that both the non-debiased (\textit{in red}) and the debiased (\textit{in green}) estimators converge fast as the step number increases. Moreover, the advanced version with bias reduction yields a better estimated value which is closer to the true limiting variance. Here we set the constant step size to be $0.025$. We also investigate other scenarios with different step sizes and we observe the similar results that the debiased estimator robustly outperform the non-debiased version. We defer the details to Figure~\ref{fig_diff} in Supplement \ref{sec_more_simu}.

\subsection{Coverage Probabilities of Confidence Intervals}
In this section, we construct online confidence intervals for the linear functional $\One_d^{\top}\vartheta_\infty^*(\gamma)$ using our two online estimators of the long-run covariance matrix. Both intervals target $\One_d^{\top}\vartheta_\infty^*(\gamma)$. By Theorem~\ref{thm_bias_expansion}, $\vartheta_\infty^*(\gamma)-\theta^*=O(\gamma^{5/2})$ for the three-step extrapolation.
Both linear regression and logistic regression are investigated. In particular, at step $k$, using our online estimator $\hat\Sigma$ of $\Sigma$, we can construct the $100(1-\alpha)\%$ confidence intervals for some given $\alpha\in(0,1)$ as follows:
\begin{equation}
    \label{eq_CI}
    \Big[\One_d^{\top}\hat\theta_k(\gamma) - z_{1-\alpha/2}\sqrt{(\One_d^{\top}\hat\Sigma_k(\gamma)\One_d)/k}, \, \One_d^{\top}\hat\theta_k(\gamma) + z_{1-\alpha/2}\sqrt{(\One_d^{\top}\hat\Sigma_k(\gamma)\One_d)/k}\Big],
\end{equation}
where $z_{1-\alpha/2}$ is the $(1-\alpha/2)$-percentile of the standard normal distribution. The bias-reduced confidence interval can be similarly constructed with $\hat\Sigma_{n}$ in (\ref{eq_CI}) replaced by $\tilde\Sigma_{n}$.

\begin{figure}[!tb]
    \centering
    \begin{subfigure}{0.49\textwidth}
        \centering
        \includegraphics[width=1\linewidth]{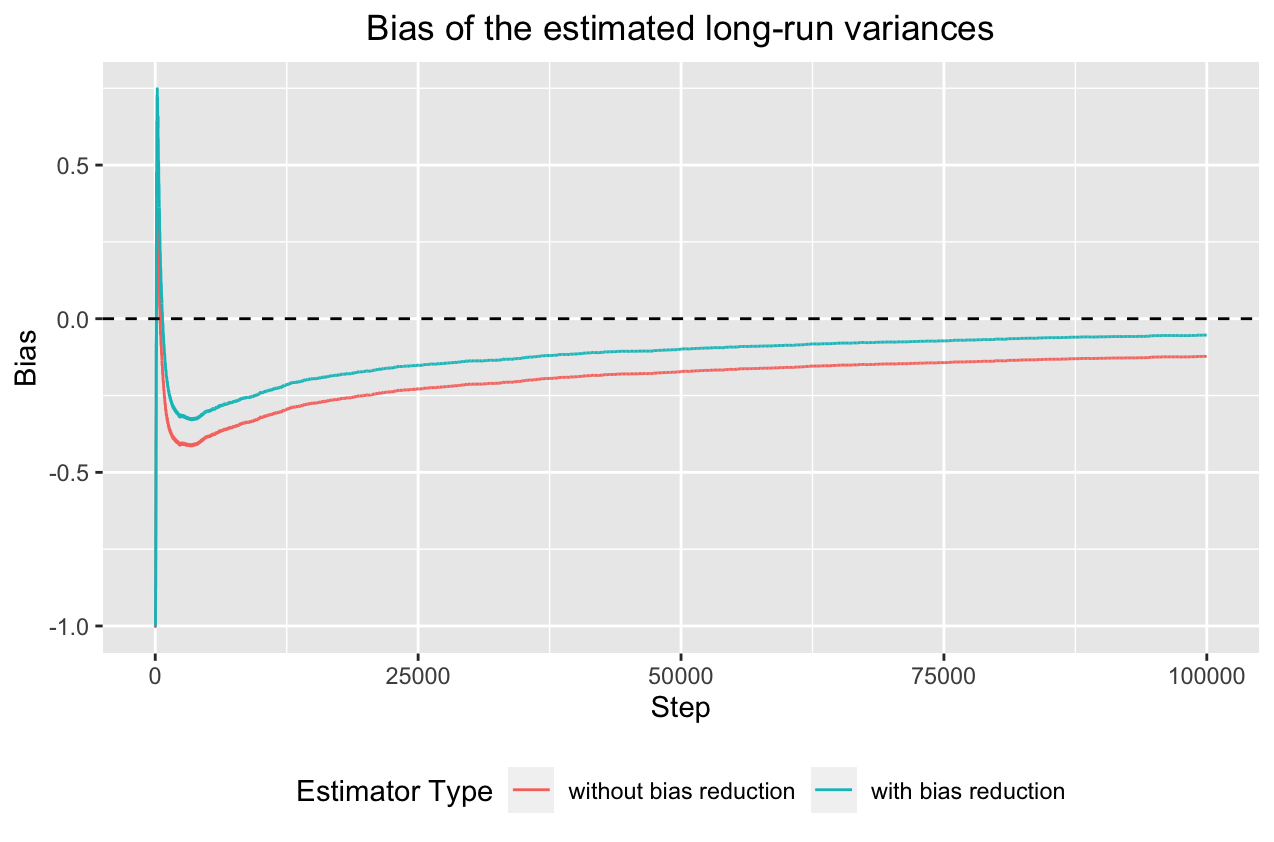}
        \caption{Bias of estimators.}
        \label{fig_var_compare}
    \end{subfigure}   
    \begin{subfigure}{0.49\textwidth}
        \centering
        \includegraphics[width=1\linewidth]{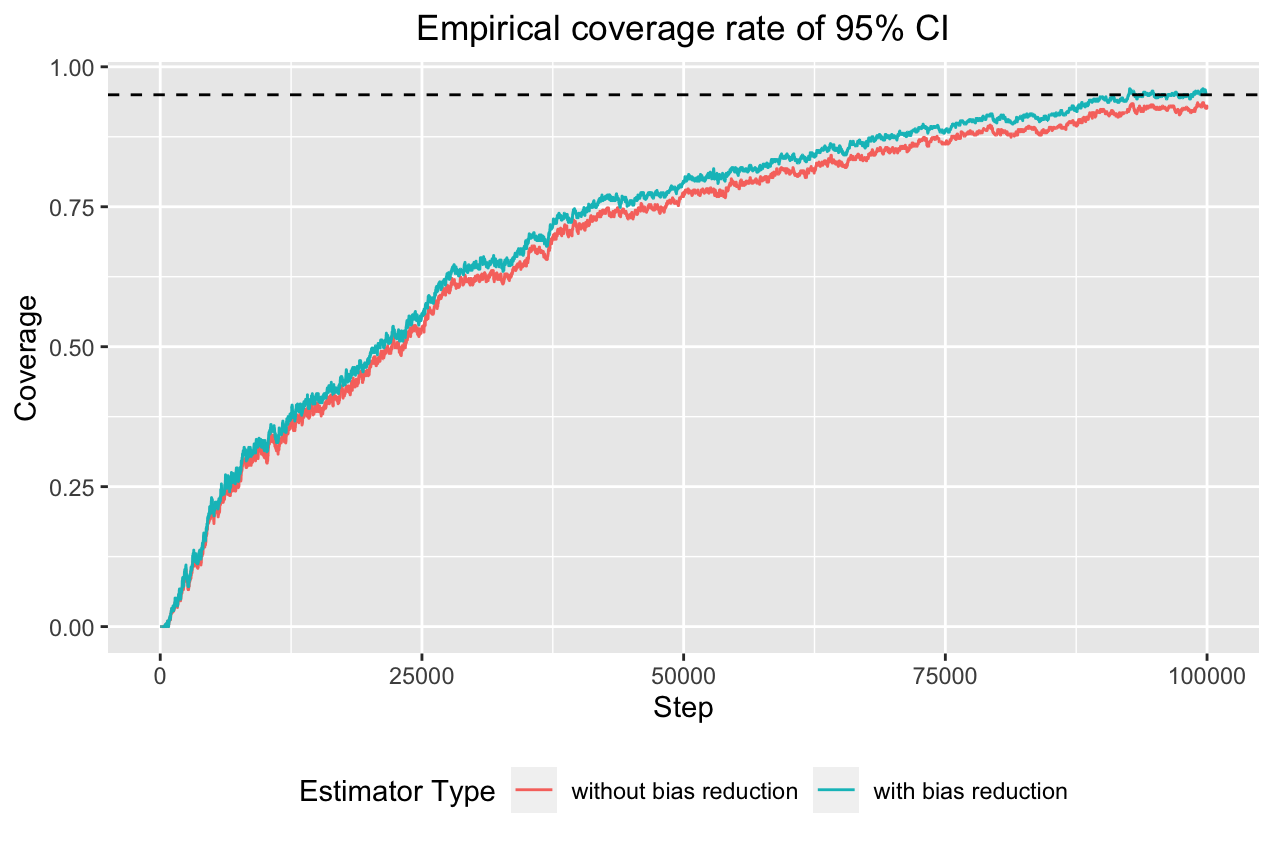}
        \caption{Empirical coverage probability.}
        \label{fig_coverage_compare}
    \end{subfigure}   
    \caption{Recursively estimated long-run variance averaged over 500 independent runs (Linear regression $\gamma=0.025$). (Left) Bias of the two estimators. (Right) Empirical coverage probability of 95\% confidence interval. The black dashed line indicates the nominal probability = 0.95.}
\end{figure}

To evaluate the coverage probability, we take the average of 500 independent repetitions. In Figure~\ref{fig_coverage_compare}, we present the convergence trace of the empirical coverage rate of 95\% CI with the constant step size $\gamma=0.025$ in linear regression. For the cases with other constant step sizes or the logistic regression, we report the coverage rates in Table~\ref{table_mse_longrun} with the standard errors shown in the brackets. Our results show that for both the non-debiased (\textit{in red}) and the debiased (\textit{in green}), the empirical
coverage rates converge to the nominal probability 95\%, while the debiased version exhibits an enhanced coverage rate.

\begin{table}[!tbhp]
    \centering
    \caption{Empirical coverage probability of 95\% confidence interval for linear regression and logistic regression over 500 independent runs. For each step size, the first row represents the estimator without bias reduction, and the second row represents the estimator with bias reduction. The standard error is reported in the brackets.}
    \label{table_mse_longrun}
    \begin{tabular}{c|c|c|c|c}
         \hline\hline
         \multicolumn{5}{c}{Linear regression} \\
         \hline
        step size & $n=5000$ & $n=10^4$ & $n=5\times10^4$ & $n=10^5$ \\
        \hline
         \multirow{2}{4em}{$0.0125$} & 0.194 (0.0177) & 0.332 (0.0211) & 0.796 (0.0180) & 0.904 (0.0132) \\
         & 0.198 (0.0178) & 0.364 (0.0215) & 0.854 (0.0158) & 0.948 (0.0099) \\
         \hline
         \multirow{2}{4em}{$0.025$}  & 0.278 (0.0200) & 0.402 (0.0219) & 0.844 (0.0162) & 0.912 (0.0127) \\
         & 0.286 (0.0202) & 0.408 (0.0220) & 0.854 (0.0158) & 0.954 (0.0094) \\            \hline
         \multirow{2}{4em}{$0.05$}  & 0.266 (0.0198) & 0.380 (0.0217) & 0.810 (0.0175) & 0.910 (0.0128) \\
         & 0.280 (0.0201) & 0.396 (0.0219) & 0.832 (0.0167) & 0.946 (0.0101) \\
         \hline\hline
         \multicolumn{5}{c}{Logistic regression} \\
         \hline
         step size & $n=5\times10^4$ & $n=10^5$ & $n=5\times10^5$ & $n=10^6$ \\
        \hline
         \multirow{2}{4em}{$0.0125$}  & 0.174 (0.0170) & 0.268 (0.0198) & 0.718 (0.0201) & 0.894 (0.0138) \\
         & 0.176 (0.0170) & 0.272 (0.0199) & 0.720 (0.0201) & 0.906 (0.0131) \\
         \hline
         \multirow{2}{4em}{$0.025$}  & 0.180 (0.0172) & 0.324 (0.0209) & 0.720 (0.0201) & 0.902 (0.0133) \\
         & 0.184 (0.0173) & 0.328 (0.0210) & 0.722 (0.0200) & 0.914 (0.0125) \\
         \hline
         \multirow{2}{4em}{$0.05$}  & 0.252 (0.0194) & 0.380 (0.0217) & 0.774 (0.0187) & 0.908 (0.0129) \\
         & 0.254 (0.0195) & 0.386 (0.0218) & 0.780 (0.0185) & 0.922 (0.0120) \\
         \hline\hline
    \end{tabular}
\end{table}

\section*{Acknowledgment}
The authors are grateful to the Joint Editor, Associate Editor and two reviewers for their thoughtful comments, which led to substantial improvements to the paper. Jiaqi Li gratefully acknowledges the support of the Department of Statistics at the University of Chicago, where most of this work was carried out during her visit as a PhD student from 2022 to 2024.

\section*{Conflict of Interest}
The authors declare no conflicts of interest.

\section*{Funding}
Jiaqi Li’s research is partially supported by NSF DMS-2515926. Wei Biao Wu has received funding from NSF DMS-2610302 and NSF DMS-2311249.

\section*{Data Availability}
All data used in this study are simulated. The code for
generating the data and reproducing the numerical results
is available at \url{https://github.com/jiaqili97/SGD_GMC}.



\printbibliography[section=0]

\clearpage
\newrefsection

\setcounter{page}{1}

\begin{appendix}

\begin{center}
    {\Large\textbf{Supplemental Material}}\\[0.5em]
    \textbf{Stochastic Gradient Descent:
    A Nonlinear Time Series Perspective}\\[0.5em]
    Jiaqi Li, Zhipeng Lou, Stefan Richter and Wei Biao Wu
\end{center}

This supplement provides the construction and implementation of the long-run covariance estimators in Section~\ref{sec_longrun_appendix}, results for heavy-tailed and projected SGD in Section~\ref{subsec_heavy_tail}, proofs of the main results in Section~\ref{sec_proof}, additional simulation results in Section~\ref{sec_more_simu}, and additional references in Section~\ref{sec_additional_literature}.

\section{Estimation of Long-Run Covariance}\label{sec_longrun_appendix}

In this section, we first show the GMC property of the Richardson-Romberg extrapolated estimator $\hat{\theta}_n(\gamma)$, and then introduce the details of the two proposed online long-run covariance estimators $\hat\Sigma_n(\gamma)$ and $\tilde\Sigma_n(\gamma)$. The pseudo codes for implementation are provided in Algorithms~\ref{algo1}~and~\ref{algo2}.

Recall the Richardson-Romberg extrapolated estimator $\hat{\theta}_{n} (\gamma) = b_1\bar{\theta}_{n}(a_1\gamma) + b_2\bar{\theta}_{n}(a_2\gamma) + b_3\bar{\theta}_{n}(a_3\gamma)$ defined in~\eqref{eq_extra_asgd}. For pairwise distinct $a_1,a_2,a_3$, by solving the equations $b_1+b_2+b_3=1$, $a_1b_1+a_2b_2+a_3b_3=0$ and $a_1^2b_1+a_2^2b_2+a_3^2b_3=0$, we obtain
\begin{align}
    \label{eq_extrapo_b}
    b_1 & = \frac{a_2a_3}{a_1^2-a_1a_2-a_1a_3+a_2a_3}, \nonumber \\
    b_2 & = -\frac{a_1a_3}{a_1a_2-a_1a_3-a_2^2+a_2a_3}, \nonumber \\
    b_3 & = \frac{a_1a_2}{a_1a_2-a_1a_3-a_2a_3+a_3^2}.
\end{align}
When two of the coefficients coincide, say $a_1\neq a_2= a_3$, then by solving $b_1+b_2=1$ and $a_1b_1+a_2b_2=0$, one can achieve $b_1=a_2/(a_2-a_1)$, $b_2=-a_1/(a_2-a_1)$. Regarding the bias of the resulting extrapolated estimator, only the first-order bias cancels, leaving $O(\gamma^2)$.

Next, we show that $\hat{\theta}_{n}(\gamma)$ satisfies the GMC property. By~\eqref{eq_extrapo_b}, it follows that 
\begin{align*}
    |b_{1}| + |b_{2}| + |b_{3}| \leq \frac{3 \max_{1\leq i\leq 3} |a_{i}|^{2}}{\min_{1\leq i\neq j\leq 3} |a_{i} - a_{j}|^{2}}. 
\end{align*}
Throughout this section, we assume that $a_{i} \gamma \in (0, \gamma(p))$ for all $1\leq i\leq 3$. Let $\bar{\rho} = \max\{\rho_{a_{1} \gamma, p}, \rho_{a_{2} \gamma, p}, \rho_{a_{3} \gamma, p}\}$. By Theorem~\ref{thm_GMC}, 
\begin{align*}
    \|\vartheta_{k}(\gamma) - \vartheta_{k}^{\circ}(\gamma)\|_{p} &\leq \sum_{i = 1}^{3} |b_{i}| \|\theta_{k}(a_{i} \gamma) - \theta_{k}^{\circ}(a_{i} \gamma)\|_{p} \cr 
    &\leq \sum_{i = 1}^{3} |b_{i}| \rho_{a_{i} \gamma, p}^{k} \|\theta_{0} - \theta_{0}^{\circ}(a_{i}\gamma)\|_{p} \cr 
    &\leq \frac{3 \max_{1\leq i\leq 3} |a_{i}|^{2}}{\min_{1\leq i\neq j\leq 3} |a_{i} - a_{j}|^{2}} \cdot \max_{1\leq i\leq 3} \|\theta_{0} - \theta_{0}^{\circ}(a_{i}\gamma)\|_{p} \cdot \bar{\rho}^{k} \cr 
    &=: C(a, \theta_{0}) \bar{\rho}^{k}. 
\end{align*}


\subsection{Online Estimator for Long-Run Covariance Matrices}
\label{subsec_online_inference}
In this section, we construct recursive consistent estimator $\hat\Sigma_n(\gamma)$ for the long-run covariance matrix $\Sigma(\gamma)$ defined in~\eqref{eq_long_run_variance}. 

Recall that $\{\eta_{m}\}_{m \in \mathbb{N}}$ is a strictly increasing positive integer-valued sequence. For each $m \in \mathbb{N}$, we have defined the consecutive blocks $\{B_{m}\}_{m \in \mathbb{N}}$ as   
\begin{align*}
    B_{m} = \{\eta_{m}, \eta_{m} + 1, \ldots, \eta_{m + 1} - 1\}.  
\end{align*}
Throughout, we assume that $\eta_{m + 1} - \eta_{m} \to \infty$ as $m \to \infty$. Define $\psi(n) = \max\{m \in \mathbb{N} : \eta_{m} \leq n\}$ and $\phi(n) = \eta_{\psi(n)}$ for each $n \in \mathbb{N}$. Then the nonoverlapping recursive estimator of the long-run covariance matrix $\Sigma(\gamma)$ is defined by $V_{n}(\gamma)/n$, where
\begin{align}
\label{eq_recursive_estimator_sigma_hat}
    V_{n}(\gamma) = \sum_{m = 1}^{\psi(n) - 1} \Bigg(\sum_{k \in B_{m}} \{\vartheta_{k}(\gamma) - \hat{\theta}_{n}(\gamma)\}\Bigg)^{\otimes 2} + \Bigg(\sum_{k = \phi(n)}^{n} \{\vartheta_{k}(\gamma) - \hat{\theta}_{n}(\gamma)\}\Bigg)^{\otimes 2}. 
\end{align}
For simplicity of notation, we write $\delta_{\eta}(n) = n - \phi(n) + 1$,  
\begin{align*}
    \mathcal{S}_{m}(\gamma) = \sum_{k \in B_{m}} \vartheta_{k}(\gamma) \enspace \quad\mathrm{and}\quad \enspace \mathcal{R}_{n}(\gamma) = \sum_{k = \phi(n)}^{n} \vartheta_{k}(\gamma). 
\end{align*}
To facilitate the recursive computation of $V_{n}(\gamma)$, we rewrite it as follows, 
\begin{align}
\label{eq_Vn_equation}
    V_{n}(\gamma) &= \Bigg(\sum_{m = 1}^{\psi(n) - 1} \mathcal{S}_{m}(\gamma)^{\otimes 2} + \mathcal{R}_{n}(\gamma)^{\otimes 2}\Bigg) + \Bigg(\sum_{m = 1}^{\psi(n) - 1} |B_{m}|^{2} + |\delta_{\eta}(n)|^{2}\Bigg) \hat{\theta}_{n}(\gamma)^{\otimes 2} \cr 
    & \quad - \Bigg(\sum_{m = 1}^{\psi(n) - 1} |B_{m}| \mathcal{S}_{m}(\gamma) + \delta_{\eta}(n) \mathcal{R}_{n}(\gamma)\Bigg) \hat{\theta}_{n}(\gamma)^{\top} \cr
    & \quad - \hat{\theta}_{n}(\gamma) \Bigg(\sum_{m = 1}^{\psi(n) - 1} |B_{m}| \mathcal{S}_{m}(\gamma) + \delta_{\eta}(n) \mathcal{R}_{n}(\gamma)\Bigg)^{\top} \cr
    &=: \mathcal{V}_{n}(\gamma) + K_{n} \hat{\theta}_{n}(\gamma)^{\otimes 2} - H_{n}(\gamma) \hat{\theta}_{n}(\gamma)^{\top} - \hat{\theta}_{n}(\gamma) H_{n}(\gamma)^{\top}. 
\end{align}
Then, it reduces to recursively compute $\{\mathcal{V}_{n}(\gamma), K_{n}, H_{n}(\gamma), \hat{\theta}_{n}(\gamma)\}$ for any $n \in \mathbb{N}$. To this end, we propose a recursive algorithm which only requires $O(1)$ storage and establish the convergence rate of the proposed estimator.

The rational behind the recursive computation for $V_{n}(\gamma)$ is as follows: if $n + 1 < \eta_{\psi(n) + 1}$, then $n + 1$ still belongs to the block $B_{\psi(n)}$ and $\psi(n + 1) = \psi(n)$. Also we have $\mathcal{R}_{n + 1}(\gamma) = \mathcal{R}_{n}(\gamma) + \vartheta_{n + 1}(\gamma)$ and $\delta_{\eta}(n + 1) = \delta_{\eta}(n) + 1$. Consequently $\{K_{n + 1}, \mathcal{V}_{n + 1}(\gamma), H_{n + 1}(\gamma)\}$ can be recursively updated via $K_{n + 1} = K_{n} - |\delta_{\eta}(n)|^{2} + |\delta_{\eta}(n + 1)|^{2}$,
\begin{align*}
    \mathcal{V}_{n + 1}(\gamma) &= \mathcal{V}_{n}(\gamma) - \mathcal{R}_{n}(\gamma)^{\otimes 2} + \mathcal{R}_{n + 1}(\gamma)^{\otimes 2}, \cr
    H_{n + 1}(\gamma) &= H_{n}(\gamma) - \delta_{\eta}(n) \mathcal{R}_{n}(\gamma) + \delta_{\eta}(n + 1) \mathcal{R}_{n + 1}(\gamma).
\end{align*}
Otherwise, if $n + 1 = \eta_{\psi(n) + 1}$, we have $\psi(n + 1) = \psi(n) + 1$. Hence $\mathcal{R}_{n + 1}(\gamma) = \vartheta_{n + 1}(\gamma)$ and $\delta_{\eta}(n + 1) = 1$. In this case, $\{K_{n + 1}, \mathcal{V}_{n + 1}(\gamma), H_{n + 1}(\gamma)\}$ can be recursively updated via $K_{n + 1} = K_{n} + 1$, 
\begin{align*}
    \mathcal{V}_{n + 1}(\gamma) = \mathcal{V}_{n}(\gamma) + \mathcal{R}_{n + 1}(\gamma)^{\otimes 2} \enspace \mathrm{and} \enspace H_{n + 1}(\gamma) = H_{n}(\gamma) + \mathcal{R}_{n + 1}(\gamma).
\end{align*}
Consequently, given $\vartheta_{1}(\gamma), \ldots, \vartheta_{n}(\gamma)$, our estimator for the long-run covariance matrix $\Sigma (\gamma)$ is given by 
\begin{align}
    \label{eq_Simga_hat}
    \hat{\Sigma}_{n}(\gamma) = \frac{1}{n} V_{n}(\gamma). 
\end{align}
In summary, our recursive algorithm for $\hat\Sigma_{n}(\gamma)$ is stated in Algorithm~\ref{algo1} below.





\begin{algorithm}[hbt!]
   \caption{Recursive estimation of the long-run covariance matrix}
   \label{algo1}
   \footnotesize
   \KwData{Observations $X_1,X_2,\ldots,X_n$; objective function $g(\cdot)$; constant step size $\gamma$; predefined sequence $\{\eta_m\}_{m\in\NN}$}
   \KwResult{Extrapolated ASGD estimator $\hat\theta_{n+1}(\gamma)$; estimated long-run covariance matrix $\hat{\Sigma}_{n+1}(\gamma)$} 
   \Init{ $\theta_0(a_{1} \gamma)=\theta_0(a_{2} \gamma)= \theta_0(a_{3} \gamma)= \hat{\theta}_0(\gamma)=\vartheta_0(\gamma)=\mathcal{R}_0(\gamma)\gets0$,}{$\psi(0)\gets1$, $\delta_{\eta}(0)\gets0$, $K_0=H_0(\gamma)\gets0$, $\mathcal{V}_0(\gamma)=V_0(\gamma)\gets0$}
   \For{$n=0,1,2,3,\ldots$}{
   $\theta_{n+1}(a_{1} \gamma)\gets\theta_n(a_{1}\gamma)-a_{1}\gamma \nabla g(\theta_n(a_{1}\gamma),X_{n+1})$\Comment*[r]{SGD estimate}
   $\theta_{n+1}(a_{2}\gamma)\gets\theta_n(a_{2}\gamma)-a_{2}\gamma \nabla g(\theta_n(a_{2}\gamma),X_{n+1})$\;
   $\theta_{n+1}(a_{3}\gamma)\gets\theta_n(a_{3}\gamma)-a_{3}\gamma \nabla g(\theta_n(a_{3}\gamma),X_{n+1})$\;
   $\vartheta_{n+1}(\gamma)\gets b_{1} \theta_{n + 1} (a_{1} \gamma) + b_{2} \theta_{n + 1} (a_{2} \gamma) + b_{3} \theta_{n + 1} (a_{3} \gamma)$\;
   $\hat{\theta}_{n + 1}(\gamma) \gets \{n\hat{\theta}_{n}(\gamma) + \vartheta_{n + 1}(\gamma)\}/(n + 1)$ \Comment*[r]{ASGD estimate}
   \eIf{$n + 1 < \eta_{\psi(n) + 1}$}{
   $\mathcal{R}_{n + 1}(\gamma) \gets \mathcal{R}_{n}(\gamma) + \vartheta_{n + 1}(\gamma)$, $\delta_{\eta}(n + 1) \gets \delta_{\eta}(n) + 1$\;
   $K_{n + 1} \gets K_{n} - \delta_{\eta}^{2}(n) + \delta_{\eta}^{2}(n + 1)$, $\psi(n + 1) \gets \psi(n)$\;
   $H_{n + 1}(\gamma) \gets H_{n}(\gamma) - \delta_{\eta}(n) \mathcal{R}_{n}(\gamma) + \delta_{\eta}(n + 1) \mathcal{R}_{n + 1}(\gamma)$\;
   $\mathcal{V}_{n + 1}(\gamma) \gets \mathcal{V}_{n}(\gamma) - \mathcal{R}_{n}(\gamma)^{\otimes 2} + \mathcal{R}_{n + 1}(\gamma)^{\otimes 2}$\;
   }{
   $\mathcal{R}_{n + 1}(\gamma) \gets \vartheta_{n + 1}(\gamma)$, $\delta_{\eta}(n + 1) \gets 1$\;
   $\psi(n + 1) \gets \psi(n) + 1$\;
   $K_{n + 1} \gets K_{n} + 1$, $H_{n + 1}(\gamma) \gets H_{n}(\gamma) + \mathcal{R}_{n + 1}(\gamma)$\;
   $\mathcal{V}_{n + 1}(\gamma) \gets \mathcal{V}_{n}(\gamma) + \mathcal{R}_{n + 1}(\gamma)^{\otimes 2}$\;
   }
   $V_{n + 1}(\gamma) \gets \mathcal{V}_{n + 1}(\gamma) + K_{n + 1} \hat{\theta}_{n + 1}(\gamma)^{\otimes 2} - H_{n + 1}(\gamma) \hat{\theta}_{n + 1}(\gamma)^{\top} - \hat{\theta}_{n + 1}(\gamma) H_{n + 1}(\gamma)^{\top}$\;
   $\hat{\Sigma}_{n+1}(\gamma) \gets V_{n+1}(\gamma)/(n+1)$\Comment*[r]{Recursive long-run covariance matrix estimate}
   }
\end{algorithm}

\subsection{Bias-Reduced Online Estimator}\label{subsec_online_inference_bias_reduce}
In this section, we introduce a new estimator for the long-run variance based on an idea for bias reduction~(\cite{Xiao2011}). To illustrate the idea, we first consider the recursive estimator based on the stationary sequence $\{\vartheta_{k}^{\circ}(\gamma)\}_{k \in \mathbb{Z}}$ in~\eqref{eq_stationary_difference}. Recall that $\phi(k) = \eta_{m}$ for any $k \in \{1,...,n\}$ such that $\eta_{m} \leq k < \eta_{m + 1}$. Also, recall in (\ref{eq_SGD_mean}), we write $\theta_{\infty}^*(\gamma) = \mathbb{E} \{\theta_{k}^{\circ}(\gamma)\}$ for $\gamma > 0$. Based on $\{\vartheta_{k}^{\circ}(\gamma)\}_{k \in \mathbb{Z}}$, define $V_{n}^{\circ}(\gamma) = \sum_{k = 1}^{n} Q_{k}^{\circ}(\gamma)$, where for each $k \in \{1, \ldots, n\}$, 
\begin{align*}
    Q_{k}^{\circ}(\gamma) &= (\vartheta_{k}^{\circ}(\gamma) - \vartheta_{\infty}^*(\gamma))^{\otimes 2} + \sum_{\ell = \phi(k)}^{k - 1} (\vartheta_{k}^{\circ}(\gamma) - \vartheta_{\infty}^*(\gamma))(\vartheta_{\ell}^{\circ}(\gamma) - \vartheta_{\infty}^*(\gamma))^{\top}\cr
    &+ \sum_{\ell = \phi(k)}^{k - 1} (\vartheta_{\ell}^{\circ}(\gamma) - \vartheta_{\infty}^*(\gamma))(\vartheta_{k}^{\circ}(\gamma) - \vartheta_{\infty}^*(\gamma))^{\top}. 
\end{align*}
In view of~\eqref{eq_long_run_variance}, the bias of $Q_{k}^{\circ}(\gamma)$ is  
\begin{align}
\label{eq_stationary_estimation}
    \mathbb{E}\{Q_{k}^{\circ}(\gamma)\} - \Sigma (\gamma) = - \sum_{|\ell| > k - \phi(k)} \mathrm{cov}\{\vartheta_{0}^{\circ}(\gamma), \vartheta_{\ell}^{\circ}(\gamma)\}. 
\end{align}
which reveals that $Q_{k}^{\circ}(\gamma)$ can have a larger bias for smaller value of $k - \phi(k)$. This motivates us to propose a new estimator by excluding these terms with small value of $k - \phi(k)$. More specifically, we define a sequence of positive integers $\{\tau(m)\}_{m \in \mathbb{N}}$ such that $\tau(m) \leq (\eta_{m + 1} - \eta_{m})$ for all $m \in \mathbb{N}$. Then, the bias reduced estimator of $V_{n}^{\circ}(\gamma)$ is defined by excluding all these $Q_{k}^{\circ}(\gamma)$ with $k - \phi(k)$ smaller than the threshold $\tau(\psi(k))$, that is, 
\begin{align*}
    U_{n}^{\circ}(\gamma) = \sum_{k = 1}^{n} Q_{k}^{\circ}(\gamma) \mathbb{I}\{k - \phi(k) \geq \tau(\psi(k))\}.
\end{align*}
This motivates us to propose the bias-reduced counterpart of $V_{n}(\gamma)$ as follows: 
\begin{align*}
    U_{n}(\gamma) = \sum_{k = 1}^{n} \Bigg((&\vartheta_{k}(\gamma) - \hat{\theta}_{n}(\gamma))^{\otimes 2} + \sum_{\ell = \phi(k)}^{k - 1} \{\vartheta_{k}(\gamma) - \hat{\theta}_{n}(\gamma)\}\{\vartheta_{\ell}(\gamma) - \hat{\theta}_{n}(\gamma)\}^{\top}\cr
    &+ \sum_{\ell = \phi(k)}^{k - 1} \{\vartheta_{\ell}(\gamma) - \hat{\theta}_{n}(\gamma)\}\{\vartheta_{k}(\gamma) - \hat{\theta}_{n}(\gamma)\}^{\top}\Bigg) \mathbb{I}\{k - \phi(k) \geq \tau(\psi(k))\}. 
\end{align*}
For simplicity of notation, we write $E_{k}(\tau) = \{k - \phi(k) \geq \tau(\psi(k))\}$ for each $k \in \mathbb{N}$. For each block $B_{m}$, we use $\mathcal{S}_{m}^{\tau}(\gamma) = \sum_{k = \eta_{m}}^{\eta_{m} + \tau(m) - 1} \vartheta_{k}(\gamma)$ to denote the part excluded for bias reduction from $\mathcal{S}_{m}(\gamma)$. To facilitate the recursive computation of $U_{n}(\gamma)$, we first rewrite it as 
\begin{align}
\label{eq_estimation_Vn_0}
    U_{n}(\gamma) = \mathcal{U}_{n}(\gamma) + \mathcal{K}_{n} \hat{\theta}_{n}(\gamma)^{\otimes 2} - \mathcal{H}_{n}(\gamma) \hat{\theta}_{n}(\gamma)^{\top} - \hat{\theta}_{n}(\gamma) \mathcal{H}_{n}(\gamma)^{\top}, 
\end{align}
where
\begin{align}
\label{eq_estimatino_Vn_00}
    \mathcal{U}_{n}(\gamma) &= \sum_{k = 1}^{n} \Bigg(\vartheta_{k}(\gamma)^{\otimes 2} + \sum_{\ell = \phi(k)}^{k - 1} \vartheta_{k}(\gamma) \vartheta_{\ell}(\gamma)^{\top} + \sum_{\ell = \phi(k)}^{k - 1} \vartheta_{\ell}(\gamma) \vartheta_{k}(\gamma)^{\top}\Bigg) \times \mathbb{I}\{E_{k}(\tau)\}, \cr  
    \mathcal{K}_{n} &= \sum_{m = 1}^{\psi(n) - 1} \{|B_{m}|^{2} - \tau^{2}(m)\} + \{|\delta_{\eta}(n)|^{2} - \tau^{2}(\psi(n))\} \times \mathbb{I}\{E_{n}(\tau)\},\cr 
    \mathcal{H}_{n}(\gamma) &= \sum_{m = 1}^{\psi(n) - 1} \{|B_{m}| \mathcal{S}_{m}(\gamma) - \tau(m) \mathcal{S}_{m}^{\tau}(\gamma)\} \cr
    & \quad + \{\delta_{\eta}(n) \mathcal{R}_{n}(\gamma) - \tau(\psi(n)) \mathcal{S}_{\psi(n)}^{\tau}(\gamma)\} \times \mathbb{I}\{E_{n}(\tau)\}. 
\end{align}
Given $\{\vartheta_{k}(\gamma)\}_{k \in \{1,...,n\}}$, our estimator for $\Sigma(\gamma)$ with bias reduction is then defined by 
\begin{align}
    \label{eq_recursive_longrun_est_debias}
    \tilde{\Sigma}_{n} (\gamma) = \frac{1}{\kappa(n)} U_{n}(\gamma), \enspace \mathrm{where} \enspace \kappa(n) = n - \sum_{m = 1}^{\psi(n) - 1} \tau(m) - \min\{n - \phi(n), \tau(\psi(n))\}.
\end{align}
Following the idea of Algorithm \ref{algo1}, our algorithm for recursively computing $\tilde\Sigma_n(\gamma)$ is summarized in Algorithm~\ref{algo2}.

\begin{algorithm}[hbt!]
   \caption{Recursive estimation of the bias-reduced long-run covariance matrix}
   \label{algo2}
   \footnotesize
   \KwData{Observations $X_1,X_2,\ldots,X_n$; objective function $g(\cdot)$; constant step size $\gamma$; predefined sequences $\{\eta_m\}_{m\in\NN}$ and $\{\tau(m)\}_{m\in\NN}$}
   \KwResult{Extrapolated ASGD estimator $\hat\theta_{n+1}(\gamma)$; estimated long-run covariance matrix $\tilde{\Sigma}_{n+1}(\gamma)$} 
   \Init{$\theta_0(a_{1} \gamma)=\theta_0(a_{2} \gamma)= \theta_0(a_{3} \gamma)=\hat{\theta}_0(\gamma)=\vartheta_0(\gamma)=\mathcal{R}_0(\gamma)=\mathcal{U}_0(\gamma)=\mathcal{S}_{\psi(0)}^{\tau}(\gamma)\gets0$,}{$\psi(0)\gets1$, $\phi(0)\gets\eta_1$, $\delta_{\eta}(0)\gets0$,} $\mathcal{K}_0=\mathcal{H}_0(\gamma)\gets0$, $\mathcal{V}_0(\gamma)=V_0(\gamma)\gets0$, $\kappa(0)\gets1$
   \For{$n=0,1,2,3,\ldots$}{
   $\theta_{n+1}(a_{1} \gamma)\gets\theta_n(a_{1}\gamma)-a_{1}\gamma \nabla g(\theta_n(a_{1}\gamma),X_{n+1})$\Comment*[r]{SGD estimate}
   $\theta_{n+1}(a_{2}\gamma)\gets\theta_n(a_{2}\gamma)-a_{2}\gamma \nabla g(\theta_n(a_{2}\gamma),X_{n+1})$\;
   $\theta_{n+1}(a_{3}\gamma)\gets\theta_n(a_{3}\gamma)-a_{3}\gamma \nabla g(\theta_n(a_{3}\gamma),X_{n+1})$\;
   $\vartheta_{n+1}(\gamma)\gets b_{1} \theta_{n + 1} (a_{1} \gamma) + b_{2} \theta_{n + 1} (a_{2} \gamma) + b_{3} \theta_{n + 1} (a_{3} \gamma)$\;
   $\hat{\theta}_{n + 1}(\gamma) \gets \{n\hat{\theta}_{n}(\gamma) + \vartheta_{n + 1}(\gamma)\}/(n + 1)$ \Comment*[r]{ASGD estimate}
   \eIf{$n + 1 < \eta_{\psi(n) + 1}$}{
   $\psi(n+1)\gets\psi(n)$, $\phi(n+1)\gets\phi(n)$, $\delta_{\eta}(n+1)\gets\delta_{\eta}(n)+1$\;
   $\mathcal{R}_{n + 1}(\gamma) \gets \mathcal{R}_{n}(\gamma) + \vartheta_{n + 1}(\gamma)$\;
   \eIf{$n + 1 \geq \phi(n + 1)+ \tau(\psi(n + 1))$}{
   \eIf{$n \geq \phi(n) + \tau(\psi(n))$}{
   $\mathcal{U}_{n + 1}(\gamma) \gets \mathcal{U}_{n}(\gamma) + \vartheta_{n + 1}(\gamma)^{\otimes 2} + \vartheta_{n + 1}(\gamma) \mathcal{R}_{n}(\gamma)^{\top} + \mathcal{R}_{n}(\gamma) \vartheta_{n + 1}(\gamma)^{\top}$\;
   $\mathcal{K}_{n + 1} \gets \mathcal{K}_{n} - \delta_{\eta}^{2}(n) + \delta_{\eta}^{2}(n + 1)$, $\mathcal{S}_{\psi(n + 1)}^{\tau}(\gamma) \gets \mathcal{S}_{\psi(n)}^{\tau}(\gamma)$\;
   $\mathcal{H}_{n + 1}(\gamma) \gets \mathcal{H}_{n}(\gamma) - \delta_{\eta}(n) \mathcal{R}_{n}(\gamma) + \delta_{\eta}(n + 1) \mathcal{R}_{n + 1}(\gamma)$\;
   }{
   $\mathcal{U}_{n + 1}(\gamma) \gets \mathcal{U}_{n}(\gamma) + \vartheta_{n + 1}(\gamma)^{\otimes 2} + \vartheta_{n + 1}(\gamma) \mathcal{R}_{n}(\gamma)^{\top} + \mathcal{R}_{n}(\gamma) \vartheta_{n + 1}(\gamma)^{\top}$\;
   $\mathcal{K}_{n + 1} \gets \mathcal{K}_{n} + \delta_{\eta}^{2}(n + 1) - \tau^{2}(\psi(n + 1))$\;
   $\mathcal{H}_{n + 1}(\gamma) \gets \mathcal{H}_{n}(\gamma) + \delta_{\eta}(n + 1)\mathcal{R}_{n + 1}(\gamma) - \tau(\psi(n + 1))\mathcal{S}_{\psi(n + 1)}^{\tau}(\gamma)$\;
   }
   }{
   $\mathcal{K}_{n + 1} \gets \mathcal{K}_{n}$, {$\mathcal{S}_{\psi(n + 1)}^{\tau}(\gamma) \gets \mathcal{S}_{\psi(n)}^{\tau}(\gamma) + \vartheta_{n+1}(\gamma)$}\;
   $\mathcal{U}_{n + 1}(\gamma) \gets \mathcal{U}_{n}(\gamma)$, $\mathcal{H}_{n + 1} \gets \mathcal{H}_{n}$\;
   }
   }{
   $\psi(n + 1) \gets \psi(n) + 1$, $\phi(n + 1) = n + 1$, $\delta_{\eta}(n + 1) = 1$\;
   $\mathcal{R}_{n + 1}(\gamma) \gets \vartheta_{n + 1}(\gamma)$\; 
   \eIf{$\tau(\psi(n+1))=0$}{$\mathcal S_{\psi(n+1)}^\tau(\gamma)\gets0$, $\mathcal U_{n+1}(\gamma) \gets\mathcal U_n(\gamma)+\vartheta_{n+1}(\gamma)^{\otimes2}$, $\mathcal K_{n+1}\gets\mathcal K_n+1$, $\mathcal H_{n+1}(\gamma)\gets\mathcal H_n(\gamma)+\vartheta_{n+1}(\gamma)$\;
    }{
    $\mathcal S_{\psi(n+1)}^\tau(\gamma)\gets\vartheta_{n+1}(\gamma)$, $\mathcal U_{n+1}(\gamma)\gets\mathcal U_n(\gamma)$, $\mathcal K_{n+1}\gets\mathcal K_n$, $\mathcal H_{n+1}(\gamma)\gets\mathcal H_n(\gamma)$\;
    }
   }
   $U_{n + 1}(\gamma) \gets \mathcal{U}_{n + 1}(\gamma) + \mathcal{K}_{n + 1} \hat{\theta}_{n + 1}(\gamma)^{\otimes 2} - \mathcal{H}_{n + 1}(\gamma) \hat{\theta}_{n + 1}(\gamma)^{\top} - \hat{\theta}_{n + 1}(\gamma) \mathcal{H}_{n + 1}(\gamma)^{\top}$\;
   $\kappa(n+1) \gets n+1 - \sum_{m = 1}^{\psi(n+1) - 1} \tau(m) - \min\{n+1 - \phi(n+1), \tau(\psi(n+1))\}$\;
   $\tilde{\Sigma}_{n+1}(\gamma) \gets U_{n+1}(\gamma)/\kappa(n+1)$\Comment*[r]{Recursive long-run covariance matrix estimate with bias reduction}
   }
\end{algorithm}

\section{Convergence of SGD with Infinite Variances}\label{subsec_heavy_tail}

\subsection{Quenched \texorpdfstring{$\alpha$}{alpha}-Stable Convergence}
In Sections~\ref{subsec_clt}~and~\ref{subsec_invar_prin}, we assume that the stochastic gradient has a finite variance at the global optimum $\theta^*$ and derive quenched asymptotic normality. In this section, we consider the SGD iterates $\{\theta_k(\gamma)\}_{k\in\NN}$ with heavy-tailed gradient noises and provide a quenched $\alpha$-stable convergence of the ASGD sequence $\{\bar\theta_n(\gamma)\}_{n\in\NN}$ when the condition on the regularly varying tail (cf. Assumption \ref{asm_regular_vary_tail} below) holds. When recursion (\ref{eq_SGD}) can be written into an affine iterated random function (\cite{gurbuzbalaban_heavy-tail_2021, mirek_heavy_2011}) or an additive functional of linear processes (\cite{wu_additive_2003}), the $\alpha$-stable convergence can be proved. In this case we can have quenched $\alpha$-stable convergence; see Theorem \ref{thm_alpha_stable} below. However, for a general nonlinear SGD algorithm with infinite variance noises, the limiting distribution remains unknown. To bridge this gap, we develop a robust quenched version for the constant step-size SGD algorithms (cf. Corollary~\ref{cor_gmc_proj}), which asserts geometric moment contraction of projected SGD so that Theorems~\ref{thm2_quenched_CLT} and~\ref{thm_invar_prin} are applicable.

Recall the definition of the stacked SGD sequence $\{\varepsilon_{k}^{\circ}\}_{k}$ in~\eqref{eq_clt_partial_sum}. Following~\textcite{Basrak2009regularly}, we now introduce the regularly varying tail assumption of $\{\varepsilon_{k}^{\circ}\}_{k}$. 

\begin{assumption}[Regularly varying tails]
\label{asm_regular_vary_tail}
The function $x \mapsto \mathbb{P}(|\varepsilon_{0}^{\circ}| > x)$ is regularly varying of index $-\alpha$ for some $\alpha \in (1, 2)$, that is, $\PP(|\varepsilon_0^\circ|>x)=x^{-\alpha}L(x)$, as $x\to\infty$
for some slowly varying function $L$, i.e., $\lim_{x\to\infty}L(cx)/L(x)=1$, $c>0$.
The sequence of stationary process $\{\varepsilon_{t}^{\circ}\}_{t \in \mathbb{Z}}$ is jointly regularly varying with index $\alpha$. That is, there exist a random variable $Y$ with $\mathbb{P}(Y > y) = y^{-\alpha}$ for $y \geq 1$, and a stochastic process $\{\Xi_{t}\}_{t \in \mathbb{Z}}$ in $\mathbb{R}^{d \ell}$ (called the spectral tail process), independent of $Y$, such that for every $h \geq 0$,  
    \begin{align}
    \label{eq_joint_regular_vary}
        \mathbb{P} \left\{\left(\frac{\varepsilon_{-h}^{\circ}}{x},\ldots, \frac{\varepsilon_{h}^{\circ}}{x}\right) \in \cdot | |\varepsilon_{0}^{\circ}| > x\right\} \to \mathbb{P} \{(Y \Xi_{-h}, \ldots, Y \Xi_{h}) \in \cdot\}, \,\, \mathrm{as} \enspace x \to \infty. 
    \end{align}
\end{assumption}

Let $\{a_{n}\}_{n \geq 1}$ be a sequence of positive numbers such that $n \mathbb{P}(|\varepsilon_{0}^{\circ}| > a_{n}) \to 1$, as $n \to \infty$. For example, $a_{n}$ can be the $(1 - 1/n)$-quantile of $|\varepsilon_{0}^{\circ}|$. The following theorem establishes the quenched asymptotic distribution of $S_{n} = \sum_{k = 1}^{n} \varepsilon_{k}$. 

\begin{theorem}[Quenched $\alpha$-stable convergence]
\label{thm_alpha_stable}
    Suppose that Assumptions~\ref{asm_conv}~and~\ref{asm_Lip} hold with some $p \in (1, \alpha)$ and that Assumption~\ref{asm_regular_vary_tail} holds. Let $\gamma_j\in(0,\gamma(p))$ for $j=1,\ldots,\ell$. Then, as $n \to \infty$, we have the weak convergence $S_{n}/a_{n} \Rightarrow \mathcal{S}(\alpha)$ for an arbitrary initial point $\mathrm{vec}(\theta_0(\gamma_1),\ldots,\theta_0(\gamma_{\ell}))$. Here $\mathcal{S}(\alpha)$ stands for the $\alpha$-stable distribution with characteristic function 
    $$\varphi_{\alpha}(\nu) = \exp\big\{- c_{\alpha} \sigma^{\alpha} (\nu) \big(1 - i \beta(\nu) \tan(\alpha \pi /2)\big)\big\}, \quad \nu \in \mathbb{R}^{d \ell},$$ 
    where $c_{\alpha} = \Gamma (2 - \alpha) \cos (\alpha \pi/2) /(1 - \alpha)$, 
    \begin{align*}
        \sigma^{\alpha}(\nu) = \mathbb{E}\Big[\frac{ \left|\sum_{t \in \mathbb{Z}} \nu^{\top} \Xi_{t}\right|^{\alpha}}{\sum_{t\in\ZZ}|\Xi_t|^{\alpha}}\Big], 
        \quad
        \beta(\nu) = \frac{\mathbb{E} \big[\big((\sum_{t \in \mathbb{Z}} \nu^{\top} \Xi_{t})_{+}^{\alpha} - (\sum_{t \in \mathbb{Z}} \nu^{\top} \Xi_{t})_{-}^{\alpha}\big)/\sum_{t\in\ZZ}|\Xi_t|^{\alpha}\big]}{\mathbb{E} \big[|\sum_{t \in \mathbb{Z}} \nu^{\top} \Xi_{t}|^{\alpha}/\sum_{t\in\ZZ}|\Xi_t|^{\alpha}\big]},
    \end{align*}
    and $\beta(\nu)=0$ if $\sigma^{\alpha}(\nu)=0$.
\end{theorem}

\subsection{Projected SGD}
The regularly varying tail Assumption \ref{asm_regular_vary_tail} can be quite difficult to verify. Following the idea of robust statistics, we can use truncation by applying projected SGD to handle the setting with heavy-tailed gradient noises. Our approach is similar to the widely used gradient clipping and mirror descent methods (\cite{zhang_why_2019,vural_mirror_2022}). Specifically, we consider the SGD iteration
\begin{align}
    \label{eq_proj_sgd}
    \theta_k^{\diamondsuit}(\gamma) = \Phi_{\Theta^{\star}}\big(\theta_{k-1}^{\diamondsuit}(\gamma) - \gamma\nabla g(\theta_{k-1}^{\diamondsuit}(\gamma), X_k)\big) =: F_{X_k}^{\mathrm{Proj}}(\theta_{k-1}^{\diamondsuit}),
\end{align}
where $\Phi_{\Theta^\star}(\cdot)=\arg\min_{v\in \Theta^\star}|\cdot-v|$ is a projection operator onto the set $\Theta^\star\subset\RR^d$ with $\theta^*\in\Theta^{\star}$. Such type of projection is widely adopted in the SGD literature, especially in the studies on generalization bounds (\cite{lei_fine-grained_2020,park_generalization_2022,sachs_generalization_2023}). By mapping the $k$-th SGD iterate $\theta_k(\gamma)$ onto some set $\Theta^\star$, this projected SGD is robust to heavy-tailed noises because the projection step limits the impact of extreme updates by constraining the iterates to a feasible set $\Theta^{\star}$ (e.g., bounded or convex set). The GMC property of the iterated random function $F_{X}(\theta)$ in Theorem~\ref{thm_GMC} similarly holds for this projected version $F_{X}^{\mathrm{Proj}}(\theta)$, and we only need Assumptions~\ref{asm_conv}~and~\ref{asm_Lip} to hold on the set $\Theta^{\star}$ instead of $\RR^d$.

\begin{corollary}[Geometric moment contraction of projected SGD]
\label{cor_gmc_proj}
Suppose that Assumptions~\ref{asm_conv}~and~\ref{asm_Lip} hold for some $p>1$ on a closed and convex set $\Theta^{\star}\subset\RR^d$ with $\theta^*\in\Theta^{\star}$ and $\gamma\in(0,\gamma(p))$. Then, the geometric-moment contraction in Theorem~\ref{thm_GMC} holds for $F_X^{\mathrm{Proj}}(\theta)$ and $\{\theta_k^{\diamondsuit}(\gamma)\}_{k\in\NN}$, with the stationary distribution $\pi_{\gamma}$ therein replaced by $\pi_{\gamma,\Theta^{\star}}$.

\end{corollary}
Corollary~\ref{cor_gmc_proj} relies on the standard result that Euclidean projections are non-expansive, which have been proved in the literature; see for example Theorem 6.42 of \textcite{beck_first-order_2017} and Proposition 7.7 of \textcite{wright_optimization_2022}. 
Suppose additionally that $\Theta^\star$ is bounded. If Assumptions~\ref{asm_conv}~and~\ref{asm_Lip}(ii) hold on $\Theta^\star$ with $p=2$, then the projected iterates have finite second moments, and the argument of Theorem~\ref{thm2_quenched_CLT} applies with $\pi_\gamma$ and $\theta_\infty^*(\gamma)$ replaced by $\pi_{\gamma,\Theta^\star}$ and
\begin{align*}
    \theta_{\infty,\Theta^\star}^*(\gamma)
    =\int_{\Theta^\star}y\,\pi_{\gamma,\Theta^\star}(dy),
\end{align*}
respectively. Under the corresponding assumptions with $p>2$, the invariance principle also applies. Notably, these limits concern the projected stationary mean
$\theta_{\infty,\Theta^\star}^*(\gamma)$, which may differ from the population minimizer $\theta^*$.

\subsection{Proof of Theorem~\ref{thm_alpha_stable}}
\begin{proof}[Proof of Theorem~\ref{thm_alpha_stable}] 
    As discussed in~\textcite{wu_limit_2004}, by H\"{o}lder's inequality, it follows that for any $0 < p^{\dagger} \leq 1$, there exists a constant $\rho^{\dagger} \in (0, 1)$ such that for $0 < \gamma < \gamma(p)$, 
    \begin{align*}
        \|F_{X}(\theta) - F_{X}(\theta')\|_{p^{\dagger}} \leq \rho^{\dagger} |\theta - \theta'|, \quad \theta, \theta' \in \mathbb{R}^{d}. 
    \end{align*}
    Hence, the contractive condition (5.7) in Proposition 5.2 of~\textcite{Matsui2023self} is satisfied. This together with Assumptions~\ref{asm_Lip}~and~\ref{asm_regular_vary_tail}, we can obtain $S_{n}^{\circ}/a_{n} \Rightarrow \mathcal{S}(\alpha)$ following from a similar argument to that in the proof of Proposition 5.2 in~\textcite{Matsui2023self}. Consequently, for any initial point $\mathrm{vec}(\theta_{0}(\gamma_{1}), \ldots, \theta_{0}(\gamma_{\ell}))$, $S_{n}/a_{n} \Rightarrow \mathcal{S}(\alpha)$ holds in view of $a_{n} \to \infty$ and 
    \begin{align*}
        \frac{\|S_{n} - S_{n}^{\circ}\|_{p}}{a_{n}} &\leq \frac{\sum_{k = 1}^{n} \sum_{j = 1}^{\ell} \|\theta_{k}(\gamma_{j}) - \theta_{k}^{\circ}(\gamma_{j})\|_{p}}{a_{n}} \cr 
        &\leq \frac{\sum_{k = 1}^{n} \sum_{j = 1}^{\ell} \rho_{\gamma_{j}, p}^{k} \|\theta_{0}(\gamma_{j}) - \theta_{0}^{\circ}(\gamma_{j})\|_{p}}{a_{n}} \cr 
        &\le \frac{\sum_{j = 1}^{\ell} \|\theta_{0}(\gamma_{j}) - \theta_{0}^{\circ}(\gamma_{j})\|_{p} / (1- \rho_{\gamma_{j}, p}) }{a_{n}} \to 0, \quad \mathrm{as} \enspace n \to \infty. 
    \end{align*}
\end{proof}

\section{Postponed Proofs}\label{sec_proof}

\subsection{Proofs of Theorems~\ref{thm_GMC}~\&~\ref{thm_asgd_converge}}

The proof of Theorem \ref{thm_GMC} is essentially motivated by Lemma~\ref{lemma_power_ineq} below, which itself can be of independent interests in the bounds of the power functions $|\cdot|^p$, for $p>1$. 
\begin{lemma}\label{lemma_power_ineq}
    For any two vectors $x,y\in\RR^d$, the following two inequalities hold:

    (i) for $p\ge2$, $||x+y|^p - |x|^p - p|x|^{p-2}x^{\top}y| \le (|x|+|y|)^p -|x|^p-p|x|^{p-1}|y|$;

    (ii) for $1<p<2$, $||x+y|^p - |x|^p - p|x|^{p-2}x^{\top}y| \le 2^{2-p} |y|^{p}$.
\end{lemma}
The proof of this lemma and its application to Theorem \ref{thm_GMC} are provided in the Proof of Theorem~\ref{thm_GMC} below.

\begin{proof}[Proof of Theorem~\ref{thm_GMC}]
We first show the contraction in (\ref{eq_contraction}) for two cases with $p\ge2$ and $1<p<2$ respectively.

(i) Case $p\geq 2$:

Let $x=\theta-\theta'$, $y=-\gamma[\nabla g(\theta,X)-\nabla g(\theta',X)]$. Since the contraction in (\ref{eq_contraction}) holds when $|\theta-\theta'|=0$, w.l.o.g., we assume that $|x|>0$. Let $\theta=\theta'+\delta v$, where $v$ is a unit vector in $\RR^d$ and $\delta>0$. This gives us $x=\delta v$ and $|x|=|\theta-\theta'|=\delta$.

We first show the following analytical inequality:
\begin{align}
    \label{thm1_pgtr2_inq}
    \big||x+y|^p - |x|^p - p|x|^{p-2}x^{\top}y\big| \le \big(|x|+|y|\big)^p -|x|^p-p|x|^{p-1}|y|.
\end{align}
Let $y=a\delta v+be$, where $e$ is a unit vector in $\RR^d$ which is orthogonal to $v$, and $a,b$ are some constants in $\RR$. We define $R^2=(a\delta)^2+b^2$ and thus $|y|=R$. Then, since $x^{\top}y=a\delta^2$, we have 
\begin{align}
    \big||x+y|^p - |x|^p - p|x|^{p-2}x^{\top}y\big| & = \big|[(1+a)^2\delta^2+b^2]^{p/2} - \delta^p - pa\delta^p\big| \nonumber \\
    & = \big|(\delta^2 + 2a\delta^2 +R^2)^{p/2} - \delta^p - pa\delta^p\big|.
\end{align}
Let $a\delta=Ru$ with some constant $|u|\le1$. Then,
\begin{align}
    (\delta^2 + 2a\delta^2 +R^2)^{p/2} - \delta^p - pa\delta^p = (\delta^2 + 2\delta Ru + R^2)^{p/2} - \delta^p - p\delta^{p-1}Ru =: f(u).
\end{align}
By taking the derivative of the function $f(u)$, we obtain
\begin{align}
    \frac{d}{du}f(u) & = \frac{p}{2}(\delta^2+2\delta Ru+R^2)^{p/2-1}\cdot 2 \delta R - p\delta^{p-1}R \nonumber \\
    & = p\delta R\big[(\delta^2+2\delta Ru+R^2)^{p/2-1}-\delta^{p-2}\big],
\end{align}
which indicates that $f(u)$ is decreasing in $[-1,-R/(2\delta)]$ and increasing in $(-R/(2\delta),1]$. 
Set $g(R)=f(1)-f(-1)=(\delta+R)^p-|\delta-R|^p -2p\delta^{p-1}R$. Then $g(0)=0$. For $R\le\delta$, $g'(R)\ge0$ by convexity of $t^{p-1}$. For $R>\delta$, $g'(R)=p\{(R+\delta)^{p-1}-(R-\delta)^{p-1}-2\delta^{p-1}\}\ge0$. Thus $f(1)\ge f(-1)$, which indicates that for $p\ge2$,
\begin{align}
    \max_{|u|\le1}|f(u)| = f(1) = (\delta+R)^p - \delta^p- p\delta^{p-1}R.
\end{align}
This, along with the fact that $|x|=\delta$ and $|y|=R$ leads to the inequality in (\ref{thm1_pgtr2_inq}).

Further, we take the expectation on the both sides of the inequality (\ref{thm1_pgtr2_inq}) and achieve
\begin{align}
    \label{eq_thm1_expectation}
    \EE|x+y|^p - |x|^p- p|x|^{p-2}\EE(x^{\top}y) \le \EE(|x|+|y|)^p -|x|^p-p|x|^{p-1}\EE|y|.
\end{align}
Recall that $x=\theta-\theta'$, $y=-\gamma[\nabla g(\theta,X)-\nabla g(\theta',X)]$. We aim to bound $\EE|x+y|^p$, which can be decomposed into two parts as follows:
\begin{align*}
    \EE|x+y|^p & = \big[\EE|x+y|^p - |x|^p- p|x|^{p-2}\EE(x^{\top}y)\big] + \big[|x|^p+ p|x|^{p-2}\EE(x^{\top}y)\big] \nonumber \\
    & =: \mathbb{I}_1 + \mathbb{I}_2.
\end{align*}
For the part $\mathbb{I}_2$, it follows from Assumption \ref{asm_conv} that 
\begin{align}
    \label{eq_thm1_conv}
    \EE(x^{\top}y) = -\gamma\langle \theta-\theta',m(\theta)-m(\theta')\rangle \le -\mu\gamma|\theta-\theta'|^2.
\end{align}
Next, for the part $\mathbb{I}_1$, we note that, for any $p>1$ ($p$ is not necessarily an integer), we can write the right-hand side of (\ref{eq_thm1_expectation}) into
\begin{align*}
    \EE(|x|+|y|)^p-|x|^p - p|x|^{p-1}\EE|y|
    & = \int_{s=0}^1\int_{t=0}^sp(p-1)\EE\big[(|x|+|y|t)^{p-2}|y|^2\big]dtds.
\end{align*}
By Hölder's inequality and the triangle inequality, we have
\begin{align*}
    \EE\big[(|x|+|y|t)^{p-2}|y|^2\big] & \le \big(\EE(|x|+|y|t)^p\big)^{(p-2)/p} \big(\EE|y|^p\big)^{2/p} \nonumber \\
    & \le (|x| + \|y\|_pt)^{p-2}\|y\|_p^2.
\end{align*}
Furthermore, for each $p>1$, it follows from Assumption \ref{asm_Lip} that
\begin{align*}
    \|y\|_p = \big\|-\gamma[\nabla g(\theta,X) - \nabla g(\theta',X)]\big\|_p \le \gamma L_p|\theta-\theta'| = \gamma L_p|x|.
\end{align*}
Hence, we can achieve 
\begin{align}
    \label{eq_thm1_Lip}
    \EE(|x|+|y|)^p -|x|^p-p|x|^{p-1}\EE|y| \le (1+\gamma L_p)^p|x|^p - |x|^p - p\gamma L_p|x|^p.
\end{align}

Finally, by combining the results for $\mathbb{I}_1$ and $\mathbb{I}_2$, we can obtain
\begin{align}
    & \quad \EE|\theta-\theta' - \gamma(\nabla g(\theta,X) - \nabla g(\theta',X))|^p \nonumber \\
    & \le \big[1-p\mu\gamma + (1+\gamma L_p)^p - 1 - p\gamma L_p\big]|\theta-\theta'|^p.
\end{align}
As a direct consequence, the upper bound for $\gamma$ is the solution to the following equation:
\begin{align}
    (1+\gamma L_p)^p - 1 - p\gamma(\mu+L_p)=0.
\end{align}

(ii) Case $1<p<2$:

First, for real $d$-dimensional vectors $x,y$ the analytic inequality
    \begin{equation}
        |x + y|^{p} \le |x|^{p} + p |x|^{p-2} x^{\top} y + 2^{2-p} |y|^{p}\label{lemma1_qoal2}
    \end{equation}
    is shown. For $x=0$ or $y=0$ this inequality holds obviously. For $|x|, |y| > 0$, define  
    \begin{align*}
        \omega = \frac{|x|}{|y|} > 0 \enspace \mathrm{and} \enspace \varrho = \frac{x^{\top} y}{|x| \cdot |y|}. 
    \end{align*}
    Then $|\varrho| \leq 1$ and 
    \begin{align*}
        \frac{|x + y|^{p} - |x|^{p} - p |x|^{p-2} x^{\top} y}{|y|^{p}} = (\omega^{2} + 1 + 2 \omega \varrho)^{p/2} - \omega^{p} - p \omega^{p - 1} \varrho =: \psi(\omega, \varrho). 
    \end{align*}
    Note that 
    \begin{align*}
        \frac{\partial}{\partial \varrho} \psi(\omega, \varrho) = p \omega^{p - 1} \left\{\left(1 + \frac{2 \varrho}{\omega} + \frac{1}{\omega^{2}}\right)^{p/2 - 1} - 1\right\}. 
    \end{align*}
    If $\omega \geq 1/2$, 
    \begin{align*}
        \sup_{\varrho \in [-1, 1]} \psi(\omega, \varrho) = \psi\left(\omega, -\frac{1}{2\omega}\right) = \frac{p \omega^{p - 2}}{2} \leq p 2^{1 - p} \leq 2^{2 - p}. 
    \end{align*}
If $0 < \omega < 1/2$, 
by Proposition 1.8 in~\textcite{pinelis_best_2015}, it follows that 
    \begin{align*}
        \sup_{\varrho \in [-1, 1]} \psi(\omega, \varrho) = \psi(\omega, -1) = (1 - \omega)^{p} - \omega^{p} + p \omega^{p - 1}  \leq 2^{2 - p}. 
    \end{align*}
    In total, we have that for all $\rho \in [-1,1]$, $\omega > 0$ that
    \[
        \psi(\omega,\rho) \le 2^{2-p},
    \]
    leading to
    \[
        \frac{|x + y|^{p} - |x|^{p} - p |x|^{p-2} x^{\top} y}{|y|^{p}} \le 2^{2-p}
    \]
    and thus to (\ref{lemma1_qoal2}). Next, recall that $x=\theta-\theta'$, $y=-\gamma[\nabla g(\theta,X)-\nabla g(\theta',X)]$. By taking the expectation on the both sides of inequality (\ref{lemma1_qoal2}), we have
\begin{align}
    & \quad \EE|\theta-\theta' - \gamma(\nabla g(\theta,X) - \nabla g(\theta',X))|^p \nonumber \\
    & \le |\theta-\theta'|^p - p|\theta-\theta'|^{p-2}(\theta-\theta')^{\top}\EE\big[\gamma\big(\nabla g(\theta,X) - \nabla g(\theta',X)\big)\big] \nonumber \\
    &\quad + 2^{2-p}\EE\big|\gamma\big(\nabla g(\theta,X) - \nabla g(\theta',X)\big)\big|^p \nonumber \\
    & \le \big(1-p\mu\gamma + 2^{2-p}L_p^p\gamma^p\big)|\theta-\theta'|^p.
\end{align}
This completes the proof of the geometric contraction condition.

Next, we shall show the existence and uniqueness of the stationary distribution $\pi_{\gamma}$ based on the GMC in (\ref{eq_GMC}) by using Theorem 2 in \textcite{wu_limit_2004}. To this end, recall the recursive function $F_X(\theta)$ defined in (\ref{eq_F_recursive}) and the global optimum $\theta^*$ defined in (\ref{eq_opt_goal}). Note that, for $p>1$, it follows from Assumption \ref{asm_Lip}(i) that
\begin{align}
    \label{eq_pi_cond1}
    \EE|\theta^* - F_X(\theta^*)|^p = \EE|\gamma\nabla g(\theta^*,X)|^p <\infty.
\end{align}
In addition, note that we have shown for any $p>1$ and $\rho_{\gamma,p}\in(0,1)$,
\begin{align}
    \label{eq_pi_cond2}
    \EE|F_{X_n}\circ\cdots\circ F_{X_1}(\theta) - F_{X_n}\circ\cdots\circ F_{X_1}(\theta_0)|^p \le (\rho_{\gamma,p}^p)^n|\theta-\theta_0|^p
\end{align}
holds for all $\theta,\theta_0\in\RR^d$, $n\in\NN$. In fact, expressions (\ref{eq_pi_cond1}) and (\ref{eq_pi_cond2}) are also valid for all $p\le1$ by Hölder's inequality. To see this, we let $p'\in(0,p)$ and $\rho_{\gamma,p'}=\rho_{\gamma,p}^{p'/p}\in(0,1)$. Then, we have
\begin{align}
    & \quad \EE|F_{X_n}\circ\cdots\circ F_{X_1}(\theta) - F_{X_n}\circ\cdots\circ F_{X_1}(\theta_0)|^{p'} \nonumber \\
    & \le \big(\EE|F_{X_n}\circ\cdots\circ F_{X_1}(\theta) - F_{X_n}\circ\cdots\circ F_{X_1}(\theta_0)|^p \big)^{p'/p} \nonumber \\
    & \le \big[(\rho_{\gamma,p}^p)^n|\theta-\theta_0|^p\big]^{p'/p} = (\rho_{\gamma,p'}^p)^n|\theta-\theta_0|^{p'}.
\end{align}
Therefore, both expressions (\ref{eq_pi_cond1}) and (\ref{eq_pi_cond2}) hold for all $p>0$, which indicates that Conditions 1 and 2 in \textcite{wu_limit_2004} are satisfied for our recursive function $F_X(\theta)$.

Denote the forward iteration process by 
\begin{equation}
    \label{forward} 
   A_n(\theta) = F_{X_n} \circ \cdots \circ F_{X_1}(\theta) 
\end{equation}
and the backward iteration process (\cite{propp1996exact})
\begin{equation} \label{backward} 
    B_n(\theta)= F_{X_0} \circ F_{X_{-1}} \cdots \circ F_{X_{1-n}}(\theta).
\end{equation}
Note that for all $\theta\in\RR^d$, we have the distributional equality $A_n(\theta) \stackrel{\mathcal{D}}{=}B_n(\theta)$. By Theorem 2 in \textcite{wu_limit_2004}, there exists a random variable $B_{\infty}$ which is $\sigma(X_0, X_{-1}, \ldots)$ measurable such that for all $\theta\in\RR^d$, $B_n(\theta)\rightarrow B_{\infty}$ almost surely as $n\rightarrow\infty$. Thus, $A_n(\theta)$ converges to $B_{\infty}$ in distribution. Then, as shown by Remark 1 in \textcite{wu_limit_2004}, Theorem 1 in \textcite{diaconis_freedman_1999} is also implied. Hence, there exists a unique stationary distribution $\pi$ with $B_{\infty}\sim\pi$, which does not depend $\theta$. 

Note that $\theta_n(\gamma)=A_n(\theta_0)$. Let $B_{\infty} = \theta_0^{\circ}(\gamma)$. Then, there exists a unique stationary distribution $\pi_{\gamma}$ such that $\theta_0^{\circ}(\gamma)\sim\pi_{\gamma}$ and therefore, $\theta_k(\gamma)\Rightarrow\pi_{\gamma}$. As a direct consequence of (\ref{eq_GMC}), $\pi_{\gamma}$ has finite $p$-th moment. Moreover, by applying the induction to (\ref{eq_GMC}), we can obtain the geometric moment contraction in (\ref{21}).


Finally, we shall show $\|\theta_n^{\circ}(\gamma)-\theta^*\|_p^2=O(\gamma)$ as $\gamma\rightarrow0$. For simplicity of notation, we write $\theta_n^{\circ}=\theta_n^{\circ}(\gamma)$ for $n \geq 1$. Since $\EE[\nabla g(\theta_{n-1}^{\circ},X_n)-m(\theta_{n-1}^{\circ})]=0$, by Rio's inequality (\cite{rio_moment_2009}), we have
\begin{align}
    \label{eq_apply_rio}
    \lVert \theta_n^{\circ}-\theta^*\rVert_p^2
    &\leq \lVert \theta_{n-1}^{\circ}-\theta^*-\gamma m(\theta_{n-1}^{\circ})\rVert_p^2+(p-1)\gamma^2\lVert\nabla g(\theta_{n-1}^{\circ},X_n)-m(\theta_{n-1}^{\circ})\rVert_p^2.
\end{align}
For the second term in (\ref{eq_apply_rio}), we apply Minkowski's inequality, which gives
\begin{align*}
    & \quad \lVert\nabla g(\theta_{n-1}^{\circ},X_n)-m(\theta_{n-1}^{\circ})\rVert_p \\
    &\leq\lVert\nabla g(\theta_{n-1}^{\circ},X_n)-\nabla g(\theta^*,X_n)\rVert_p+\lVert m(\theta_{n-1}^{\circ})-m(\theta^*)\rVert_p+\lVert \nabla g(\theta^*,X_n)\rVert_p.
\end{align*}
Thus, by Assumption \ref{asm_Lip}, we have 
\begin{align*}
    \lVert\nabla g(\theta_{n-1}^{\circ},X_n)-m(\theta_{n-1}^{\circ})\rVert_p^2
    &\leq 6L_p^2\lVert \theta_{n-1}^{\circ}-\theta^*\rVert_p^2+3\lVert \nabla g(\theta^*,X_n)\rVert_p^2.
\end{align*}
Regarding the first term in (\ref{eq_apply_rio}), since $m(\theta^*)=0$, we notice that 
\begin{align*}
    & \quad \lVert \theta_{n-1}^{\circ}-\theta^*-\gamma m(\theta_{n-1}^{\circ})\rVert_p^2 \\
    &= 
    \lVert\langle \theta_{n-1}^{\circ}-\theta^*,\theta_{n-1}^{\circ}-\theta^*\rangle-2\gamma\langle \theta_{n-1}^{\circ}-\theta^*,m(\theta_{n-1}^{\circ})-m(\theta^*)\rangle\\
    &\;\;\;+\gamma^2\langle m(\theta_{n-1}^{\circ}) - m(\theta^*),m(\theta_{n-1}^{\circ}) - m(\theta^*)\rangle\rVert_{p/2} \\
    & \le \lVert\langle \theta_{n-1}^{\circ}-\theta^*,\theta_{n-1}^{\circ}-\theta^*\rangle-2\gamma\langle \theta_{n-1}^{\circ}-\theta^*,m(\theta_{n-1}^{\circ})-m(\theta^*)\rangle\|_{p/2} + \gamma^2\|m(\theta_{n-1}^{\circ}) - m(\theta^*)\|_p^2.
\end{align*}
Then, by Minkowski's inequality, Jensen's inequality, and Assumptions \ref{asm_conv}~and~\ref{asm_Lip}, we have
\begin{align*}
    \|\theta_{n-1}^{\circ}-\theta^*-\gamma m(\theta_{n-1}^{\circ})\|^2_p
    &\le (1-2\gamma\mu+\gamma^2L_p^2)\|\theta_{n-1}^{\circ}-\theta^*\|^2_p.
\end{align*}
After combining the results above, we obtain 
\begin{align*}
\lVert \theta_n^{\circ}-\theta^*\rVert_p^2
    &\leq (1-2\gamma\mu+(6p-5)\gamma^2L_p^2)\lVert \theta_{n-1}^{\circ}-\theta^*\rVert_p^2+3(p-1)\gamma^2\lVert \nabla g(\theta^*,X_n)\rVert_p^2.
\end{align*}
By Assumption~\ref{asm_Lip},
\[
3(p - 1) \gamma^2 \lVert \nabla g(\theta^*, X_n) \rVert_p^2 = O(\gamma^2),
\]
and for sufficiently small $\gamma>0$,
\[
r_p(\gamma) = 1 - 2\gamma \mu + (6p - 5) \gamma^2 L_p^2 <1,
\]
we can express the recursive relationship as
\[
\lVert \theta_n^{\circ} - \theta^* \rVert_p^2 \leq r_p(\gamma) \lVert \theta_{n-1}^{\circ} - \theta^* \rVert_p^2 + O(\gamma^2).
\]
By iterating this inequality, we can conclude that
\begin{align*} 
    \lVert \theta_n^{\circ} - \theta^* \rVert_p^2 = O(\gamma^2)\sum_{j=0}^{\infty}(r_p(\gamma))^j = \frac{1}{1-r_p(\gamma)}O(\gamma^2) = O(\gamma),
\end{align*}
which completes the proof.
\end{proof}

\begin{proof}[Proof of Theorem~\ref{thm_asgd_converge}]
Let $\theta_{k,\{0\}}^\circ(\gamma)$ denote the stationary iterate obtained by replacing $X_0$ with an independent copy. By GMC in Theorem~\ref{thm_GMC},
\begin{align*}
    \sum_{k=0}^{\infty}\|\theta_k^\circ(\gamma)-\theta_{k,\{0\}}^\circ(\gamma)\|_p\leq \sum_{k=0}^{\infty}\rho^k\|\theta_0^\circ(\gamma)-\theta_{0,\{0\}}^\circ(\gamma)\|_p<\infty.
\end{align*}
Hence, Theorem~1(iii) of \textcite{wu_strong_2007} gives
\begin{align*}
    \Big\|\sum_{k=1}^n[\theta_k^\circ(\gamma)-\theta_\infty^*(\gamma)]\Big\|_p=O\big(n^{1/(p\wedge2)}\big).
\end{align*}
Moreover, by~\eqref{21}, $\|\sum_{k=1}^n[\theta_k(\gamma)-\theta_k^\circ(\gamma)]\|_p
=O(1)$. Dividing by $n$ proves the stated moment bound.


For the almost sure bounds, GMC verifies the conditions of Corollary~2(iii) of \textcite{wu_strong_2007} when $1<p<2$, Theorem~2(ii) when $p=2$, and Theorem~2(i) when $p>2$. These results first apply to the stationary sequence. By~\eqref{21}, $\sum_{k=1}^{\infty}|\theta_k(\gamma)-\theta_k^\circ(\gamma)|<\infty$ almost surely, so the same rates hold for 
$\bar\theta_n(\gamma)$.

\end{proof}

\subsection{Proof of Theorem~\ref{thm2_quenched_CLT}}

\begin{proof}[Proof of Theorem~\ref{thm2_quenched_CLT}]
Let $\{\theta_k^{\circ}(\gamma)\}_{k\in\NN}$ be the sequence of solutions from the SGD recursive function (\ref{eq_SGD}) with the stationary distribution $\pi_{\gamma}$. As a result, $\{\theta_k^{\circ}(\gamma)\}_{k\in\NN}$ is a stationary ergodic Markov chain. Recall the recursive function $F_X(\theta)$ in (\ref{eq_F_recursive}) and the partial sum $S_n^{\circ}=\sum_{k=1}^n\varepsilon_k^{\circ}\in\RR^{d\ell}$ in (\ref{eq_clt_partial_sum}). Note that for each $0<\gamma_s<\gamma(2)$, $s=1,\ldots,\ell$, we have
\begin{align}
    \sum_{i=1}^n\big[\theta_i^{\circ}(\gamma_s) - \theta_{\infty}^*(\gamma_s)\big] 
    & = F_{X_1}(\theta_0^{\circ}(\gamma_s)) - \EE[F_{X_1}(\theta_0^{\circ}(\gamma_s))]+\cdots \nonumber \\
    & \quad + F_{X_n}\circ \cdots\circ F_{X_1}(\theta_0^{\circ}(\gamma_s)) - \EE[F_{X_n}\circ \cdots\circ F_{X_1}(\theta_0^{\circ}(\gamma_s))].
\end{align}
By Equation~(44) and Theorem~3(ii) of \textcite{wu_asymptotic_2011}, since GMC ensures the short-range dependence of $\theta_i^{\circ}(\gamma_s)$, the $\sqrt{n}$-central limit theorem holds for the partial sum $S_n^{\circ}=\sum_{k=1}^n\varepsilon_k^{\circ}$ via the Cramér-Wold device, that is,
\begin{equation}
    \label{eq_thm2_clt}
    n^{-1/2}S_n^{\circ} \Rightarrow N(0,\Sigma^{\circ}),
\end{equation}
where the asymptotic covariance matrix $\Sigma^{\circ}$ is defined by the long-run covariance matrix of $\varepsilon_k^{\circ}$ by Equation~(48) in~\textcite{wu_asymptotic_2011}, i.e., $\Sigma^{\circ}=\sum_{k=-\infty}^{\infty}\mathrm{cov}(\varepsilon_0^{\circ},\varepsilon_k^{\circ})$.
Similarly, we recall the partial sum $S_n$ with any initial point $\theta_0\in\RR^d$ defined in Theorem~\ref{thm2_quenched_CLT}. Since $\theta_0$ may not follow the stationary distribution $\pi_{\gamma_s}$, the sequence $\{\theta_k(\gamma_s)\}_{k\in\NN}$ can be non-stationary. Therefore, we need to use the GMC property in Theorem~\ref{thm_GMC} to show that $\{\theta_k(\gamma_s)\}_{k\in\NN}$ is asymptotically stationary and the quenched CLT holds. By Theorem~\ref{thm_GMC} and the triangle inequality, it follows that, for each $0<\gamma_s<\gamma(2)$, $s=1,\ldots,\ell$,
\begin{align}
    \label{eq_thm2_diff}
    & \quad \Big\|\sum_{i=1}^n\big[\theta_i^{\circ}(\gamma_s) - \theta_i(\gamma_s)\big] \Big\|_2 \nonumber \\
    & = \big\|F_{X_1}(\theta_0^{\circ}(\gamma_s))+F_{X_2}\circ F_{X_1}(\theta_0^{\circ}(\gamma_s)) + \cdots + F_{X_n}\circ \cdots\circ F_{X_1}(\theta_0^{\circ}(\gamma_s)) \nonumber \\
    &\quad - \big[F_{X_1}(\theta_0(\gamma_s))+F_{X_2}\circ F_{X_1}(\theta_0(\gamma_s)) + \cdots + F_{X_n}\circ \cdots\circ F_{X_1}(\theta_0(\gamma_s))\big]\big\|_2 \nonumber \\
    & \le \sum_{i=1}^n[\rho_{\gamma,2}]^i\|\theta_0^{\circ}(\gamma_s)-\theta_0(\gamma_s)\|_2
    \le \frac{\rho_{\gamma,2}}{1-\rho_{\gamma,2}} 
    \|\theta_0^{\circ}(\gamma_s)-\theta_0(\gamma_s)\|_2,
\end{align}
where $\rho_{\gamma,2}$ is defined in Theorem \ref{thm_GMC}. Thus $|S_n^{\circ} - S_n| = O_{\PP}(1)$, which by \eqref{eq_thm2_clt} yields the quenched CLT
\begin{equation}
    \label{eq_thm2_clt_quench}
    n^{-1/2}S_n \Rightarrow N(0,\Sigma^{\circ}), 
\end{equation}
for any starting point $\theta_0\in\RR^d$, which completes the proof.
\end{proof}

\subsection{Proof of Theorem~\ref{thm_invar_prin}}
Before showing the proof for Theorem~\ref{thm_invar_prin}, we first introduce the functional dependence measure (\cite{wu_nonlinear_2005}), which is the key towards the proof. In particular, we provide a detailed form of the functional dependence measure tailored for the stationary SGD process.

Let $X_i$, $X_j'$, $i,j\in\ZZ$ be i.i.d. random variables following the distribution $\Pi$ in (\ref{eq_opt_goal}). Define the filtration $\F_i=(\ldots,X_{i-1},X_i)$ and the coupled version $\F_{i,\{k\}}=(\ldots,X_{k-1},X_k',X_{k+1},$ $\ldots,X_i)$. Then, there exists some measurable function $H_{\gamma}$ such that the stationary SGD sequence $\{\theta_i^{\circ}(\gamma)\}_{i\in\NN}$ can be written as the following causal process
\begin{align}
    \theta_i^{\circ}(\gamma) = H_{\gamma}(\ldots,X_{i-1},X_i) = H_{\gamma}(\F_i).
\end{align}
Let $p>1$. We define the functional dependence measure of $\theta_i^{\circ}(\gamma)$ as
\begin{equation}
    \delta_{i,p}(\gamma) = \|\theta_i^{\circ}(\gamma)-\theta_{i,\{0\}}^{\circ}(\gamma)\|_p, \quad \text{where }\theta_{i,\{0\}}^{\circ}(\gamma) = H_{\gamma}(\F_{i,\{0\}}).
\end{equation}
Further, if $\sum_{i=0}^{\infty}\delta_{i,p}<\infty$, we define the tail of cumulative dependence measure as
\begin{equation}
    \Theta_{m,p} = \sum_{i=m}^{\infty}\delta_{i,p}.
\end{equation}

Essentially, since the geometric moment contraction of the SGD iterates indicates the exponential decay of the functional dependence measure, one can quantify the error bound of the Gaussian approximation using the sample size $n$ and the $p$-th finite moment condition in Assumption \ref{asm_Lip} with $p>2$. We can obtain the best case rate of $n^{1/p}$ for any arbitrary initial point $\theta_0\in\RR^d$.

\begin{proof}[Proof of Theorem \ref{thm_invar_prin}]
We shall first consider the random process $\{\theta_k^{\circ}(\gamma)\}_{k\in\NN}$ following the stationary distribution $\pi_{\gamma}$. 
As a direct consequence of Theorem~\ref{thm_GMC}, the functional dependence measure $\delta_{k,p}(\gamma)$ converges at an exponential rate as $k\rightarrow\infty$, that is, for some constant $0<\rho_{\gamma}<1$,
\begin{equation}
    \delta_{k,p}(\gamma)=O(\rho_{\gamma}^k),
\end{equation}
where the constant step size $\gamma\in(0,\gamma(p))$ satisfies the conditions in Theorem \ref{thm_GMC}. Therefore, there exists a constant $A>0$ such that the tail cumulative dependence measure of the stationary SGD process $\{\theta_k^{\circ}(\gamma)\}_{k\in\NN}$ can be bounded by
\begin{align}
    \Theta_{i,p}(\gamma) =\sum_{k=i}^{\infty}\delta_{k,p}(\gamma) = O\big\{i^{-\chi}(\log(i))^{-A}\big\},
\end{align}
for every $\chi>0$. Recall the stationary partial sum sequence $S_i^{\circ}=\sum_{k=1}^i\varepsilon_k^{\circ}$ defined in (\ref{eq_clt_partial_sum}). Then, it follows from Theorem 2 in \textcite{karmakar_optimal_2020} that there exists a (richer) probability space $(\tilde\Omega,\tilde\A,\tilde\PP)$ on which we can define random vectors $\tilde\varepsilon^{\circ}_k\in\RR^{d\ell}$ with the partial sum process $\tilde S_i^{\circ}=\sum_{k=1}^i\tilde\varepsilon_k^{\circ}$, and a Gaussian process $\tilde G_i^{\circ}=\sum_{k=1}^i\tilde Z_k^{\circ}$, where $\tilde Z_k^{\circ}$ is a mean zero independent Gaussian vector in $\RR^{d\ell}$ with an identity covariance matrix, such that $(\tilde S_i^{\circ})_{i\ge1}\stackrel{\D}{=}(S_i^{\circ})_{i\ge1}$ and
\begin{equation}
    \label{eq_invar_prin_stationary}
    \max_{i\le n}\big|\tilde S_i^{\circ} - \Sigma^{\circ1/2}\tilde G_i^{\circ}\big|=o_{\PP}(n^{1/p}), \quad \text{in }(\tilde\Omega,\tilde\A,\tilde\PP).
\end{equation}

Next, recall the partial sum sequence $S_i=\sum_{k=1}^i\varepsilon_k$ defined in Theorem~\ref{thm2_quenched_CLT}. We shall bound the difference $\|\max_{i\leq n}|S_i-S_i^{\circ}|\|_p$, for $p>2$. Note that, for each $0<\gamma_s<\gamma(p)$, $s=1,\ldots,\ell$, we write $S_i(\gamma_s):=\sum_{k=1}^i\theta_k(\gamma_s)$ and $S_i^{\circ}(\gamma_s):=\sum_{k=1}^i\theta_k^{\circ}(\gamma_s)$ for simplicity. Then, we have
\begin{align}
    \tilde S_i(\gamma_s) & := S_i(\gamma_s) - S_i^{\circ}(\gamma_s) \nonumber \\
    & = \big[F_{X_1}(\theta_0(\gamma_s)) - F_{X_1}(\theta_0^{\circ}(\gamma_s))\big] + \big[F_{X_2}\circ F_{X_1}(\theta_0) -  F_{X_2}\circ F_{X_1}(\theta_0^{\circ}(\gamma_s))\big]+ \cdots  \nonumber \\
    &\quad + \big[F_{X_i}\circ \cdots\circ F_{X_1}(\theta_0(\gamma_s))- F_{X_i}\circ \cdots\circ F_{X_1}(\theta_0^{\circ}(\gamma_s))\big]\nonumber \\
    & =: \sum_{k=1}^i \tilde Y_i(\theta_0(\gamma_s),\theta_0^{\circ}(\gamma_s)).
\end{align}
Recall the geometric-moment contraction in Theorem~\ref{thm_GMC} which holds for any $\gamma_s\in(0,\gamma(p))$. Given the contraction constant $0<\rho_{\gamma,p}<1$ defined in Theorem~\ref{thm_GMC}, we have
\begin{align*}
    \|\tilde Y_i(\theta_0(\gamma_s),\theta_0^{\circ}(\gamma_s))\|_p \le  [\rho_{\gamma,p}]^i\|\theta_0 -\theta_0^{\circ}\|_p.
\end{align*}
Write $\tilde Y_i=\tilde Y_i(\theta_0(\gamma),\theta_0^{\circ}(\gamma))$. Also note that by the triangle inequality, it follows that $|\tilde S_i(\gamma)|\le \sum_{k=1}^i|\tilde Y_k|$, which is a non-decreasing sequence as $i$ grows. Therefore, $\max_{1\le i\le n}|\tilde S_i(\gamma)| \le \max_{1\le i\le n}\sum_{k=1}^i|\tilde Y_i| = \sum_{k=1}^n|\tilde Y_i|$, which along with the triangle inequality further yields
\begin{align}
    \label{eq_thm3_diff}
    \big\|\max_{1\le i\le n}|\tilde S_i(\gamma)|\big\|_p \le \Big\|\sum_{k=1}^n|\tilde Y_i|\Big\|_p \le \sum_{i=1}^n\|\tilde Y_i\|_p \le
    \frac{\rho_{\gamma,p}(1-[\rho_{\gamma,p}]^n)}{1-\rho_{\gamma,p}}\|\theta_0-\theta_0^{\circ}\|_p.
\end{align}
Hence, when the step $n$ is large, for $p>2$, we can obtain
\begin{equation}
    \label{eq_thm3_diff_bd}
    \frac{\big\|\max_{1\le i\le n}|S_i(\gamma)-S_i^{\circ}(\gamma)|\big\|_p}{n^{1/p}} \rightarrow0, \quad \text{as }n\rightarrow\infty.
\end{equation}
Since $(\tilde S_i^{\circ})_{i\geq1}\stackrel{\D}{=}(S_i^{\circ})_{i\geq1}$, we may enlarge the probability space to construct $(\tilde S_i)_{i\geq1}$ such that
\begin{align*}
    ((\tilde S_i,\tilde S_i^{\circ}))_{i\geq1}\stackrel{\D}{=}((S_i,S_i^{\circ}))_{i\geq1}.
\end{align*}
Therefore, by~\eqref{eq_thm3_diff_bd} and~\eqref{eq_invar_prin_stationary},
\begin{align*}
    \max_{1\le i\leq n}\big|\tilde S_i-\Sigma^{\circ1/2}\tilde G_i^{\circ}\big|
    \leq \max_{1\le i\leq n}|\tilde S_i-\tilde S_i^{\circ}|
    +\max_{1\le i\leq n}\big|\tilde S_i^{\circ}-\Sigma^{\circ1/2}\tilde G_i^{\circ}\big|=o_{\PP}(n^{1/p}),
\end{align*}
which proves the result for an arbitrary initial point.
\end{proof}

\subsection{Proofs of Theorems~\ref{Theorem_first_order_derivative}~\&~\ref{Theorem_Second_Derivative_Bound}}

\begin{proof}[Proof of Theorem~\ref{Theorem_first_order_derivative}]
    Let $\delta \in \mathbb{R}^{d}$ be a unit vector, and let $\theta_{t} = \theta + t \delta$. By Theorem~\ref{thm_GMC} and Fatou's lemma, 
    \begin{align*}
        \rho^{p} &\geq \liminf_{t \downarrow 0} \mathbb{E} \left(\frac{|F_{X}(\theta_{t}) - F_{X}(\theta)|^{p}}{t^{p}}\right) \cr 
        &\geq \mathbb{E} \liminf_{t \downarrow 0} \left(\frac{|F_{X}(\theta_{t}) - F_{X}(\theta)|^{p}}{t^{p}}\right) = \mathbb{E} (|\bm{A}_{\gamma} (\theta, X) \delta|^{p}). 
    \end{align*}
    Then for any $\mathcal{F}_{k - 1}$-measurable random vector $V \in \mathbb{R}^{d}$, we have 
    \begin{align}
    \label{eq_Contraction_A}
        \mathbb{E} (|\bm{A}_{\gamma}(\theta, X_{k}) V|^{p}) = \mathbb{E} \{\mathbb{E} (|\bm{A}_{\gamma}(\theta, X_{k}) V|^{p} | \mathcal{F}_{k - 1})\} \leq \rho^{p} \mathbb{E} (|V|^{p}).   
    \end{align}
    An elementary calculation shows that the solution to~\eqref{eq_derivative} takes the following form 
    \begin{align*}
        \dot{\theta}_{k}(\gamma) = - \sum_{h = 0}^{\infty} \left(\prod_{\ell = 1}^{h} \bm{A}_{\gamma} (\theta_{k - \ell}^{\circ}(\gamma), X_{k + 1 - \ell})\right) \nabla g(\theta_{k - 1 - h}^{\circ}(\gamma), X_{k - h}), 
    \end{align*}
    where $\prod_{\ell = 1}^{0} \bm{A}_{\gamma} (\theta_{k - \ell}^{\circ}(\gamma), X_{k + 1 - \ell}) = I_{d}$ and 
    \begin{align*}
        \prod_{\ell = 1}^{h} \bm{A}_{\gamma} (\theta_{k - \ell}^{\circ}(\gamma), X_{k + 1 - \ell}) = \bm{A}_{\gamma} (\theta_{k - 1}^{\circ}(\gamma), X_{k}) \cdots \bm{A}_{\gamma} (\theta_{k - h}^{\circ}(\gamma), X_{k + 1 - h}), \quad h \geq 1. 
    \end{align*}
    Consequently, by~\eqref{eq_Contraction_A}, Assumption~\ref{asm_Lip} and the triangle inequality, we have 
    \begin{align}
    \label{eq_first_derivative_moment_bound}
        \|\dot{\theta}_{k}(\gamma)\|_{p} \leq \sum_{h = 0}^{\infty} \rho^{h} \|\nabla g(\theta_{k - 1 - h}^{\circ}(\gamma), X_{k - h})\|_{p} = \frac{1}{1-\rho}\|\nabla g(\theta_{k - 1}^{\circ}(\gamma), X_{k})\|_{p}< \infty. 
    \end{align}
    Now we upper bound $\mathbb{E} |\partial \theta_{k}(\gamma)/\partial \gamma - \dot{\theta}_{k}(\gamma)|$. Observe that  
    \begin{align*}
        \partial_{\gamma} \theta_{k}(\gamma) - \dot{\theta}_{k}(\gamma) &= \bm{A}_{\gamma} (\theta_{k - 1}(\gamma), X_{k}) (\partial_{\gamma} \theta_{k - 1}(\gamma) - \dot{\theta}_{k - 1}(\gamma)) \cr 
        &+ (\bm{A}_{\gamma} (\theta_{k - 1}(\gamma), X_{k}) - \bm{A}_{\gamma} (\theta_{k - 1}^{\circ}(\gamma), X_{k})) \dot{\theta}_{k - 1}(\gamma) \cr 
        &+ \nabla g(\theta_{k - 1}^{\circ}(\gamma), X_{k}) - \nabla g(\theta_{k - 1}(\gamma), X_{k}) \cr 
        &=: \Delta_{1}(\gamma) + \Delta_{2}(\gamma) + \Delta_{3}(\gamma). 
    \end{align*}
    It follows by~\eqref{eq_Contraction_A} that $\mathbb{E} |\Delta_{1}(\gamma)| \leq \rho \mathbb{E} |\partial_{\gamma} \theta_{k - 1}(\gamma) - \dot{\theta}_{k - 1}(\gamma)|$. Recall that we have assumed $\|\nabla^{2} g(\theta, X) - \nabla^2 g(\theta', X)\|_{p} \leq L_{p}^{\circ} |\theta - \theta'|$. This together with~\eqref{eq_Contraction_A} gives
    \begin{align}
    \label{eq_moment_bound_iota}
        \mathbb{E} |\Delta_{2}(\gamma)| &\leq \mathbb{E} \min\left\{2 \rho |\dot{\theta}_{k - 1}(\gamma)|, \gamma L_{p}^{\circ} |\theta_{k - 1}(\gamma) - \theta_{k - 1}^{\circ}(\gamma)| |\dot{\theta}_{k - 1}(\gamma)|\right\} \cr 
        &\leq \mathbb{E} \left(2 \rho |\dot{\theta}_{k - 1}(\gamma)|\right)^{1 - \iota} \left(\gamma L_{p}^{\circ} |\theta_{k - 1}(\gamma) - \theta_{k - 1}^{\circ}(\gamma)| |\dot{\theta}_{k - 1}(\gamma)|\right)^{\iota} \cr 
        &\lesssim \|\theta_{k - 1}(\gamma) - \theta_{k - 1}^{\circ}(\gamma)\|_{p}^{\iota} \|\dot{\theta}_{k - 1}(\gamma)\|_{p/(p - \iota)} \lesssim \rho^{\iota (k - 1)}, 
    \end{align}
    where $\iota = \min\{p - 1, 1\}$. Combining these with $\mathbb{E} |\Delta_{3}(\gamma)| \leq L_{p} \rho^{k - 1} \|\theta_{0} - \theta_{0}^{\circ}(\gamma)\|_{p}$ yields 
    \begin{align*}
        \mathbb{E} |\partial_{\gamma} \theta_{k}(\gamma) - \dot{\theta}_{k}(\gamma)| \leq \rho \mathbb{E} |\partial_{\gamma} \theta_{k - 1}(\gamma) - \dot{\theta}_{k - 1}(\gamma)| + C \rho^{\iota(k - 1)}. 
    \end{align*} 
    Then, we have $\mathbb{E} |\partial_{\gamma} \theta_{k}(\gamma) - \dot{\theta}_{k}(\gamma)| = O(\rho_{\circ}^{k})$, where $\rho_{\circ} = \rho^{\iota/2} \in (0, 1)$. 
\end{proof}

\begin{proof}[Proof of Theorem~\ref{Theorem_Second_Derivative_Bound}]
    For simplicity of notation, write 
    \begin{align*}
        \mathcal{E}_{k}(\gamma) = - 2 \nabla^{2} g(\theta_{k - 1}^{\circ} (\gamma), X_{k}) \dot{\theta}_{k - 1}(\gamma) - \gamma \mathrm{vec}(\dot{\theta}_{k - 1}^{\top}(\gamma) \nabla^{3} g(\theta_{k - 1}^{\circ}(\gamma), X_{k}) \dot{\theta}_{k - 1}(\gamma)), 
    \end{align*}
    Hence,
    $$\ddot{\theta}_{k}(\gamma) = \bm{A}_{\gamma} (\theta_{k - 1}^{\circ}(\gamma), X_{k}) \ddot{\theta}_{k - 1}(\gamma) + \mathcal{E}_{k}(\gamma),$$ 
    and the solution to~\eqref{eq_derivative_second} takes the following form, 
    \begin{align*}
        \ddot{\theta}_{k}(\gamma) = \sum_{h = 0}^{\infty} \left(\prod_{\ell = 1}^{h} \bm{A}_{\gamma} (\theta_{k - \ell}^{\circ}(\gamma), X_{k + 1 - \ell})\right) \mathcal{E}_{k - h}(\gamma).  
    \end{align*}
    Then, similar to~\eqref{eq_first_derivative_moment_bound}, we have 
    $$\|\ddot{\theta}_{k}(\gamma)\|_{p/2} \leq \sum_{h = 0}^{\infty} \rho^{h} \|\mathcal{E}_{k - h}(\gamma)\|_{p/2} = \frac{1}{1-\rho}\|\mathcal{E}_{k}(\gamma)\|_{p/2}< \infty.$$ 
    Now we upper bound $\mathbb{E} |\partial_{\gamma}^{2} \theta_{k}(\gamma) - \ddot{\theta}_{k}(\gamma)|$. Observe that 
    \begin{align*}
        &\partial_{\gamma}^{2} \theta_{k}(\gamma) - \ddot{\theta}_{k}(\gamma) = \bm{A}_{\gamma}(\theta_{k - 1}(\gamma), X_{k}) (\partial_{\gamma}^{2} \theta_{k - 1}(\gamma) - \ddot{\theta}_{k - 1}(\gamma)) \cr 
        &+ (\bm{A}_{\gamma} (\theta_{k - 1}(\gamma), X_{k}) - \bm{A}_{\gamma} (\theta_{k - 1}^{\circ} (\gamma), X_{k})) \ddot{\theta}_{k - 1}(\gamma) \cr 
        &+ 2 \nabla^{2} g(\theta_{k - 1}^{\circ}(\gamma), X_{k}) \dot{\theta}_{k - 1}(\gamma) - 2 \nabla^{2} g(\theta_{k - 1}(\gamma), X_{k}) \partial_{\gamma}\theta_{k - 1}(\gamma) \cr 
        &+ \gamma \mathrm{vec}(\dot{\theta}_{k - 1}^{\top}(\gamma) \nabla^{3} g(\theta_{k - 1}^{\circ}(\gamma), X_{k}) \dot{\theta}_{k - 1}(\gamma)) - \gamma \mathrm{vec}(\partial_{\gamma}\theta_{k - 1}^{\top}(\gamma) \nabla^{3} g(\theta_{k - 1} (\gamma), X_{k}) \partial_{\gamma}\theta_{k - 1}(\gamma)) \cr 
        &=: \tilde{\Delta}_{1}(\gamma) + \tilde{\Delta}_{2}(\gamma) + \tilde{\Delta}_{3}(\gamma) + \tilde{\Delta}_{4}(\gamma). 
    \end{align*}
    Since $\ddot{\theta}_k(\gamma)\in\L^{p/2}$ only, let $\iota=\min\{1,p-2\}$. Since we have assumed $\|\nabla^{2} g(\theta, X) - \nabla^2 g(\theta', X)\|_{p} \leq L_{p}^{\circ} |\theta - \theta'|$, it follows from~\eqref{eq_Contraction_A} that
    \begin{align*}
        \EE\big\{|\tilde{\Delta}_2(\gamma)|\mid\mathcal{F}_{k-1}\big\}
        &\le\min\Big\{2\rho|\ddot{\theta}_{k-1}(\gamma)|,\gamma L_p^{\circ}|\ddot{\theta}_{k-1}(\gamma)||\theta_{k-1}(\gamma)-\theta_{k-1}^{\circ}(\gamma)|\Big\}\\
        &\lesssim|\ddot{\theta}_{k-1}(\gamma)||\theta_{k-1}(\gamma)-\theta_{k-1}^{\circ}(\gamma)|^{\iota}.
    \end{align*}
    Since $\iota\leq p-2$, Hölder's inequality gives
    \begin{align*}
        \EE|\tilde{\Delta}_2(\gamma)|
        \lesssim\|\ddot{\theta}_{k-1}(\gamma)\|_{p/2}\|\theta_{k-1}(\gamma)-\theta_{k-1}^{\circ}(\gamma)\|_p^{\iota}\lesssim \rho^{\iota(k-1)}.
    \end{align*}
    For the term $\tilde{\Delta}_4(\gamma)$, we have 
    \begin{align*}
        \tilde{\Delta}_{4}(\gamma)
        &= \gamma \mathrm{vec}\big(\dot{\theta}_{k-1}^{\top}(\gamma)\nabla^{3}g(\theta_{k-1}^{\circ}(\gamma),X_k)\dot{\theta}_{k-1}(\gamma)\big)  - \gamma \mathrm{vec}\big(\partial_{\gamma}\theta_{k-1}^{\top}(\gamma)\nabla^{3}g(\theta_{k-1}(\gamma),X_k)\partial_{\gamma}\theta_{k-1}(\gamma)\big) \\
        &= \gamma \mathrm{vec}\big(\dot{\theta}_{k-1}^{\top}(\gamma)[\nabla^{3}g(\theta_{k-1}^{\circ}(\gamma),X_k)-\nabla^{3}g(\theta_{k-1}(\gamma),X_k)]\dot{\theta}_{k-1}(\gamma)\big) \\
        &\quad + \gamma \mathrm{vec}\big([\dot{\theta}_{k-1}(\gamma)-\partial_{\gamma}\theta_{k-1}(\gamma)]^{\top}\nabla^{3}g(\theta_{k-1}(\gamma),X_k)\dot{\theta}_{k-1}(\gamma)\big) \\
        &\quad + \gamma \mathrm{vec}\big(\partial_{\gamma}\theta_{k-1}^{\top}(\gamma)\nabla^{3}g(\theta_{k-1}(\gamma),X_k)[\dot{\theta}_{k-1}(\gamma)-\partial_{\gamma}\theta_{k-1}(\gamma)]\big) =: \tilde\I_1 +\tilde\I_2+\tilde\I_3.
    \end{align*}
    By the condition $\|\nabla^{2} g(\theta, X) - \nabla^2 g(\theta', X)\|_{p} \leq L_{p}^{\circ} |\theta - \theta'|$ in Theorem~\ref{Theorem_first_order_derivative} and Fatou's lemma, $\sup_{\theta\in\mathbb{R}^d}\|\nabla^3g(\theta,X)\|_p<\infty$.
    Moreover, by the third-derivative Lipschitz condition $\mathbb{E} |\nabla^{3} g(\theta, X) - \nabla^{3} g(\theta', X)| \leq L_p^{\diamond} |\theta - \theta'|$ in Theorem~\ref{Theorem_Second_Derivative_Bound} and the preceding uniform bound, we have,
    \begin{align*}
        &\quad \mathbb{E}\Big\{\big|\nabla^3g(\theta_{k-1}^{\circ}(\gamma),X_k)-\nabla^3g(\theta_{k-1}(\gamma),X_k)\big|\mid\mathcal{F}_{k-1}\Big\}\\
        &\leq\min\Big\{2\sup_{\theta\in\mathbb{R}^d}\|\nabla^3g(\theta,X)\|_p,L_p^{\diamond}|\theta_{k-1}^{\circ}(\gamma)-\theta_{k-1}(\gamma)|\Big\}\\
        &\lesssim|\theta_{k-1}^{\circ}(\gamma)-\theta_{k-1}(\gamma)|^{\iota}.
    \end{align*}
    Therefore, since $\iota\leq p-2$,
    \begin{align*}
        \mathbb{E}|\tilde{\mathcal I}_1|&\lesssim\gamma\mathbb{E}\Big[|\dot{\theta}_{k-1}(\gamma)|^2|\theta_{k-1}^{\circ}(\gamma)-\theta_{k-1}(\gamma)|^{\iota}\Big]\\
        &\lesssim \gamma\|\dot{\theta}_{k-1}(\gamma)\|_p^2\|\theta_{k-1}^{\circ}(\gamma)-\theta_{k-1}(\gamma)\|_p^{\iota}\lesssim \gamma\rho^{\iota(k-1)}.
    \end{align*}
    For the remaining two terms, conditional expectation and Hölder's inequality give
    \begin{align*}
        \mathbb{E}|\tilde{\mathcal I}_2|+\mathbb{E}|\tilde{\mathcal I}_3|
        \lesssim \gamma \|\dot{\theta}_{k-1}(\gamma) -\partial_{\gamma}\theta_{k-1}(\gamma)\|_{p/(p-1)}\big[\|\dot{\theta}_{k-1}(\gamma)\|_p+\|\partial_{\gamma}\theta_{k-1}(\gamma)\|_p\big].
    \end{align*}
    By~\eqref{eq_der_midstep}, \eqref{eq_first_derivative_moment_bound}, Assumption~\ref{asm_Lip} and Theorem~\ref{thm_GMC},
    \begin{align*}
        &\quad\sup_{k\geq0}\big\{\|\dot{\theta}_k(\gamma)\|_p+\|\partial_{\gamma}\theta_k(\gamma)\|_p\big\}\\
        &\leq\|\dot{\theta}_0(\gamma)\|_p+\|\partial_{\gamma}\theta_0(\gamma)\|_p+\frac{L_p\sup_{k\geq0}\|\theta_k(\gamma)-\theta^*\|_p+\|\nabla g(\theta^*,X)\|_p}{1-\rho}<\infty.
    \end{align*}
    Consequently, $\sup_{k\geq0}\|\dot{\theta}_k(\gamma)-\partial_{\gamma}\theta_k(\gamma)\|_p<\infty$. Together with Theorem~\ref{Theorem_first_order_derivative}, interpolation between $\L^1$ and $\L^p$ yields
    \begin{align*}
        &\quad \|\dot{\theta}_{k-1}(\gamma)-\partial_{\gamma}\theta_{k-1}(\gamma)\|_{p/(p-1)}\\
        &\leq\|\dot{\theta}_{k-1}(\gamma)-\partial_{\gamma}\theta_{k-1}(\gamma)\|_1^{(p-2)/(p-1)}\|\dot{\theta}_{k-1}(\gamma)-\partial_{\gamma}\theta_{k-1}(\gamma)\|_p^{1/(p-1)}\\
        &\lesssim\rho_{\circ}^{(k-1)(p-2)/(p-1)}.
    \end{align*}
    Let $\tilde{\rho}=\max\big\{\rho^{\iota},\rho_{\circ}^{(p-2)/(p-1)}\big\}\in(0,1)$. It follows that
    \begin{align*}
        \mathbb{E}|\tilde{\Delta}_4(\gamma)|
        &\lesssim\gamma\Big\{\rho^{\iota(k-1)}+\rho_{\circ}^{(k-1)(p-2)/(p-1)}\Big\}\lesssim \tilde{\rho}^{k}.
    \end{align*}
    Under the conditions of Theorem~\ref{Theorem_first_order_derivative} with $p > 2$, we have 
    \begin{align*}
        \mathbb{E} |\tilde{\Delta}_{3}(\gamma)| &\leq 2 \mathbb{E} |\nabla^{2} g(\theta_{k - 1}(\gamma), X_{k}) (\dot{\theta}_{k - 1}(\gamma) - \partial_{\gamma}\theta_{k - 1}(\gamma))| \cr 
        &+ 2 L_p^{\circ} \mathbb{E} |\theta_{k - 1}^{\circ} (\gamma) - \theta_{k - 1}(\gamma)| |\dot{\theta}_{k - 1} (\gamma)| \lesssim \rho_{\circ}^{k}. 
    \end{align*}
    By~\eqref{eq_Contraction_A} and conditional Jensen's inequality,
    \begin{align*}
        \mathbb{E}|\tilde{\Delta}_1(\gamma)|&\leq\rho\mathbb{E}\big|\partial_{\gamma}^{2}\theta_{k-1}(\gamma)-\ddot{\theta}_{k-1}(\gamma)\big|.
    \end{align*}
    Combining the bounds for $\tilde{\Delta}_1(\gamma),\ldots,\tilde{\Delta}_4(\gamma)$, and using $\rho^{\iota(k-1)}\lesssim\tilde{\rho}^{k}$, we have
    \begin{align*}
        &\quad \mathbb{E}\big|\partial_{\gamma}^{2}\theta_k(\gamma)-\ddot{\theta}_k(\gamma)\big|\leq\rho\mathbb{E}\big|\partial_{\gamma}^{2}\theta_{k-1}(\gamma)-\ddot{\theta}_{k-1}(\gamma)\big|+C\big\{\rho_{\circ}^{k}+\tilde{\rho}^{k}\big\}.
    \end{align*}
    Therefore, we obtain 
    $$\mathbb{E}|\partial_{\gamma}^{2} \theta_{k}(\gamma) - \ddot{\theta}_{k}(\gamma)| = O(\rho_{\diamond}^{k}),$$ 
    where $\rho_{\diamond} = \max\{\rho_{\circ}^{1/2},\tilde{\rho}^{1/2},\rho^{1/2}\} \in (0, 1)$.  
\end{proof}

\subsection{Proof of Theorem~\ref{thm_bias_expansion}}

\begin{proof}[Proof of Theorem~\ref{thm_bias_expansion}]
In this proof, we omit the dependence of $\theta_n^{\circ}(\gamma)$ on the iteration index $n$ and step size $\gamma$ with no confusion caused, and write $\theta^{\circ}=\theta_n^{\circ}(\gamma)$. For notational convenience write $\theta^{\circ}=(\theta_1,\ldots,\theta_d)^{\top}$. W.l.o.g., assume $\theta^*=0$. Since $\EE|\theta_n^{\circ}(\gamma)|^p=O(\gamma^{p/2})$ by Theorem~\ref{thm_GMC}, to find the coefficient in front of $\gamma^2$ in bias expansion, we shall solve all the unknown parameters shown in Table~\ref{table_class_eq}. Let $\mu_{\bm{b}}=\EE[\theta^{\bm{b}}]$ with $\bm{b}=(b_1,\ldots,b_d)^{\top}$, $b_i\in\{0,1,2,3,4,5\}$, where $\theta^{\bm{b}}=\theta_1^{b_1}\cdots\theta_d^{b_d}$.  
\begin{table}[!htb]
    \centering
    \begin{tabular}{ccccc}
        \hline
        Order & $\mu_{\bm{b}}$ & Types & Index & Number of Parameters \\
        \hline
        $|\bm{b}|=1$ & $\mu_i$, $1\le i\le d$ & $\EE[\theta_1]$ & $\A_1$ & $d$\\
        $|\bm{b}|=2$ & $\mu_{ij}$, $1\le i,j\le d$ &  $\EE[\theta_1\theta_2],\,\EE[\theta_1^2]$ & $\A_2$ & ${d\choose2} +d$\\
        $|\bm{b}|=3$ & $\mu_{ijk}$, $1\le i,j,k\le d$ &  $\EE[\theta_1\theta_2\theta_3],\,\EE[\theta_1^2\theta_2],\,\EE[\theta_1^3]$ & $\A_3$ & ${d\choose3} + 2{d\choose2} + d$\\
        $|\bm{b}|=4$ & $\mu_{ijkl}$, $1\le i,j,k,l\le d$ &  $\EE[\theta_1\theta_2\theta_3\theta_4],\,\EE[\theta_1^2\theta_2\theta_3],$ & $\A_4$ & ${d\choose4} + 3{d\choose3} + 3{d\choose2} + d$ \\
         &  &  $\EE[\theta_1^3\theta_2],\,\EE[\theta_1^2\theta_2^2],\,\EE[\theta_1^4]$ &  &  \\
        \hline
    \end{tabular}
    \caption{In total $\tilde d=|\A_1|+|\A_2|+|\A_3|+|\A_4|$ unknown parameters to solve, where $|\bm{b}|=\sum_{i=1}^db_i$.}
    \label{table_class_eq}
\end{table}
By the four types of parameters in Table~\ref{table_class_eq}, we have $\tilde d=|\A_1|+|\A_2|+|\A_3|+|\A_4|$ unknown parameters in total, where
\begin{equation}
    \label{eq_tilde_d}
    \tilde{d} = 4d + 6{d\choose2} +4{d\choose3} + {d\choose4}.
\end{equation}
Next, we shall prove that we can find exactly $\tilde{d}$ equations that are linearly independent to solve these $\tilde{d}$ unknown parameters. Let $\bm{a}=(a_1,\ldots,a_{\ell})$ for $\ell\in\{1,2,3,4\}$ with $a_j\in\{1,\ldots,d\}$. Denote the expectation of the product of partial derivatives of $g(\theta,X)$ by
\begin{equation}
    m_{\bm{a}}(\theta) = \EE\Bigg(\frac{\partial g(\theta,X)}{\partial\theta_{a_1}}\cdots\frac{\partial g(\theta,X)}{\partial\theta_{a_{\ell}}}\Bigg).
\end{equation}
Note that $m_{\bm{a}}(\theta)=m_{\bm{a}'}(\theta)$ when $\bm{a}'$ is a permutation of $\bm{a}$. For a function $f(\cdot):\RR^d\mapsto\RR$, write
\begin{equation}
    f^{\bm{b}}(\theta) = \Big(\frac{1}{b_1!}\cdots\frac{1}{b_d!}\Big)\frac{\partial^{|\bm{b}|}f(\theta)}{\partial\theta_1^{b_1}\ldots\partial\theta_d^{b_d}}.
\end{equation}

\textbf{Class 1: $d$ equations.} Recall the stationary SGD iterates
\begin{equation}
    \label{eq_bias_sgd_recall}
    \theta_n^{\circ}(\gamma) = \theta_{n-1}^{\circ}(\gamma) - \gamma\nabla g(\theta_{n-1}^{\circ}(\gamma),X_n), \quad n\ge1.
\end{equation}
First, we take the expectation of (\ref{eq_bias_sgd_recall}) and obtain
\begin{equation*}
    \EE [m_{\bm{a}}(\theta)]=0.
\end{equation*}
By Assumption~\ref{asm_g_smooth}, we apply the 5-th order Taylor expansion to $ m_{\bm{a}}(\theta)$ at $\theta^*=0$ and obtain
\begin{equation}
    \label{eq_class1}
    \sum_{|\bm{b}|\le 4}m_{\bm{a}}^{\bm{b}}(0)\mu_{\bm{b}} = O(\gamma^{5/2}),
\end{equation}
for $\bm{a}=1,2,\ldots,d$. Therefore, we have $d$ equations.

\textbf{Class 2: ${d\choose2}+d$ equations.} By taking the square of (\ref{eq_bias_sgd_recall}) and taking expectation, we obtain
\begin{align*}
    \EE[\theta_i\theta_j] & = \EE\Big[\big(\theta-\gamma\nabla g(\theta,X)\big)_i\big(\theta-\gamma\nabla g(\theta,X)\big)_j\Big] \\
    & = \EE[\theta_i\theta_j] - \gamma\EE\Big[\theta_j\frac{\partial g(\theta,X)}{\partial\theta_i} + \theta_i\frac{\partial g(\theta,X)}{\partial\theta_j}\Big] + \gamma^2\EE\Big[\frac{\partial g(\theta,X)}{\partial\theta_i}\cdot\frac{\partial g(\theta,X)}{\partial\theta_j}\Big],
\end{align*}
for $1\le i,j\le d$. By applying the tower rule to the second term, we have
\begin{equation*}
    \EE\big[\theta_jm_i(\theta) + \theta_im_j(\theta)\big] = \gamma \EE[m_{(i,j)}(\theta)].
\end{equation*}
By Assumption~\ref{asm_g_smooth}, we apply the third order Taylor expansion at $\theta^*=0$ to the left-hand side (LHS) and the second order to the right-hand side (RHS), which gives
\begin{align}
    \label{eq_class2}
    \sum_{|\bm{b}|\le3}m_{i}^{\bm{b}}(0)\mu_{\bm{e}_j+\bm{b}} + \sum_{|\bm{b}|\le3}m_{j}^{\bm{b}}(0)\mu_{\bm{e}_i+\bm{b}} = \gamma\Big(m_{(i,j)}(0) + \sum_{1\le|\bm{b}|\le2}m_{(i,j)}^{\bm{b}}(0)\mu_{\bm{b}}\Big) + O(\gamma^{5/2}),
\end{align}
where $\bm{e}_j=(0,\ldots,0,1,0,\ldots,0)^{\top}$ with 1 as the $j$-th element. Note that here $i$ and $j$ are exchangeable, which indicates that we have ${d\choose2}$ equations for $i\neq j$ and $d$ equations for $i=j$.

\textbf{Class 3: ${d\choose3}+2{d\choose2}+d$ equations.} By taking the cube of (\ref{eq_bias_sgd_recall}) and taking expectation, we obtain, for $1\le i,j,k\le d$,
\begin{align*}
    & \quad \EE[\theta_i\theta_j\theta_k] \\
    & = \EE\Big[\big(\theta-\gamma\nabla g(\theta,X)\big)_i\big(\theta-\gamma\nabla g(\theta,X)\big)_j\big(\theta-\gamma\nabla g(\theta,X)\big)_k\Big] \\
    & = \EE[\theta_i\theta_j\theta_k] - \gamma\EE\Big[\theta_j\theta_k\frac{\partial g(\theta,X)}{\partial\theta_i} + \theta_i\theta_k\frac{\partial g(\theta,X)}{\partial\theta_j} + \theta_i\theta_j\frac{\partial g(\theta,X)}{\partial\theta_k}\Big] \\
    & \quad + \gamma^2\EE\Big[\theta_i\frac{\partial g(\theta,X)}{\partial\theta_j}\cdot\frac{\partial g(\theta,X)}{\partial\theta_k} + \theta_j\frac{\partial g(\theta,X)}{\partial\theta_i}\cdot\frac{\partial g(\theta,X)}{\partial\theta_k} + \theta_k\frac{\partial g(\theta,X)}{\partial\theta_i}\cdot\frac{\partial g(\theta,X)}{\partial\theta_j}\Big] \\
    & \quad -\gamma^3\EE\Big[\frac{\partial g(\theta,X)}{\partial\theta_i}\cdot\frac{\partial g(\theta,X)}{\partial\theta_j}\cdot\frac{\partial g(\theta,X)}{\partial\theta_k}\Big].
\end{align*}
By applying the tower rule to the second and third term, we have
\begin{align*}
    & \quad \EE[\theta_i\theta_jm_k(\theta) + \theta_i\theta_km_j(\theta) + \theta_j\theta_km_i(\theta)] \\
    & = \gamma\EE[\theta_im_{(j,k)}(\theta) + \theta_jm_{(k,i)}(\theta) + \theta_km_{(i,j)}(\theta)] - \gamma^2\EE[m_{(i,j,k)}(\theta)].
\end{align*}
Let $\bm{e}_{(i,j)}=\bm{e}_i+\bm{e}_j$. We apply the second order Taylor expansion at $\theta^*=0$ on the LHS and the first order to the first term on RHS, which yields
\begin{align}
    \label{eq_class3}
    & \quad \sum_{|\bm{b}|\le2}\Big[m_k^{\bm{b}}(0)\mu_{\bm{e}_{(i,j)}+\bm{b}} + m_j^{\bm{b}}(0)\mu_{\bm{e}_{(i,k)}+\bm{b}} + m_i^{\bm{b}}(0)\mu_{\bm{e}_{(j,k)}+\bm{b}}\Big] \nonumber \\
    & = \gamma\sum_{|\bm{b}|\le1}\Big[m_{(j,k)}^{\bm{b}}(0)\mu_{\bm{e}_i+\bm{b}} + m_{(i,k)}^{\bm{b}}(0)\mu_{\bm{e}_j+\bm{b}} + m_{(i,j)}^{\bm{b}}(0)\mu_{\bm{e}_k+\bm{b}}\Big] \nonumber \\
    & \quad -\gamma^2m_{(i,j,k)}(0) + O(\gamma^{5/2}).
\end{align}
Since $i$, $j$ and $k$ are all exchangeable, we have ${d\choose3}$ equations for pairwise distinct $i,j,k$, $2{d\choose2}$ equations for $i=j\neq k$ and $d$ equations for $i=j=k$.

\textbf{Class 4: ${d\choose4} +3{d\choose3}+3{d\choose2}+d$ equations.} By taking (\ref{eq_bias_sgd_recall}) to the fourth power, we can obtain, for $1\le i,j,k,l\le d$,
\begin{align*}
    \EE[\theta_i\theta_j\theta_k\theta_l] = \EE\Big[\big(\theta-\gamma\nabla g(\theta,X)\big)_i\big(\theta-\gamma\nabla g(\theta,X)\big)_j\big(\theta-\gamma\nabla g(\theta,X)\big)_k\big(\theta-\gamma\nabla g(\theta,X)\big)_l\Big],
\end{align*}
which together with the tower rule and stationarity further yields
\begin{align*}
    & \quad \EE[\theta_i\theta_j\theta_km_l(\theta) + \theta_i\theta_j\theta_lm_k(\theta) + \theta_i\theta_k\theta_lm_j(\theta) + \theta_j\theta_k\theta_lm_i(\theta)] \\
    & = \gamma\EE\Big[\theta_i\theta_jm_{(k,l)}(\theta) + \theta_i\theta_km_{(j,l)}(\theta) + \theta_i\theta_lm_{(j,k)}(\theta) \\
    & \quad\quad + \theta_j\theta_km_{(i,l)}(\theta) + \theta_j\theta_lm_{(i,k)}(\theta) + \theta_k\theta_lm_{(i,j)}(\theta)\Big] \\
    & \quad -\gamma^2\EE\Big[\theta_im_{(j,k,l)}(\theta) + \theta_jm_{(i,k,l)}(\theta) + \theta_km_{(i,j,l)}(\theta) + \theta_lm_{(i,j,k)}(\theta)\Big] \\
    & \quad + \gamma^3\EE[m_{(i,j,k,l)}(\theta)].
\end{align*}
Let $\bm{e}_{(i,j,k)}=\bm{e}_i+\bm{e}_j+\bm{e}_k$. By Taylor expansion at $\theta^*=0$, we have
\begin{align}
    \label{eq_class4}
    & \quad \sum_{|\bm{b}|\le1}\Big[m_l^{\bm{b}}(0)\mu_{\bm{e}_{(i,j,k)}+\bm{b}} + m_k^{\bm{b}}(0)\mu_{\bm{e}_{(i,j,l)}+\bm{b}} + m_j^{\bm{b}}(0)\mu_{\bm{e}_{(i,k,l)}+\bm{b}} + m_i^{\bm{b}}(0)\mu_{\bm{e}_{(j,k,l)}+\bm{b}}\Big] \nonumber \\
    & = \gamma\Big[m_{(k,l)}(0)\mu_{\bm{e}_{(i,j)}} + m_{(j,l)}(0)\mu_{\bm{e}_{(i,k)}} + m_{(j,k)}(0)\mu_{\bm{e}_{(i,l)}} + m_{(i,l)}(0)\mu_{\bm{e}_{(j,k)}} \nonumber \\
    & \quad + m_{(i,k)}(0)\mu_{\bm{e}_{(j,l)}} + m_{(i,j)}(0)\mu_{\bm{e}_{(k,l)}}\Big] + O(\gamma^{5/2}).
\end{align}
Since $i,j,k$ and $l$ are all exchangeable, we have ${d\choose4}$ equations for pairwise distinct $i,j,k,l$, $3{d\choose3}$ equations for $i=j$ with $i,k,l$ pairwise distinct, $2{d\choose2}$ equations for $i=j=k\neq l$, ${d\choose2}$ equations when $i=j\neq k=l$, and $d$ equations for $i=j=k=l$.

Finally, we shall provide the general solution for the equations \eqref{eq_class1}--\eqref{eq_class4} from Classes 1--4. Specifically, we shall solve the following linear system with perturbation
\begin{equation}
    \label{eq_A0_mat}
    \bm{A}_{\gamma}\bm{\mu} = \gamma\bm{w}_{\gamma} + O(\gamma^{5/2}), \quad \text{where }\bm{A}_{\gamma} = \bm{A}_0 + \gamma\bm{B}_0,\, \bm{w}_{\gamma} = \bm{w}_0+\gamma \bm{v}.
\end{equation}
Here, $\bm{\mu}=((\mu_{\bm{a}})_{\bm{a}\in\A_1}^{\top}, (\mu_{\bm{a}})_{\bm{a}\in\A_2}^{\top}, (\mu_{\bm{a}})_{\bm{a}\in\A_3}^{\top}, (\mu_{\bm{a}})_{\bm{a}\in\A_4}^{\top})^{\top}\in\RR^{\tilde{d}}$, matrix $\bm{A}_0\in\RR^{\tilde{d}\times\tilde{d}}$ consists of the coefficients of $\mu_{\bm{a}}$ on the LHS of \eqref{eq_class1}--\eqref{eq_class4}, and matrix $\bm{B}_0\in\RR^{\tilde{d}\times\tilde{d}}$ consists of the coefficients of $\mu_{\bm{a}}$ on the RHS. The vector $\bm{w}_0\in\RR^{\tilde{d}}$ consists of the remaining parts on the RHS without $\gamma$, and $\bm{v}\in\RR^{\tilde{d}}$ consists of the coefficients of $\gamma$. Since $\bm{A}_0$ is non-singular by Lemma~\ref{lemma_nonsingular_A0}, the solution is given by
\begin{align}
    \bm{\mu} & = \big(\bm{A}_0^{-1} - \gamma\bm{A}_0^{-1}\bm{B}_0\bm{A}_0^{-1} + O(\gamma^2)\big)\gamma\bm{w}_{\gamma} + O(\gamma^{5/2}) \nonumber \\
    & = \gamma(\bm{A}_0^{-1}\bm{w}_0) + \gamma^2(\bm{A}_0^{-1}\bm{v} - \bm{A}_0^{-1}\bm{B}_0\bm{A}_0^{-1}\bm{w}_0) + O(\gamma^{5/2}).
\end{align}
Hence, $\Delta_1$ consists of the first $d$ elements of $\bm{A}_0^{-1}\bm{w}_0$ and $\Delta_2$ consists of the first $d$ elements of $\bm{A}_0^{-1}\bm{v} - \bm{A}_0^{-1}\bm{B}_0\bm{A}_0^{-1}\bm{w}_0$. This completes the proof for the multi-dimensional case. 

\textbf{Special case: dimension $d=1$.} We give an example of one-dimensional case. Assume $\theta^*=0$. For $a=1,2,3,4$ and $b=0,1,2,3,4$, define 
\begin{equation*}
    f_a^b = \frac{1}{b!}\frac{d^b}{d\theta^b}\EE[(\nabla g(\theta,X_1))^a]\Big\vert_{\theta=0}.
\end{equation*}
Then $f_1^0=m(0)=0$. Since $d=1$, we have $\tilde d=4$ unknown parameters $\mu_k=\EE(\theta^k)$, $k=1,2,3,4$. Accordingly, equations~\eqref{eq_class1}--\eqref{eq_class4} can be reduced to the following four equations
\begin{align}
    & f_1^1\mu_1 + f_1^2\mu_2 + f_1^3\mu_3 + f_1^4\mu_4 = O(\gamma^{5/2}), \label{eq_class1_d1}\\
    & f_1^1\mu_2 + f_1^2\mu_3 + f_1^3\mu_4 = \frac{\gamma}{2}\Big[f_2^0 + f_2^1\mu_1 +f_2^2\mu_2\Big] + O(\gamma^{5/2}), \label{eq_class2_d1}\\
    & f_1^1\mu_3 + f_1^2\mu_4 = \gamma\Big[f_2^0\mu_1 + f_2^1\mu_2\Big] - \frac{\gamma^2}{3}f_3^0 + O(\gamma^{5/2}), \label{eq_class3_d1}\\
    & f_1^1\mu_4 = \frac{3\gamma}{2}f_2^0\mu_2 + O(\gamma^{5/2}) \label{eq_class4_d1}.
\end{align}
By \eqref{eq_class1_d1}--\eqref{eq_class4_d1}, we can obtain the linear system $\bm{A}_{\gamma}\bm{\mu}=\gamma\bm{w}_{\gamma}+O(\gamma^{5/2})$ as follows:
\begin{align}
\begin{pmatrix}
f_1^1 & f_1^2 & f_1^3 & f_1^4 \\
-\frac{\gamma}{2}f_2^1 & f_1^1 - \frac{\gamma}{2}f_2^2 & f_1^2 & f_1^3 \\
-\gamma f_2^0 & -\gamma f_2^1 & f_1^1 & f_1^2 \\
0 & -\frac{3}{2}\gamma f_2^0 & 0 & f_1^1
\end{pmatrix}
\begin{pmatrix}
\mu_1 \\
\mu_2 \\
\mu_3 \\
\mu_4 
\end{pmatrix}
=
\begin{pmatrix}
0 \\
\frac{\gamma}{2}f_2^0 \\
- \frac{\gamma^2}{3}f_3^0 \\
0
\end{pmatrix}
+O(\gamma^{5/2}).
\end{align}
Here, we can also write $\bm{A}_{\gamma}=\bm{A}_0+\gamma\bm{B}_0$, where
\begin{align*}
    \bm{A}_0 = 
\begin{pmatrix}
f_1^1 & f_1^2 & f_1^3 & f_1^4 \\
0 & f_1^1  & f_1^2 & f_1^3 \\
0 & 0 & f_1^1 & f_1^2 \\
0 & 0 & 0 & f_1^1
\end{pmatrix},
\bm{B}_0=
\begin{pmatrix}
0 & 0 & 0 & 0 \\
-\frac{1}{2}f_2^1 & - \frac{1}{2}f_2^2 & 0 & 0 \\
-f_2^0 & -f_2^1 & 0 & 0 \\
0 & -\frac{3}{2}f_2^0 & 0 & 0
\end{pmatrix},
\bm{w}_0=
\begin{pmatrix}
0 \\
\frac{1}{2}f_2^0 \\
0 \\
0
\end{pmatrix},
\bm{v}=
\begin{pmatrix}
0 \\
0 \\
-\frac{1}{3}f_3^0 \\
0
\end{pmatrix}.
\end{align*}
By construction, $\bm{A}_0$ is an upper triangular Toeplitz matrix. Moreover, Assumption~\ref{asm_conv} implies that $f_1^1\neq0$. Hence, $\det(\bm{A}_0)=(f_1^1)^4\neq0$, and therefore $\bm{A}_0$ is non-singular. Finally, the solution is given by
\begin{align*}
    \bm{\mu} & = \big(\bm{A}_0^{-1} - \gamma\bm{A}_0^{-1}\bm{B}_0\bm{A}_0^{-1} + O(\gamma^2)\big)\gamma\bm{w}_{\gamma} + O(\gamma^{5/2}) \\
    & = \gamma(\bm{A}_0^{-1}\bm{w}_0) + \gamma^2(\bm{A}_0^{-1}\bm{v} - \bm{A}_0^{-1}\bm{B}_0\bm{A}_0^{-1}\bm{w}_0) + O(\gamma^{5/2}),
\end{align*}
where 
\begin{align*}
\bm{A}_0^{-1} = \begin{pmatrix}
\frac{1}{f_1^1} & -\frac{f_1^2}{(f_1^1)^2} & \frac{(f_1^2)^2 - f_1^1 f_1^3}{(f_1^1)^3} & -\frac{(f_1^2)^3 - 2f_1^1 f_1^2 f_1^3 + (f_1^1)^2 f_1^4}{(f_1^1)^4} \\
0 & \frac{1}{f_1^1} & -\frac{f_1^2}{(f_1^1)^2} & \frac{(f_1^2)^2 - f_1^1 f_1^3}{(f_1^1)^3} \\[10pt]
0 & 0 & \frac{1}{f_1^1} & -\frac{f_1^2}{(f_1^1)^2} \\
0 & 0 & 0 & \frac{1}{f_1^1}
\end{pmatrix}.
\end{align*}
Denote the first row of $\bm{A}_0^{-1}$ by $[\bm{A}_0^{-1}]_{(1,:)}$. Then, we can obtain
\begin{align}
    \Delta_1=[\bm{A}_0^{-1}]_{(1,:)}\bm{w}_0 = -\frac{f_1^2f_2^0}{2(f_1^1)^2},
\end{align}
and 
\begin{align}
    \Delta_2 & = [\bm{A}_0^{-1}]_{(1,:)}\bm{v} - [\bm{A}_0^{-1}\bm{B}_0\bm{A}_0^{-1}]_{(1,:)}\bm{w}_0 \nonumber \\
    & = \frac{f_1^3 f_3^0}{3 (f_1^1)^{2}} - \frac{(f_1^2)^{2} f_3^0}{3 (f_1^1)^{3}} - \frac{f_1^2 f_2^0 f_2^2}{4 (f_1^1)^{3}} - \frac{f_1^3 f_2^0 f_2^1}{2 (f_1^1)^{3}} - \frac{3 f_1^4 (f_2^0)^{2}}{4 (f_1^1)^{3}} \nonumber \\
    & \quad + \frac{3(f_1^2)^{2} f_2^0 f_2^1}{4 (f_1^1)^{4}} + \frac{2 f_1^2 f_1^3 (f_2^0)^{2}}{ (f_1^1)^{4}} - \frac{5 (f_1^2)^{3} (f_2^0)^{2}}{4 (f_1^1)^{5}},
\end{align}
where $\Delta_2$ is calculated by the package {\tt ‘SymPy’} in Python.
\end{proof}

\begin{lemma}[Non-singularity of $\bm{A}_0$]\label{lemma_nonsingular_A0}
For any $\tilde{d}$ defined in~\eqref{eq_tilde_d} with $d\ge1$, under the same conditions of Theorem~\ref{thm_bias_expansion}, the matrix $\bm{A}_0\in\RR^{\tilde{d}\times\tilde{d}}$ defined in~\eqref{eq_A0_mat} is non-singular.
\end{lemma}

\begin{proof}[Proof of Lemma~\ref{lemma_nonsingular_A0}]
Recall that $m(\theta)=\nabla G(\theta)$ as defined in Assumption~\ref{asm_conv}. W.l.o.g., assume $\theta^*=0$ and define a $d\times d$ matrix
\begin{align*}
    \bm{H} =\nabla m(0) =(h_{ij})_{i,j=1}^d,\quad \text{where }h_{ij} = m_i^{\bm{e}_j}(0),
\end{align*}
and $\bm{e}_j=(0,\ldots,0,1,0,\ldots,0)^{\top}$ with 1 as the $j$-th element. Thus, $\bm{H}$ is symmetric. By Assumption~\ref{asm_conv}, for any $u\in\RR^d$, it follows that
\begin{align*}
    u^{\top}\bm{H}u = \lim_{t\rightarrow0}\frac{\langle m(tu) - m(0), tu\rangle}{t^2} \ge \mu|u|^2.
\end{align*}
Therefore, $\bm{H}$ is a positive definite matrix with $\bm{H}\succeq \mu \bm{I}_d$. Let $\lambda_1,\ldots,\lambda_d$ be the eigenvalues of $\bm{H}$ with corresponding orthonormal eigenvectors $q_1,\ldots,q_d$. Then, for each $1\le i\le d$, $\bm{H}q_i=\lambda_iq_i$ and $|q_i|=1$, which gives
\begin{align}
    \label{eq_lower_bd_eigen_H}
    \lambda_i=q_i^{\top}\bm{H}q_i \ge \mu|q_i|^2 = \mu >0.
\end{align}

Next, we show that $\bm{A}_0$ is a block upper triangle matrix. Recall the four classes defined in Table~\ref{table_class_eq}. For $r=1,2,3,4$, define a set of multi-indices of total degree $r$, i.e.,
\begin{align}
    \label{eq_A0_index_set}
    \I_r = \{\bm{\alpha}=(\alpha_1,\ldots,\alpha_d)\in\NN^d: \, |\bm{\alpha}|=r\}, \quad N_r=|\I_r| = {d+r-1\choose r}.
\end{align}
We label rows and columns in $\bm{A}_0$ by the multi-index $\bm{\alpha}$ in the order $\I_1,\I_2,\I_3,\I_4$, which partitions $\bm{A}_0$ into blocks as follows
\begin{align*}
    \bm{A}_0 = \begin{pmatrix}
        \bm{D}_1& \bm{C}_{12} & \bm{C}_{13} & \bm{C}_{14} \\
        \bm{C}_{21} & \bm{D}_2 & \bm{C}_{23} & \bm{C}_{24} \\
        \bm{C}_{31} & \bm{C}_{32} & \bm{D}_3 & \bm{C}_{34} \\
        \bm{C}_{41} & \bm{C}_{42} & \bm{C}_{43} & \bm{D}_4
    \end{pmatrix},
\end{align*}
where $\bm{D}_r:=\bm{C}_{rr}$, and $\bm{C}_{rs}$ has rows labeled by $\bm{\alpha}\in\I_r$ and columns labeled by $\bm{\beta}\in\I_s$. For example, when $d=2$, $\I_1=\{(1,0),(0,1)\}$ and $\I_2=\{(2,0),(1,1),(0,2)\}$. Then, for $\bm{\nu}=(\nu_1,\ldots,\nu_d)^{\top}$ with $1\le |\bm{\nu}|\le 4$, write
\begin{align*}
    c_{i,\bm{\nu}} = \frac{\partial^{\bm{\nu}}m_i(0)}{\bm{\nu}!} = \frac{\partial^{\bm{\nu}}m_i(0)}{\nu_1!\cdots\nu_d!}=m
    _i^{\bm{\nu}}(0), \quad c_{i,\bm{e}_j} = h_{ij}.
\end{align*}
By the definition of $\bm{A}_0$, we group the repeated indices in the LHS of~\eqref{eq_class1}--\eqref{eq_class4}, which gives, for $\bm{\alpha}\in\I_r$, the expression whose Taylor coefficients form the corresponding row of $\bm{A}_0$ is 
\begin{align*}
    \sum_{\{i: \alpha_i>0\}}\alpha_i\theta^{\bm{\alpha}-\bm{e}_i}m_i(\theta).
\end{align*}
Since $m(0)=0$, the coefficient of $\mu_{\bm{\beta}}$, for $\bm{\beta}\in\I_s$, can be expressed by
\begin{align}
    \label{eq_A0_index}
    (\bm{A}_0)_{\bm{\alpha}\bm{\beta}} = \sum_{\{i: \alpha_i>0\}}\sum_{1\le |\bm{\nu}|\le 5-r}\alpha_ic_{i,\bm{\nu}}\One_{\{\bm{\beta}=\bm{\alpha}-\bm{e}_i+\bm{\nu}\}}.
\end{align}
Here, $\bm{\beta}=\bm{\alpha}-\bm{e}_i+\bm{\nu}$ requires $s=r-1+|\bm{\nu}|$. Then \eqref{eq_A0_index} vanishes when $s<r$. When $s=r$, it requires $|\bm{\nu}|=1$, and thus
\begin{align}
\label{eq_A0_diag_block}
    (\bm{D}_r)_{\bm{\alpha}\bm{\beta}} = \sum_{i=1}^d\sum_{j=1}^d \alpha_ih_{ij}\One_{\bm{\beta}+\bm{e}_i=\bm{\alpha}+\bm{e}_j}, \quad \bm{\alpha}, \bm{\beta}\in\I_r.
\end{align}
Consequently,
\begin{align}
    \label{eq_A0_block_mat}
    \bm{A}_0= \begin{pmatrix}
        \bm{D}_1& \bm{C}_{12} & \bm{C}_{13} & \bm{C}_{14} \\
        \bm{0} & \bm{D}_2 & \bm{C}_{23} & \bm{C}_{24} \\
        \bm{0} & \bm{0} & \bm{D}_3 & \bm{C}_{34} \\
        \bm{0} & \bm{0} & \bm{0} & \bm{D}_4
    \end{pmatrix}, \quad \bm{D}_1=\bm{H}.
\end{align}

Since $\bm{A}_0$ is a block upper triangle matrix, it suffices to show that $\bm{D}_r$ is non-singular for all $r=1,2,3,4$. For any matrix $\bm{M}=(M_{ij})_{i,j=1}^d\in\RR^{d\times d}$, define $\bm{D}_r(\bm{M})$ by~\eqref{eq_A0_diag_block} with $h_{ij}$ therein replaced by $M_{ij}$. Recall $N_r$ defined in~\eqref{eq_A0_index_set} and denote a vector of monomials by
\begin{align*}
    z_r(\theta) = (\theta^{\bm{\alpha}})_{\bm{\alpha}\in\I_r}\in\RR^{N_r}, \quad \theta\in\RR^d.
\end{align*}
For example, when $d=2$ and $r=2$, $z_2(\theta) = (\theta_1^2,\theta_1\theta_2,\theta_2^2)^{\top}$. Denote the Jacobian matrix of vector $z_r(\theta)$ by $\bm{J}_{z_r(\theta)}\in\RR^{N_r\times d}$, with $(\bm{\alpha},i)$-entry $\partial \theta^{\bm{\alpha}}/\partial \theta_i = \alpha_i\theta^{\bm{\alpha}-\bm{e}_i}$ for $\alpha_i>0$ and 0 for $\alpha_i=0$. Then, for each $\bm{\alpha}\in\I_r$,
\begin{align*}
    [\bm{J}_{z_r(\theta)}\bm{M}\theta]_{\bm{\alpha}} = \sum_{\{i: \alpha_i>0\}}\sum_{j=1}^d\alpha_i M_{ij}\theta^{\bm{\alpha}-\bm{e}_i+\bm{e}_j},
\end{align*}
which further indicates
\begin{align}
    \label{eq_Jacob_zr}
    \bm{J}_{z_r(\theta)}\bm{M}\theta = \bm{D}_r(\bm{M})z_r(\theta).
\end{align}
By taking an orthogonal diagonalization of $\bm{H}$, we obtain
\begin{align*}
    \bm{H} = \bm{Q}\bm{\Lambda} \bm{Q}^{\top}, \quad \text{where }\bm{\Lambda} = \mathrm{diag}(\lambda_1,\ldots,\lambda_d),
\end{align*}
with each $\lambda_j\ge\mu>0$ by~\eqref{eq_lower_bd_eigen_H}. Since $|\bm{\alpha}|=r$ and $[z_r(\bm{Q}\theta)]_{\bm{\alpha}} = (\bm{Q}\theta)^{\bm{\alpha}} = \prod_{i=1}^d(\sum_{j=1}^dQ_{ij}\theta_j)^{\alpha_i}$, each component of $z_r(\bm{Q}\theta)$ is a homogeneous polynomial of degree $r$. It follows that there is a unique matrix $\bm{P}_r(\bm{Q})$ satisfying
\begin{align}
    \label{eq_zr_decomp}
    z_r(\bm{Q}\theta) = \bm{P}_r(\bm{Q})z_r(\theta).
\end{align}
Moreover, since $\bm{Q}$ is orthogonal, $\theta=\bm{Q}\bm{Q}^{\top}\theta$. By applying the expression above twice, we obtain that for any $\theta\in\RR^d$, $z_r(\theta) = z_r(\bm{Q}\bm{Q}^{\top}\theta) = \bm{P}_r(\bm{Q})\bm{P}_r(\bm{Q}^{\top})z_r(\theta)$. Therefore, we have
\begin{align*}
    [\bm{I}_{N_r} - \bm{P}_r(\bm{Q})\bm{P}_r(\bm{Q}^{\top})]z_r(\theta) =0, \quad \text{for all }\theta\in\RR^d.
\end{align*}
Since the $N_r$ monomials $\{\theta^{\bm{\alpha}}:\bm{\alpha}\in \I_r\}$ are linearly independent, each row of the matrix above must have all coefficients equal to zero. Therefore, $\bm{P}_r(\bm{Q})\bm{P}_r(\bm{Q}^{\top})=\bm{I}_{N_r}$, and hence, $\bm{P}_r(\bm{Q})$ is invertible with the inverse $\bm{P}_r(\bm{Q}^{\top})$. By taking the derivative of~\eqref{eq_zr_decomp}, we obtain
\begin{align*}
    \bm{J}_{z_r(\bm{Q}\theta)}\bm{Q} = \bm{P}_r(\bm{Q})\bm{J}_{z_r(\theta)},
\end{align*}
which together with $\bm{H}\bm{Q}=\bm{Q}\bm{\Lambda}$ and~\eqref{eq_Jacob_zr} gives
\begin{align*}
    \bm{D}_r(\bm{H})\bm{P}_r(\bm{Q})z_r(\theta) & = \bm{J}_{z_r(\bm{Q}\theta)}\bm{H}\bm{Q}\theta \\
    & = \bm{J}_{z_r(\bm{Q}\theta)}\bm{Q}\bm{\Lambda} \theta \\
    & = \bm{P}_r(\bm{Q})\bm{J}_{z_r(\theta)}\bm{\Lambda} \theta \\
    & = \bm{P}_r(\bm{Q})\bm{D}_r(\bm{\Lambda})z_r(\theta), \quad \text{for all }\theta\in\RR^d.
\end{align*}
Again by the linear independence of the monomials, 
\begin{align}
    \label{eq_Dr_decomp}
    \bm{D}_r(\bm{H}) = \bm{P}_r(\bm{Q})\bm{D}_r(\bm{\Lambda})\bm{P}_r(\bm{Q}^{\top}).
\end{align}
Since $\Lambda_{ij}=\lambda_i\One_{\{i=j\}}$, it follows from expression~\eqref{eq_A0_diag_block} that
\begin{align*}
    [\bm{D}_r(\bm{\Lambda})]_{\bm{\alpha}\bm{\beta}} = \sum_{i=1}^d\alpha_i\lambda_i\One_{\{\bm{\beta}+\bm{e}_i = \bm{\alpha}+\bm{e}_i\}} = \One_{\{\bm{\alpha} = \bm{\beta}\}}\sum_{i=1}^d\alpha_i\lambda_i
\end{align*}
Thus, by~\eqref{eq_Dr_decomp} and~\eqref{eq_lower_bd_eigen_H}, we can achieve
\begin{align*}
    \mathrm{det} (\bm{D}_r) = \prod_{\bm{\alpha}\in\I_r}\Big(\sum_{i=1}^d\alpha_i\lambda_i\Big) \ge \prod_{\bm{\alpha}\in\I_r}\mu|\bm{\alpha}| = (r\mu)^{N_r} >0.
\end{align*} 
This together with~\eqref{eq_A0_block_mat} eventually yields
\begin{align*}
    \mathrm{det}(\bm{A}_0) = \prod_{r=1}^4\mathrm{det}(\bm{D}_r) >0. 
\end{align*}
Hence, $\bm{A}_0$ is non-singular and we complete the proof.
\end{proof}

\newpage
\subsection{Proofs of Theorems~\ref{Theorem_Consistency_recursive_sigma}~and~\ref{Theorem_Consistency_recursive_sigma_new}}

\begin{proof}[Proof of Theorem~\ref{Theorem_Consistency_recursive_sigma}]
W.l.o.g., we show the one-dimensional case here. The multi-dimensional case can follow the similar arguments and thus the proof is omitted. Let $\hat\sigma_n^2(\gamma)$ (resp. $\sigma_n^2(\gamma)$) be $\hat\Sigma_n(\gamma)$ (resp. $\Sigma_n(\gamma)$) with $d=1$. It is straightforward to verify that as $M \to \infty$, we have $\eta_{M + 1}/\eta_{M} \to 1$ and 
\begin{align*}
    \sum_{m = 1}^{M} (\eta_{m + 1} - \eta_{m})^{2} \asymp \eta_{M + 1}(\eta_{M + 1} - \eta_{M}). 
\end{align*}
Recall~\eqref{eq_stationary_difference} for the stationary process $\{\vartheta_{k}^{\circ}(\gamma)\}_{k \in \mathbb{Z}}$. Note that  
\begin{align*}
    \mathbb{E} \{\vartheta_{k}^{\circ}(\gamma)\} = b_{1} \theta_{\infty}^{*} (a_{1} \gamma) + b_{2} \theta_{\infty}^{*} (a_{2} \gamma) + b_{3} \theta_{\infty}^{*} (a_{3} \gamma). 
\end{align*}
For simplicity of notation, we write 
\begin{align*}
    \tilde{\theta}_{n}(\gamma) &= \hat{\theta}_{n}(\gamma) - \mathbb{E} \{\vartheta_{k}^{\circ}(\gamma)\}, \cr 
    \bar{\mathcal{S}}_{m}(\gamma) &= \sum_{k \in B_{m}} [\vartheta_{k}(\gamma) - \mathbb{E} \{\vartheta_{k}^{\circ}(\gamma)\}], \cr 
    \bar{\mathcal{R}}_{n}(\gamma) &= \sum_{k = \phi(n)}^{n} [\vartheta_{k}(\gamma) - \mathbb{E} \{\vartheta_{k}^{\circ}(\gamma)\}]. 
\end{align*}
Then, similar to~\eqref{eq_Vn_equation}, we have 
\begin{align*}
    V_{n}(\gamma) &= \Bigg(\sum_{m = 1}^{\psi(n) - 1} \bar{\mathcal{S}}_{m}(\gamma)^{\otimes 2} + \bar{\mathcal{R}}_{n}(\gamma)^{\otimes 2}\Bigg) + \Bigg(\sum_{m = 1}^{\psi(n) - 1} |B_{m}|^{2} + |\delta_{\eta}(n)|^{2}\Bigg) \tilde{\theta}_{n}(\gamma)^{\otimes 2}\cr 
    & \quad - \Bigg(\sum_{m = 1}^{\psi(n) - 1} |B_{m}| \bar{\mathcal{S}}_{m}(\gamma) + \delta_{\eta}(n) \bar{\mathcal{R}}_{n}(\gamma)\Bigg) \tilde{\theta}_{n}(\gamma)^{\top} \\
    & \quad - \tilde{\theta}_{n}(\gamma) \Bigg(\sum_{m = 1}^{\psi(n) - 1} |B_{m}| \bar{\mathcal{S}}_{m}(\gamma) + \delta_{\eta}(n) \bar{\mathcal{R}}_{n}(\gamma)\Bigg)^{\top}\cr
    &=: \bar{\mathcal{V}}_{n}(\gamma) + K_{n} \tilde{\theta}_{n}(\gamma)^{\otimes 2} - \bar{H}_{n}(\gamma) \tilde{\theta}_{n}(\gamma)^{\top} - \tilde{\theta}_{n}(\gamma) \bar{H}_{n}(\gamma)^{\top}. 
\end{align*}
Following the construction of $V_{n}(\gamma)$ in~\eqref{eq_recursive_estimator_sigma_hat}, we define 
\begin{align*}
    V_{n}^{\circ}(\gamma) = \sum_{m = 1}^{\psi(n) - 1} \Bigg|\sum_{k \in B_{m}} \mathbb{E}_{0}\{\vartheta_{k}^{\circ}(\gamma)\}\Bigg|^{2} + \Bigg|\sum_{k = \phi(n)}^{n} \mathbb{E}_{0} \{\vartheta_{k}^{\circ}(\gamma)\}\Bigg|^{2} =: \sum_{m = 1}^{\psi(n) - 1} |\mathcal{S}_{m}^{\circ}(\gamma)|^{2} + |\mathcal{R}_{n}^{\circ}(\gamma)|^{2}. 
\end{align*}
By~\eqref{eq_stationary_estimation} and the fact that $\vartheta_{k}^{\circ}(\gamma)$ satisfies GMC with $\bar{\rho} \in (0, 1)$, it follows that 
\begin{align}
\label{eq_estimation_0}
    |\mathbb{E} \{V_{n}^{\circ}(\gamma)\} - n \sigma^{2} (\gamma)| &\lesssim \sum_{m = 1}^{\psi(n) - 1} \sum_{k \in B_{m}} \sum_{|l| > k - \eta_{m}} \bar{\rho}^{|l|} + \sum_{k = \phi(n)}^{n} \sum_{|l| > k - \phi(n)} \bar{\rho}^{|l|} \cr 
    &\lesssim \sum_{m = 1}^{\psi(n) - 1} \sum_{k \in B_{m}} \bar{\rho}^{k - \eta_{m}} + \sum_{k = \phi(n)}^{n} \bar{\rho}^{k - \phi(n)} \cr 
    &\lesssim \psi(n) + 1 \asymp n^{1/\beta}.
\end{align}
By Theorem 1 in~\textcite{Xiao2011}, it follows that 
\begin{align*}
    \|V_{n}^{\circ}(\gamma) - \mathbb{E} \{V_{n}^{\circ}(\gamma)\}\|_{2} \lesssim n^{1/2} (\eta_{\psi(n)} - \eta_{\psi(n) - 1})^{1/2} \asymp n^{1 - 1/(2\beta)}. 
\end{align*}
Combining this with~\eqref{eq_estimation_0} yields 
\begin{align*}
    \mathbb{E}\left|\frac{V_{n}^{\circ}(\gamma)}{n\sigma^{2}(\gamma)} - 1\right|^{2} \lesssim n^{(2/\beta - 2)\vee (-1/\beta)}. 
\end{align*}
Then it suffices to upper bound $\mathbb{E}|V_{n}(\gamma) - V_{n}^{\circ}(\gamma)|^{2}$. To this end, note that 
\begin{align}
\label{eq_difference_V_circ}
    \|V_{n}(\gamma) - V_{n}^{\circ}(\gamma)\|_{2} \leq \|\bar{\mathcal{V}}_{n}(\gamma) - V_{n}^{\circ}(\gamma)\|_{2} + 2\|\bar{H}_{n}(\gamma) \tilde{\theta}_{n}(\gamma)\|_{2} + K_{n}\|\tilde{\theta}_{n}(\gamma)\|_{4}^{2}. 
\end{align}
By Theorem~\ref{thm_GMC}, it follows that 
\begin{align}
\label{eq_V_bound}
    \|\bar{\mathcal{V}}_{n}(\gamma) - V_{n}^{\circ}(\gamma)\|_{2} &\leq \sum_{m = 1}^{\psi(n) - 1} \|\bar{\mathcal{S}}_{m}(\gamma) - \mathcal{S}_{m}^{\circ}(\gamma)\|_{4} \|\bar{\mathcal{S}}_{m}(\gamma) + \mathcal{S}_{m}^{\circ}(\gamma)\|_{4}\cr 
    &+ \|\bar{\mathcal{R}}_{n}(\gamma) - \mathcal{R}_{n}^{\circ}(\gamma)\|_{4} \|\bar{\mathcal{R}}_{n}(\gamma) + \mathcal{R}_{n}^{\circ}(\gamma)\|_{4}\cr 
    &\lesssim \sum_{m = 1}^{\psi(n) - 1} \bar{\rho}^{\eta_{m}} (\eta_{m + 1} - \eta_{m})^{1/2} + \bar{\rho}^{\phi(n)}(n - \phi(n) + 1)^{1/2} \leq C. 
\end{align}
Similarly, we have 
\begin{equation*}
    \|\tilde{\theta}_{n}(\gamma)\|_{4}^{2} \lesssim \left\|\frac{1}{n} \sum_{k = 1}^{n} \mathbb{E}_{0}\{\vartheta_{k}^{\circ}(\gamma)\}\right\|_{4}^{2} + \left\|\frac{1}{n} \sum_{k = 1}^{n} \{\vartheta_{k}(\gamma) - \vartheta_{k}^{\circ}(\gamma)\}\right\|_{4}^{2} \asymp \frac{1}{n}, 
\end{equation*}
and 
\begin{align}
\label{eq_H_bound}
    \|\bar{H}_{n}(\gamma)\|_{4} &\leq \sum_{m = 1}^{\psi(n) - 1} |B_{m}| \|\bar{\mathcal{S}}_{m}(\gamma) - \mathcal{S}_{m}^{\circ} (\gamma)\|_{4} + \delta_{\eta}(n) \|\bar{\mathcal{R}}_{n}(\gamma) - \mathcal{R}_{n}^{\circ}(\gamma)\|_{4} \cr 
    &+ \left\|\sum_{m = 1}^{\psi(n) - 1} |B_{m}| \mathcal{S}_{m}^{\circ}(\gamma) + \delta_{\eta}(n) \mathcal{R}_{n}^{\circ}(\gamma)\right\|_{4} \lesssim n^{3/2 - 1/\beta}.    
\end{align}
Combining these with the fact that $K_{n} \asymp n^{2-1/\beta}$ yields $K_{n} \|\tilde{\theta}_{n}(\gamma)\|_{4}^{2} \asymp n^{1 - 1/\beta}$ and $\|\bar{H}_{n}(\gamma) \tilde{\theta}_{n}(\gamma)\|_{2} \lesssim n^{1 - 1/\beta}$. Putting all these pieces together, we obtain $\|V_{n}(\gamma) - V_{n}^{\circ}(\gamma)\|_{2} \lesssim n^{1 - 1/\beta}$ in view of~\eqref{eq_difference_V_circ}. Consequently, it follows that 
\begin{align*}
    \mathbb{E} \left|\frac{\hat{\sigma}_{n}^{2}(\gamma)}{\sigma^{2}(\gamma)} - 1\right|^{2} \lesssim \mathbb{E}\left|\frac{V_{n}^{\circ}(\gamma)}{n\sigma^{2}(\gamma)} - 1\right|^{2} + \mathbb{E} \left|\frac{V_{n}^{\circ}(\gamma) - V_{n}(\gamma)}{n \sigma^{2}(\gamma)}\right|^{2} \lesssim n^{(2/\beta - 2) \vee (-1/\beta)}. 
\end{align*}
\end{proof}

\begin{proof}[Proof of Theorem~\ref{Theorem_Consistency_recursive_sigma_new}]
For simplicity of notation, we write  
\begin{align*}
    \bar{\mathcal{S}}_{m}^{\tau}(\gamma) = \sum_{k = \eta_{m}}^{\eta_{m} + \tau(m) - 1} [\vartheta_{k}(\gamma) - \mathbb{E} \{\vartheta_{k}^{\circ} (\gamma)\}].  
\end{align*}
Similar to~\eqref{eq_estimation_Vn_0} and~\eqref{eq_estimatino_Vn_00}, it follows that 
\begin{align*}
    U_{n}(\gamma) = \bar{\mathcal{U}}_{n}(\gamma) + \mathcal{K}_{n} \tilde{\theta}_{n}(\gamma)^{\otimes 2} - \bar{\mathcal{H}}_{n}(\gamma) \tilde{\theta}_{n}(\gamma)^{\top} - \tilde{\theta}_{n}(\gamma) \bar{\mathcal{H}}_{n}(\gamma)^{\top}, 
\end{align*}
where
\begin{align*}
    \bar{\mathcal{U}}_{n}(\gamma) &= \sum_{k = 1}^{n} \Bigg(\mathbb{E}_{0}\{\vartheta_{k}(\gamma)\}^{\otimes 2} + \sum_{\ell = \phi(k)}^{k - 1} \mathbb{E}_{0}\{\vartheta_{k}(\gamma)\} \mathbb{E}_{0}\{\vartheta_{\ell}(\gamma)\}^{\top} \cr 
    &\quad + \sum_{\ell = \phi(k)}^{k - 1} \mathbb{E}_{0}\{\vartheta_{\ell}(\gamma)\} \mathbb{E}_{0}\{\vartheta_{k}(\gamma)\}^{\top}\Bigg) \times \mathbb{I}\{E_{k}(\tau)\}, \cr   
    \bar{\mathcal{H}}_{n}(\gamma) &= \sum_{m = 1}^{\psi(n) - 1} \{|B_{m}| \bar{\mathcal{S}}_{m}(\gamma) - \tau(m) \bar{\mathcal{S}}_{m}^{\tau}(\gamma)\} \\
    & \quad + \{\delta_{\eta}(n) \bar{\mathcal{R}}_{n}(\gamma) - \tau(\psi(n)) \bar{\mathcal{S}}_{\psi(n)}^{\tau}(\gamma)\} \times \mathbb{I}\{E_{n}(\tau)\}. 
\end{align*}

\begin{itemize}
    \item[(1)] When $\eta_{m} = \lfloor c_{1} m^{\beta} \rfloor$ and $\tau(m) = \lfloor c_{2} \log m \rfloor$ for some $c_{1} > 0$, $\beta > 1$ and $c_{2} > 0$, by~\eqref{eq_stationary_estimation} and the fact that $\vartheta_{k}^{\circ}(\gamma)$ satisfies GMC with $\bar{\rho} \in (0, 1)$, it follows that 
    \begin{align}
    \label{eq_E_Un_0}
        |\mathbb{E} \{U_{n}^{\circ} (\gamma)\} &- \kappa(n) \sigma^{2} (\gamma)| \leq \sum_{k = 1}^{n} |\mathbb{E} \{Q_{k}^{\circ} (\gamma)\} - \sigma^{2} (\gamma)| \times \mathbb{I} \{E_{k}(\tau)\} \cr 
        &\lesssim \sum_{m = 1}^{\psi(n) - 1} \sum_{k = \eta_{m} + \tau(m)}^{\eta_{m + 1} - 1} \sum_{|l| > k - \eta_{m}} \bar{\rho}^{|l|} + \sum_{k = \phi(n)}^{n} \sum_{|l| > k - \phi(n)} \bar{\rho}^{|l|} \times \mathbb{I}\{E_{n}(\tau)\} \cr 
        &\lesssim \sum_{m = 1}^{\psi(n) - 1} \bar{\rho}^{\tau(m)} + \sum_{k = \phi(n)}^{n} \bar{\rho}^{k - \phi(n)} \lesssim \max\left\{n^{(c_{2} \log (\bar{\rho}) + 1)/\beta}, \log n\right\}. 
    \end{align}
    By Theorem 1 in~\textcite{Xiao2011}, 
    \begin{align*}
        \|U_{n}^{\circ} (\gamma) - \mathbb{E} \{U_{n}^{\circ} (\gamma)\}\|_{2} \lesssim n^{1/2} (\eta_{\psi(n)} - \eta_{\psi(n) - 1})^{1/2} \asymp n^{1 - 1/(2\beta)}. 
    \end{align*}
    Consequently, since $2 c_{2} \log (\bar{\rho}) \leq 2 \beta - 3$, we have 
    \begin{align}
    \label{eq_MSE_Un_0}
        \mathbb{E} \left|\frac{U_{n}^{\circ} (\gamma)}{\kappa(n)} - \sigma^{2}(\gamma)\right|^{2} \lesssim n^{-1/\beta}.  
    \end{align} 
    Then, it suffices to derive an upper bound for $\|U_{n}^{\circ} (\gamma) - U_{n} (\gamma)\|_{2}$. Similar to~\eqref{eq_V_bound} and~\eqref{eq_H_bound}, it follows from Theorem~\ref{thm_GMC} that $\|U_{n}^{\circ}(\gamma) - \bar{\mathcal{U}}_{n}(\gamma)\|_{2} \leq C$ and $\|\bar{\mathcal{H}}_{n}(\gamma)\|_{4} \lesssim n^{3/2 - 1/\beta}$. Then 
    \begin{align*}
        \|U_{n}^{\circ} (\gamma) - U_{n} (\gamma)\|_{2} \lesssim n^{1 - 1/\beta}.  
    \end{align*}
    Combining this with~\eqref{eq_MSE_Un_0} yields 
    \begin{align*}
        \mathbb{E} \left|\frac{U_{n} (\gamma)}{\kappa(n)} - \sigma^{2}(\gamma)\right|^{2} \lesssim n^{-1/\beta}.  
    \end{align*}

    \item[(2)] When $\eta_{m} = \lfloor \bar{c}_{1} m \log m \rfloor$ and $\tau(m) = \lfloor \bar{c}_{2} \log m \rfloor$ for some $\bar{c}_{1} > \bar{c}_{2} > 0$ with $2 \bar{c}_{2} \log (\bar{\rho}) \leq -1$, by a similar argument to that in~\eqref{eq_E_Un_0}, we have  
    \begin{align*}
        |\mathbb{E} \{U_{n}^{\circ} (\gamma)\} - \kappa(n) \sigma^{2} (\gamma)| \lesssim n^{1/2}. 
    \end{align*}
    By Theorem 1 in~\textcite{Xiao2011}, 
    \begin{align*}
        \|U_{n}^{\circ} (\gamma) - \mathbb{E} \{U_{n}^{\circ} (\gamma)\}\|_{2} \lesssim n^{1/2} (\eta_{\psi(n)} - \eta_{\psi(n) - 1})^{1/2}\lesssim (n \log n)^{1/2}.  
    \end{align*}
    Since $\eta_{m} = \lfloor \bar{c}_{1} m \log m \rfloor$, elementary calculations yield $\mathcal{K}_{n} \lesssim n \log n$ and $\|\bar{\mathcal{H}}_{n}(\gamma)\|_{4} \lesssim n^{1/2} \log n$. Consequently, we obtain $\|U_{n}(\gamma) - U_{n}^{\circ}(\gamma)\|_{2} \lesssim \log n$ and 
    \begin{align*}
        \mathbb{E} \left|\frac{U_{n} (\gamma)}{\kappa(n)} - \sigma^{2}(\gamma)\right|^{2} \lesssim \frac{\log n}{n}. 
    \end{align*}    
\end{itemize}
\end{proof}

\clearpage

\section{Additional Information of Simulation Studies}\label{sec_more_simu}

\subsection{Improved ASGD Estimates by Multiple Step Sizes}

We consider the SGD procedure with three different constant step sizes: $\gamma=0.0125$, 0.025 and 0.05. For simplicity, we denote respectively the extrapolated estimators using two and three different step sizes by
$$\hat{\theta}_n^{(2)}(\gamma) = 2\bar\theta_n(\gamma) - \bar\theta_n(2\gamma),$$
and
$$\hat{\theta}_n^{(3)}(\gamma) = \frac{8}{3}\bar\theta_n(\gamma) - 2\bar\theta_n(2\gamma) + \frac{1}{3}\bar\theta_n(4\gamma).$$

\begin{table}[!hbtp]
    \centering
    \caption{Logistic regression. Absolute value of the average bias over 100 independent runs, where the target is penalized population minimizer.}
    \label{table_linear}
    \begin{tabular}{c|c|c|c|c|c|c|c}
      \hline\hline
      step size & estimator & $n=100$ & $n=1000$ & $n=5000$ & $n=10^4$ & $n=10^5$ & $n=10^6$ \\
      \hline
        \multirow{3}{5em}{$\gamma=0.025$} & $\bar\theta_n(\gamma)$ & 1.942e-01 & 1.782e-02 & 4.520e-03 & 1.923e-03 & 1.823e-04 & 9.154e-05 \\
         & $\hat{\theta}_n^{(2)}(\gamma)$ & 1.375e-01 & 1.462e-02 & 2.915e-03 & 1.355e-03 & 6.670e-05 & \textbf{8.489e-02} \\
         & $\hat{\theta}_n^{(3)}(\gamma)$ & 8.638e-02 & 8.363e-03 & 2.823e-03 & 9.776e-04 & 2.470e-05 & \textbf{1.725e-02} \\
      \hline
        \multirow{3}{4em}{$\gamma=0.05$} & $\bar\theta_n(\gamma)$ & 1.145e-01 & 1.911e-02 & 5.352e-03 & 2.632e-03 & 5.348e-04 & 1.327e-04 \\
         & $\hat{\theta}_n^{(2)}(\gamma)$ & 8.051e-02 & 7.328e-03 & 3.207e-03 & 1.066e-03 & 3.083e-04 & \textbf{8.604e-05} \\
         & $\hat{\theta}_n^{(3)}(\gamma)$ & 5.274e-02 & 4.713e-03 & 2.490e-03 & 8.798e-04 & 1.410e-04 & \textbf{3.584e-02} \\ 
      \hline\hline
    \end{tabular}
\end{table}

\subsection{Comparison of Two Online Estimators of Long-Run Covariance}
\begin{figure}[!tbhp]
        \centering
        \includegraphics[width=0.7\textwidth]{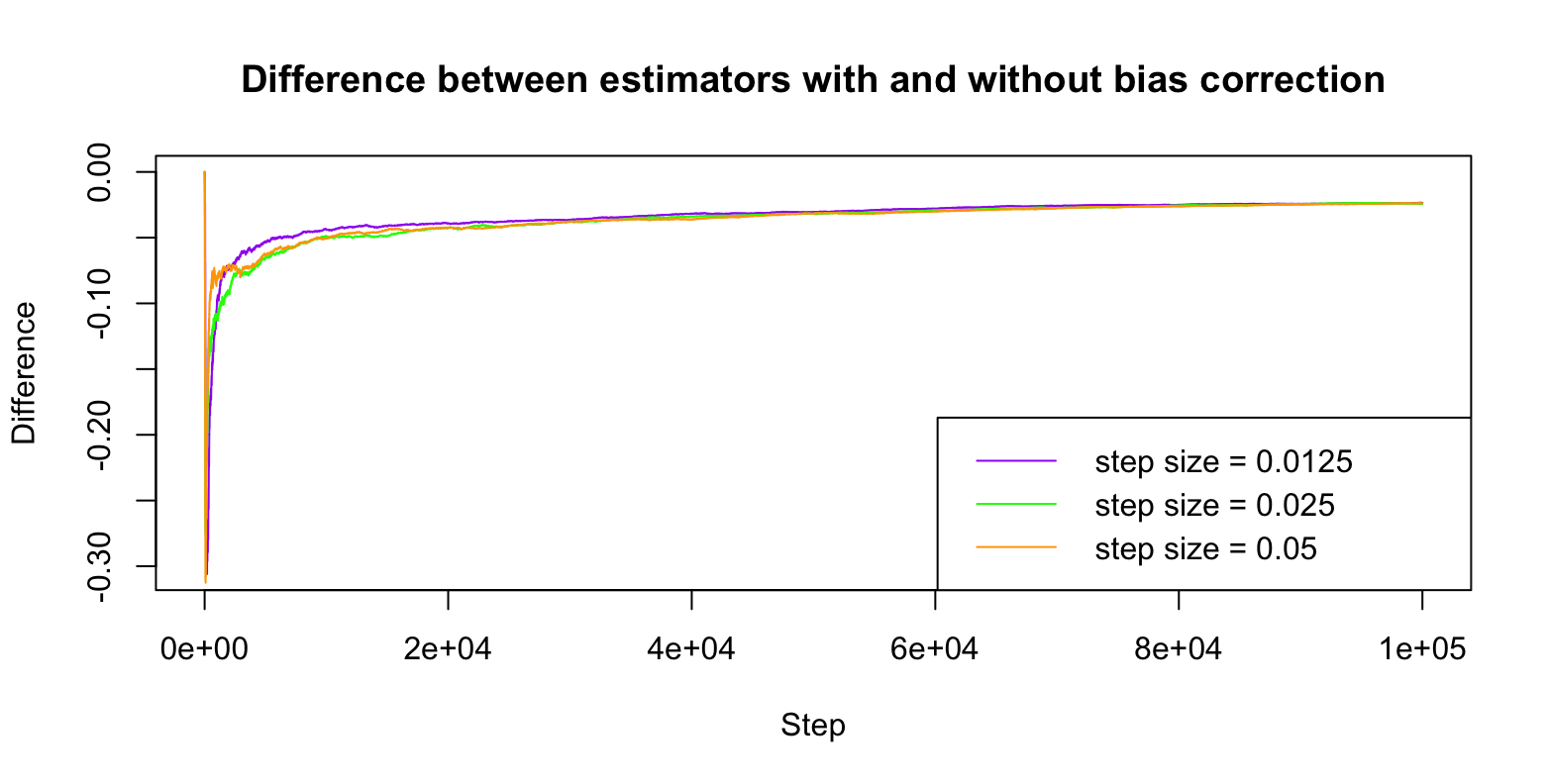}
    \caption{Difference between the estimators with and without bias correction, i.e., $\hat\sigma-\tilde\sigma$. The accuracy of the bias-reduced estimator outperforms the non-debiased one regardless of different step sizes.}
    \label{fig_diff}
\end{figure}

We report the convergence trace of our proposed online estimators with different constant step sizes. In Figure \ref{fig_diff}, we show the difference between the bias of the non-debiased and the bias-reduced estimators. One can observe that our bias reduction works robustly across different step sizes. 

\begin{figure}[!tb]
    \centering
    
    \begin{subfigure}{0.45\textwidth}
        \centering
        \includegraphics[width=1\linewidth]{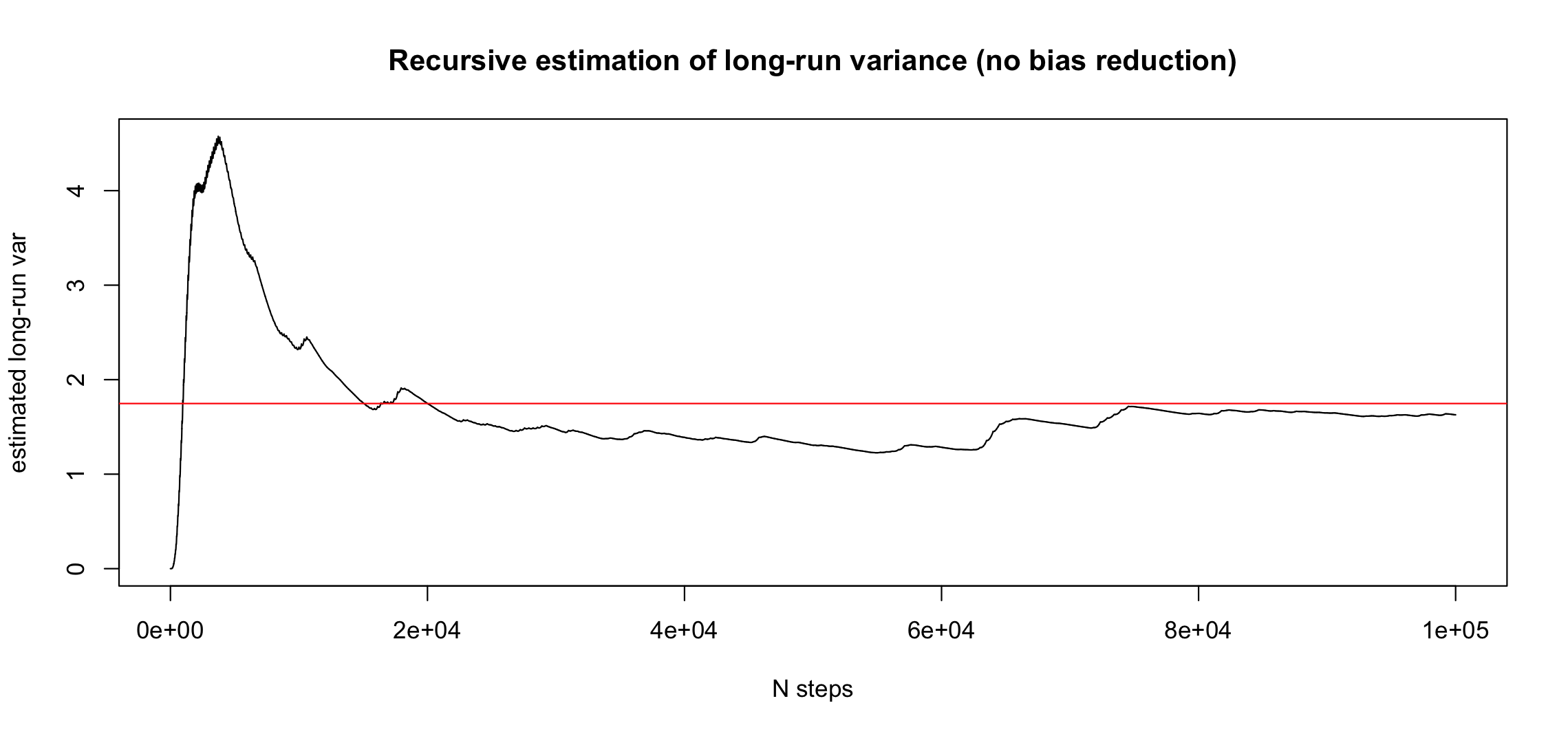}  
        \caption{Linear regression, $\gamma=0.01$}
    \end{subfigure}
    \begin{subfigure}{0.45\textwidth}
        \centering
        \includegraphics[width=1\linewidth]{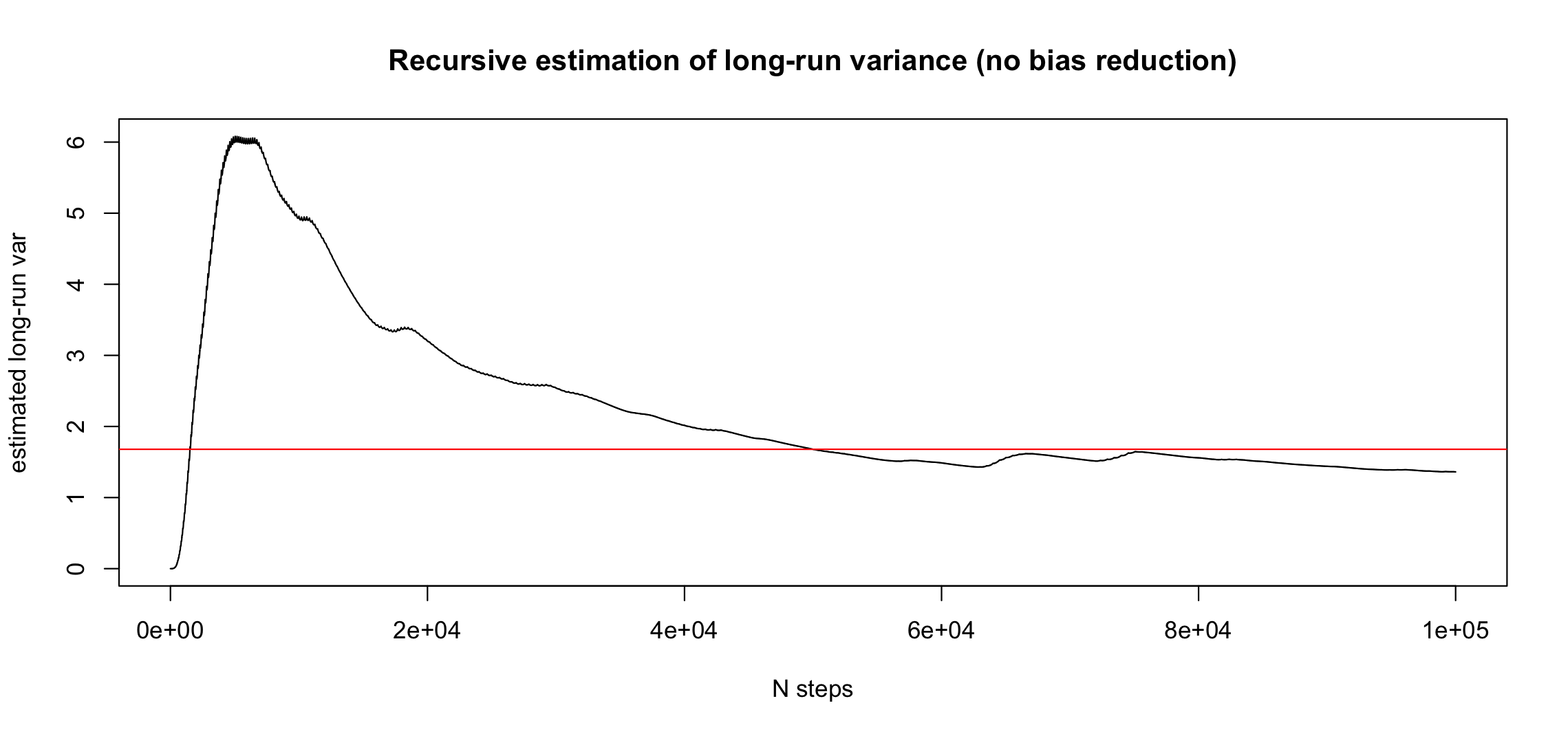}  
        \caption{Linear regression, $\gamma=0.005$}
    \end{subfigure}
        \caption{Recursive estimation of the long-run variance \textbf{without} bias reduction. The red line is the oracle value with $n=0.1$ billion.}
    \label{fig:linear_1d_online}
\end{figure}

\begin{figure}[tb!]
    \centering
    
    \begin{subfigure}{0.45\textwidth}
        \centering
        \includegraphics[width=1\linewidth]{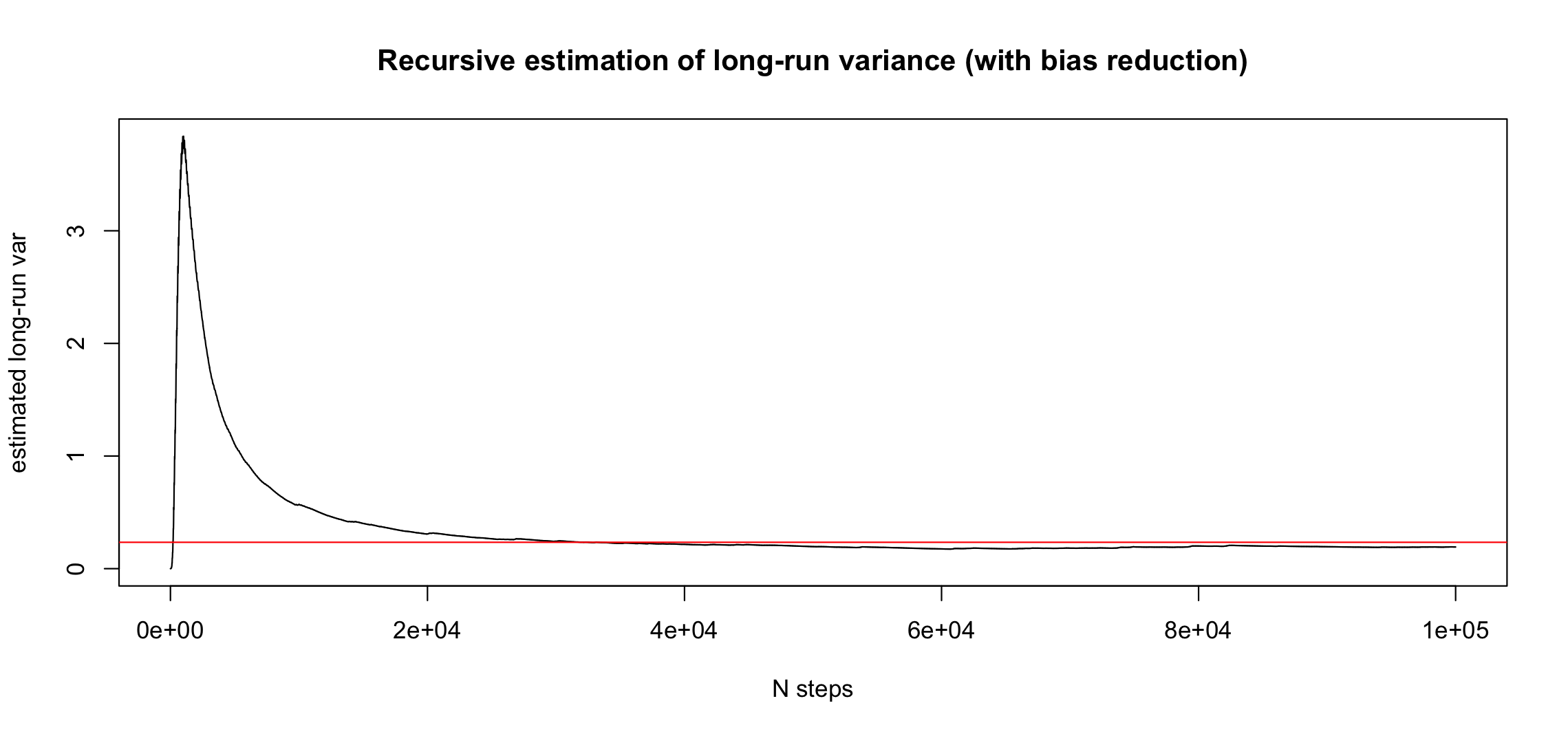}  
        \caption{Linear regression, $\gamma=0.01$}
    \end{subfigure}
    \begin{subfigure}{0.45\textwidth}
        \centering
        \includegraphics[width=1\linewidth]{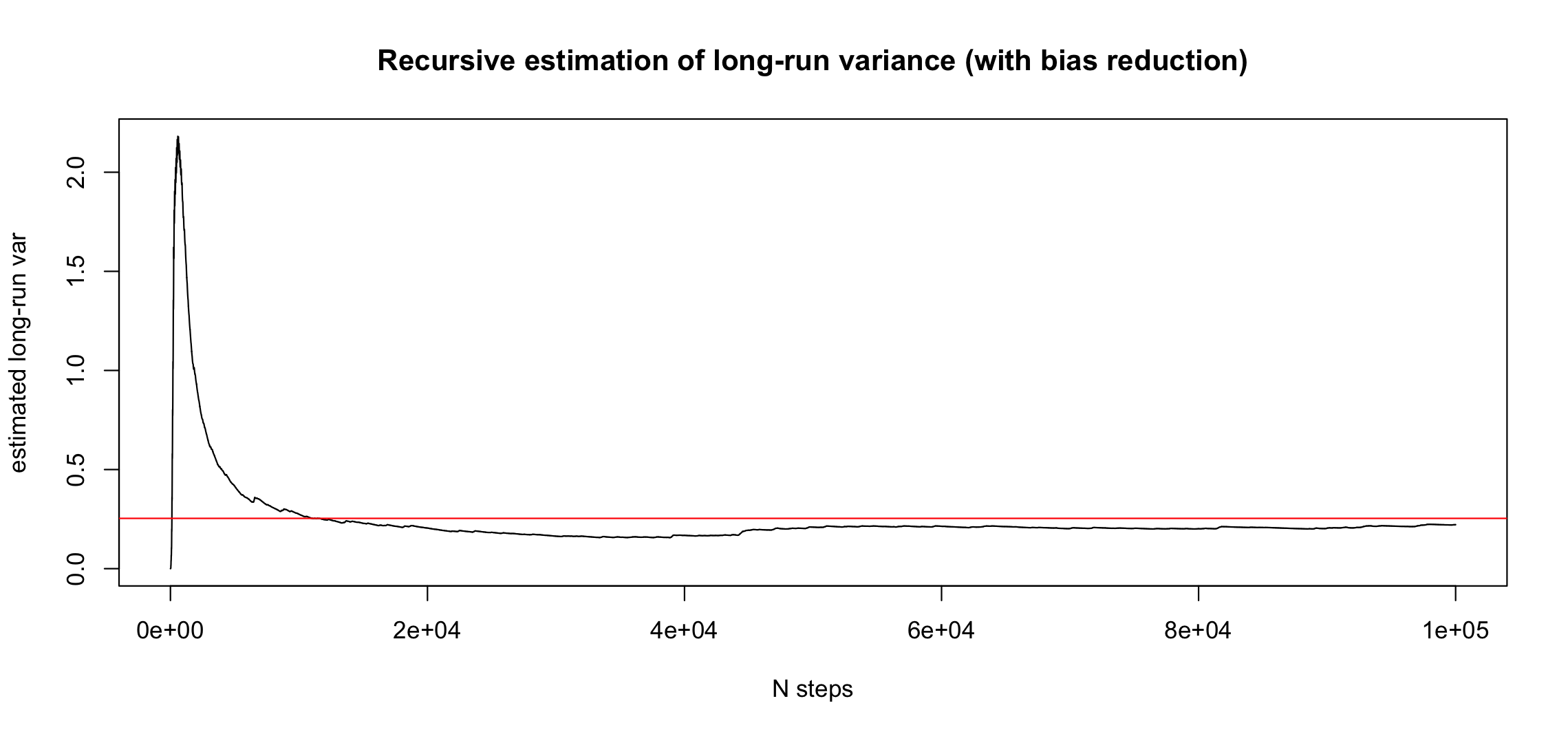}  
        \caption{Linear regression, $\gamma=0.005$}
    \end{subfigure}
        \caption{Recursive estimation of the long-run variance \textbf{with} bias reduction. The red line is the oracle value with $n=0.1$ billion.}
    \label{fig:linear_1d_online_bias_reduct}
\end{figure}

\begin{figure}[tb!]
    \centering
    \begin{subfigure}{0.45\textwidth}
        \centering
        \includegraphics[width=1\linewidth]{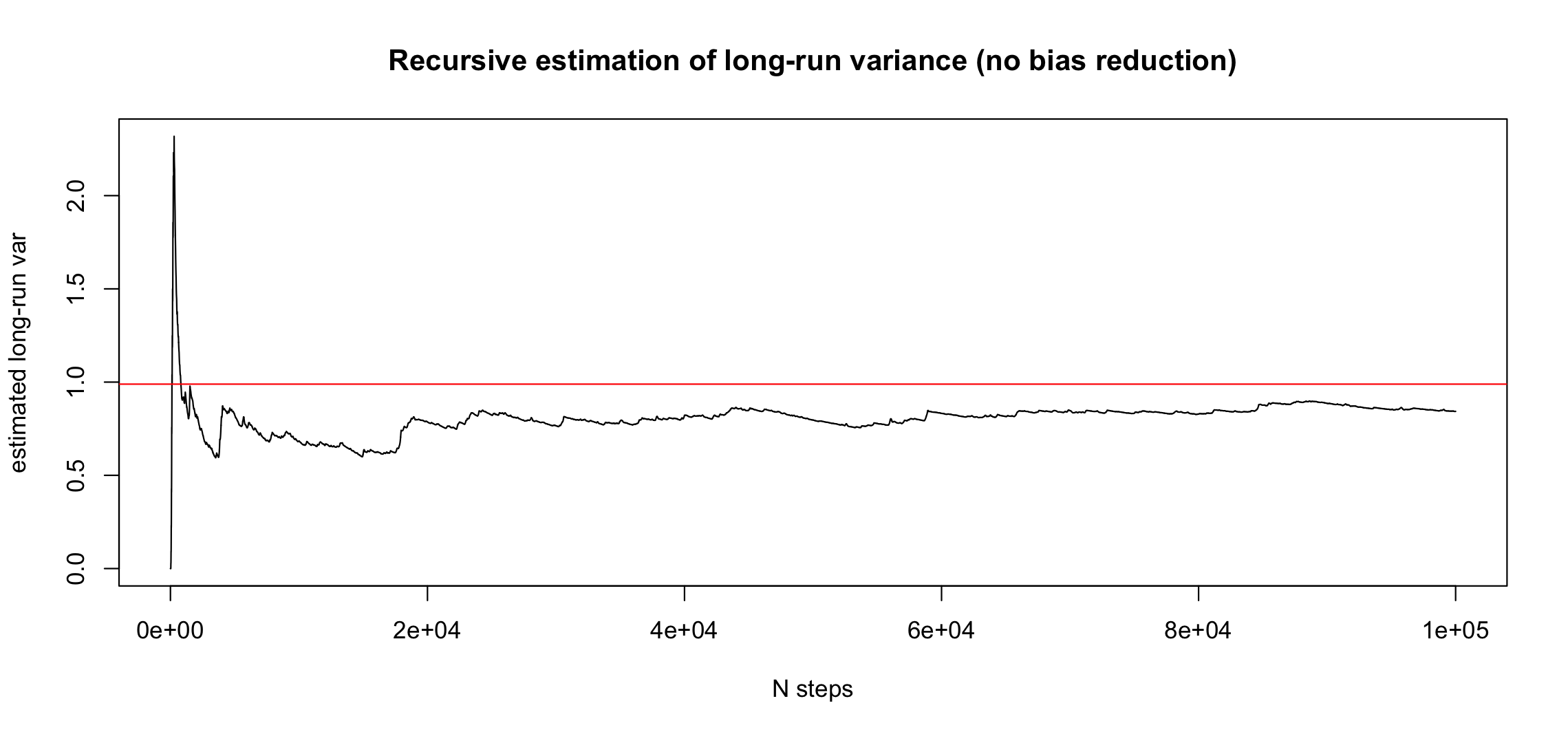}  
        \caption{Logistic regression, $\gamma=0.01$}
    \end{subfigure}
    \begin{subfigure}{0.45\textwidth}
        \centering
        \includegraphics[width=1\linewidth]{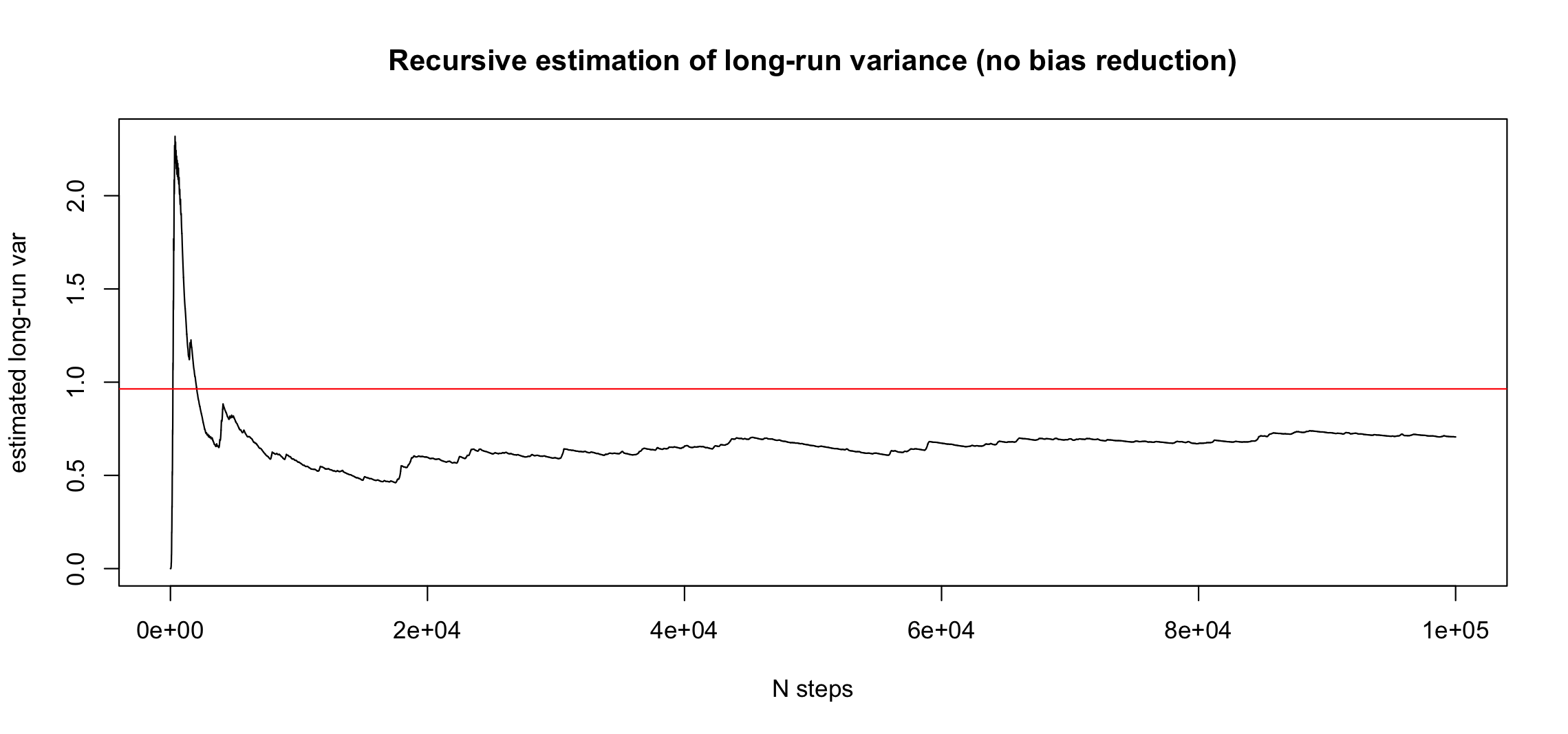}  
        \caption{Logistic regression, $\gamma=0.005$}
    \end{subfigure}

    \caption{Recursive estimation of the long-run variance \textbf{without} bias reduction. The red line is the oracle value with $n=0.1$ billion.}
    \label{fig:logi_1d_online}
\end{figure}

\begin{figure}[tb!]
    \centering
    \begin{subfigure}{0.45\textwidth}
        \centering
        \includegraphics[width=1\linewidth]{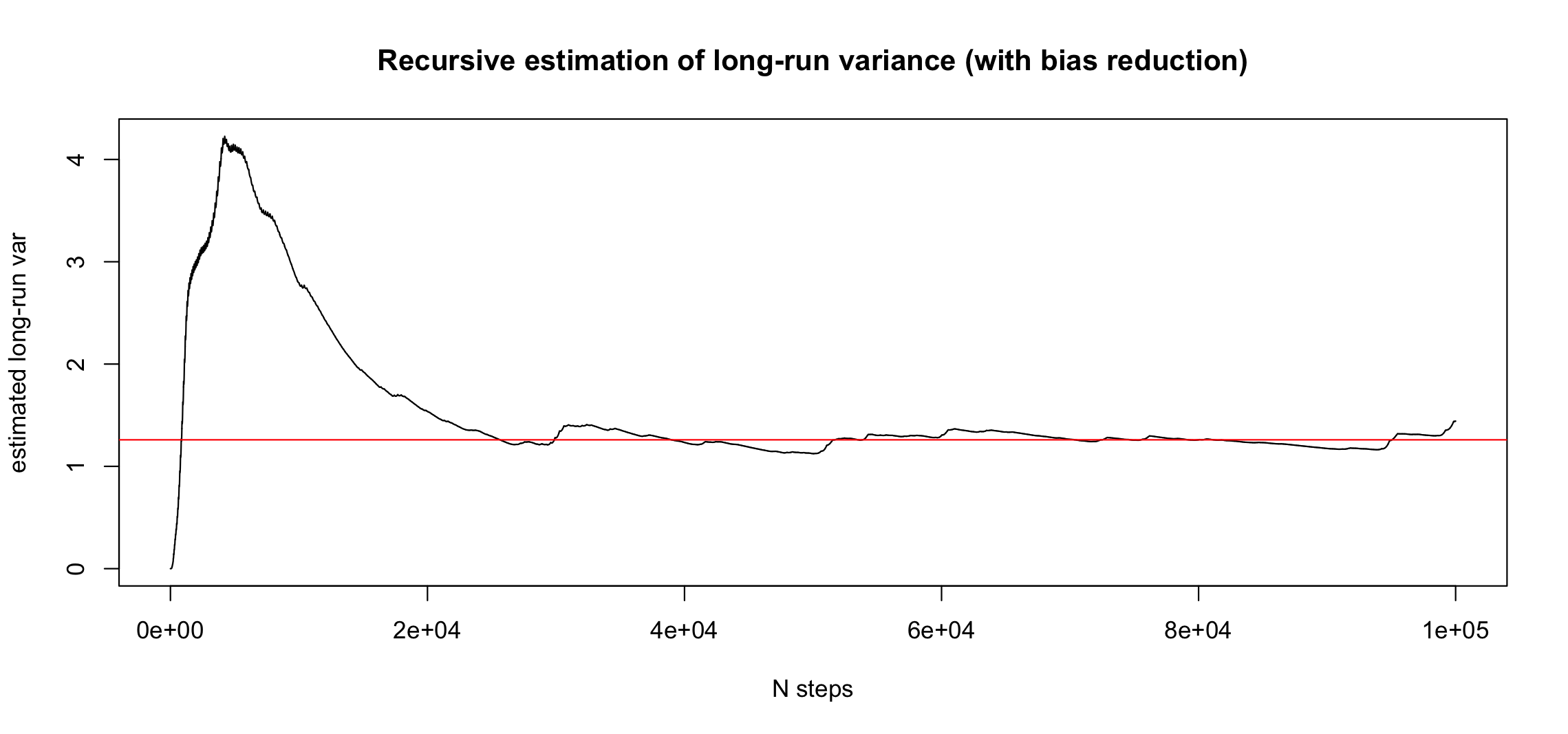}  
        \caption{Logistic regression, $\gamma=0.01$}
    \end{subfigure}
    \begin{subfigure}{0.45\textwidth}
        \centering
        \includegraphics[width=1\linewidth]{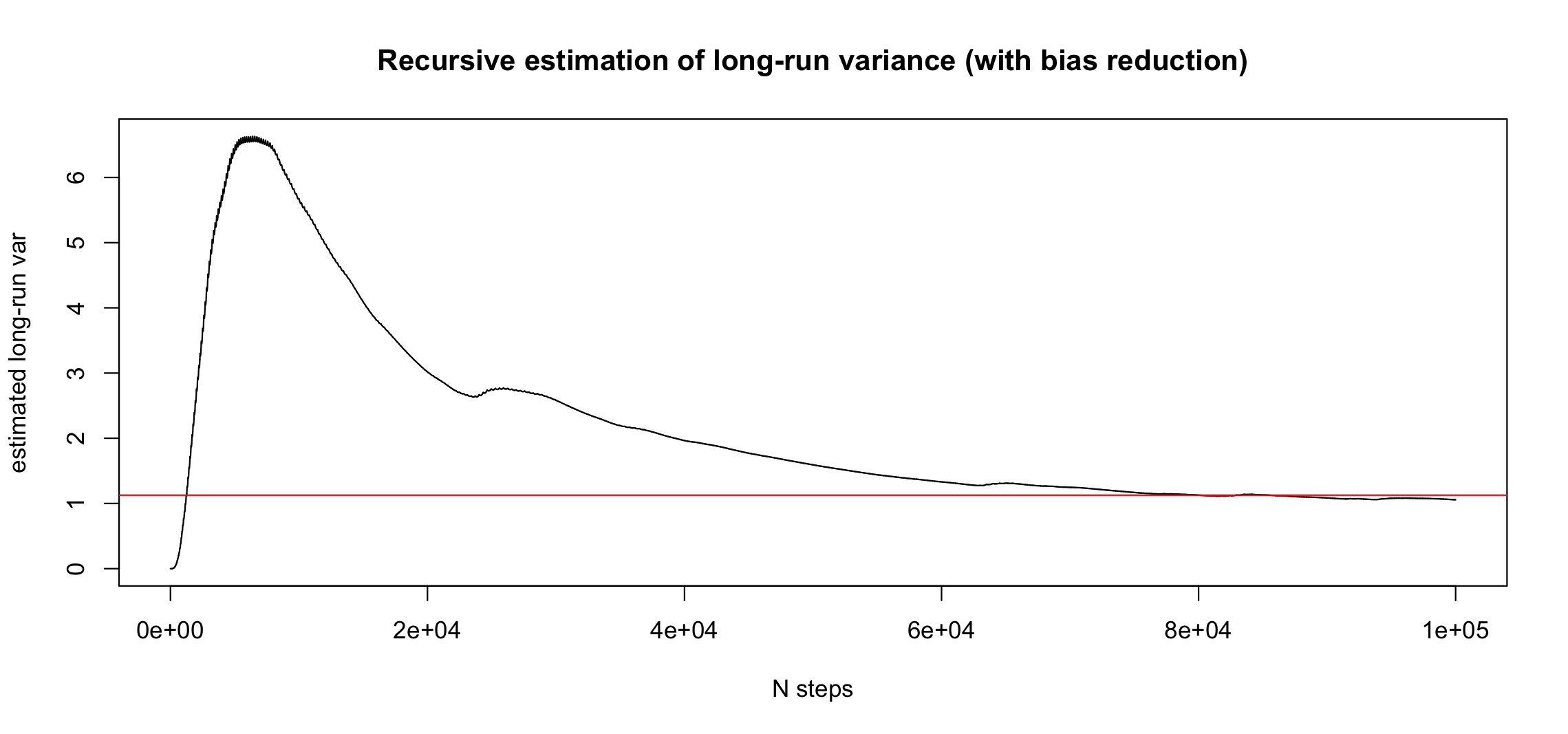}  
        \caption{Logistic regression, $\gamma=0.005$}
    \end{subfigure}

    \caption{Recursive estimation of the long-run variance \textbf{with} bias reduction. The red line is the oracle value with $n=0.1$ billion.}
    \label{fig:logi_1d_online_bias_reduct}
\end{figure}

Further, we report more cases of the estimated long-run variances in both linear regression and logistic regression. Since we do not have a closed-form solution for the true long-run variance of the logistic regression, we simply choose the estimated long-run but with an extremely large iteration step number, i.e., 0.1 billion, as the baseline. As shown in Figure \ref{fig:linear_1d_online}--\ref{fig:logi_1d_online_bias_reduct}, our two proposed online estimators can converge to the true long-run as the step number increases.

\clearpage

\section{Additional References}
\label{sec_additional_literature}

The classical theory of stochastic approximation provides foundations for studying convergence of recursive estimators. Early convergence results include \textcite{blum_approximation_1954,dvoretzky_stochastic_1956}, while \textcite{sacks_asymptotic_1958} studies asymptotic distributions. The almost-supermartingale convergence theorem of \textcite{robbins_convergence_1971} provides a general tool for establishing almost sure convergence. \textcite{ljung_analysis_1977} relates the asymptotic behavior of recursive stochastic algorithms to an associated ordinary differential equation. For a broader historical overview, see \textcite{lai_stochastic_2003}.

Subsequent developments address computational constraints, applications and more general noise distributions. \textcite{wang_stochastic_2010} study stochastic resource allocation over fading communication channels, and \textcite{gandikota_vqsgd_2022} develop vector quantization schemes that reduce communication costs in distributed SGD with convergence guarantees. \textcite{lou_beyond_2022} establish concentration bounds for averaged SGD in linear models under finite-moment conditions, allowing noise beyond the sub-Gaussian setting. These developments illustrate the importance of accounting for both the implementation of stochastic algorithms and the distribution of their noise.

A complementary approach studies stochastic recursions through iterated random functions. \textcite{dubins_invariant_1966} investigate invariant probabilities for a class of Markov processes, while \textcite{barnsley_iterated_1985,barnsley_recurrent_1989} develop iterated function systems and their recurrent extensions. \textcite{stenflo_survey_2012} surveys systems that are contractive on average. Beyond the existence of stationary distributions, \textcite{mirek_heavy_2011} studies heavy-tailed stationary laws and stable limits for sums generated by iterated Lipschitz maps. This perspective motivates our treatment of constant step-size SGD as a stochastic recursion, with GMC providing quantitative control of its dependence on
initialization.

\printbibliography[title={References}]

@book{wright_optimization_2022,
	address = {Cambridge},
	author = {Wright, Stephen J. and Recht, Benjamin},
	publisher = {Cambridge University Press},
	title = {Optimization for {Data} {Analysis}},
	year = {2022}}

@book{beck_first-order_2017,
	author = {Beck, Amir},
	month = oct,
	publisher = {Society for Industrial and Applied Mathematics},
	series = {{MOS}-{SIAM} {Series} on {Optimization}},
	title = {First-{Order} {Methods} in {Optimization}},
	year = {2017}}

@inproceedings{pedregosa_hyperparameter_2016,
	author = {Pedregosa, Fabian},
	booktitle = {Proceedings of {The} 33rd {International} {Conference} on {Machine} {Learning}},
	month = jun,
	pages = {737--746},
	publisher = {PMLR},
	title = {Hyperparameter optimization with approximate gradient},
	year = {2016}}

@inproceedings{maclaurin_gradient-based_2015,
	author = {Maclaurin, Dougal and Duvenaud, David and Adams, Ryan},
	booktitle = {Proceedings of the 32nd {International} {Conference} on {Machine} {Learning}},
	month = jun,
	pages = {2113--2122},
	publisher = {PMLR},
	title = {Gradient-based {Hyperparameter} {Optimization} through {Reversible} {Learning}},
	year = {2015}}

@inproceedings{franceschi_forward_2017,
	author = {Franceschi, Luca and Donini, Michele and Frasconi, Paolo and Pontil, Massimiliano},
	booktitle = {Proceedings of the 34th {International} {Conference} on {Machine} {Learning}},
	month = jul,
	pages = {1165--1173},
	publisher = {PMLR},
	title = {Forward and {Reverse} {Gradient}-{Based} {Hyperparameter} {Optimization}},
	year = {2017}}

@inproceedings{mehmood_automatic_2019,
	author = {Mehmood, Sheheryar and Ochs, Peter},
	month = oct,
	title = {Automatic {Differentiation} of {Some} {First}-{Order} {Methods} in {Parametric} {Optimization}},
    booktitle = {International Conference on Artificial Intelligence and Statistics},
	year = {2019}}

@article{ehrhardt_analyzing_2024,
	author = {Ehrhardt, Matthias J and Roberts, Lindon},
	journal = {IMA Journal of Applied Mathematics},
	month = jan,
	number = {1},
	pages = {254--278},
	title = {Analyzing inexact hypergradients for bilevel learning},
	volume = {89},
	year = {2024}}

@article{sirignano_stochastic_2020,
	author = {Sirignano, Justin and Spiliopoulos, Konstantinos},
	journal = {Stochastic Systems},
	month = jun,
	number = {2},
	pages = {124--151},
	title = {Stochastic {Gradient} {Descent} in {Continuous} {Time}: {A} {Central} {Limit} {Theorem}},
	volume = {10},
	year = {2020}}

@article{li_stochastic_2019,
	author = {Li, Qianxiao and Tai, Cheng and E, Weinan},
	journal = {Journal of Machine Learning Research},
	number = {40},
	pages = {1--47},
	title = {Stochastic {Modified} {Equations} and {Dynamics} of {Stochastic} {Gradient} {Algorithms} {I}: {Mathematical} {Foundations}},
	volume = {20},
	year = {2019}}

@inproceedings{li_stochastic_2017,
	author = {Li, Qianxiao and Tai, Cheng and E, Weinan},
	booktitle = {Proceedings of the 34th {International} {Conference} on {Machine} {Learning}},
	month = jul,
	pages = {2101--2110},
	publisher = {PMLR},
	title = {Stochastic {Modified} {Equations} and {Adaptive} {Stochastic} {Gradient} {Algorithms}},
	year = {2017}}

@article{flamand_functional_2026,
	author = {Flamand, Kessang and Brunel, Victor-Emmanuel},
	month = feb,
	title = {Functional {Central} {Limit} {Theorem} for {Stochastic} {Gradient} {Descent}},
    journal={arXiv preprint 	arXiv:2602.15538},
	year = {2026}}

@article{feng_uniform--time_2020,
	author = {Feng, Yuanyuan and Gao, Tingran and Li, Lei and Liu, Jian-Guo and Lu, Yulong},
	journal = {Communications in Mathematical Sciences},
	number = {1},
	pages = {163--188},
	title = {Uniform-in-time weak error analysis for stochastic gradient descent algorithms via diffusion approximation},
	volume = {18},
	year = {2020}}

@article{abdulle_high_2014,
	author = {Abdulle, Assyr and Vilmart, Gilles and Zygalakis, Konstantinos C.},
	journal = {SIAM Journal on Numerical Analysis},
	month = jan,
	number = {4},
	pages = {1600--1622},
	title = {High {Order} {Numerical} {Approximation} of the {Invariant} {Measure} of {Ergodic} {SDEs}},
	volume = {52},
	year = {2014}}

@inproceedings{wu_how_2018,
	author = {Wu, Lei and Ma, Chao and E, Weinan},
	booktitle = {Advances in {Neural} {Information} {Processing} {Systems}},
	publisher = {Curran Associates, Inc.},
	title = {How {SGD} {Selects} the {Global} {Minima} in {Over}-parameterized {Learning}: {A} {Dynamical} {Stability} {Perspective}},
	volume = {31},
	year = {2018}}

@article{bottou_optimization_2018,
	author = {Bottou, L{\'e}on and Curtis, Frank E. and Nocedal, Jorge},
	journal = {SIAM Review},
	month = jan,
	number = {2},
	pages = {223--311},
	title = {Optimization {Methods} for {Large}-{Scale} {Machine} {Learning}},
	volume = {60},
	year = {2018}}

@article{stenflo_survey_2012,
	author = {Stenflo, {\"O}rjan},
	journal = {Journal of Difference Equations and Applications},
	number = {8},
	pages = {1355--1380},
	title = {A survey of average contractive iterated function systems},
	volume = {18},
	year = {2012}}

@inproceedings{hodgkinson_multiplicative_2021,
	author = {Hodgkinson, Liam and Mahoney, Michael},
	booktitle = {Proceedings of the 38th {International} {Conference} on {Machine} {Learning}},
	pages = {4262--4274},
	publisher = {PMLR},
	title = {Multiplicative {Noise} and {Heavy} {Tails} in {Stochastic} {Optimization}},
	year = {2021}}

@article{dubins_invariant_1966,
	author = {Dubins, Lester E. and Freedman, David A.},
	journal = {The Annals of Mathematical Statistics},
	number = {4},
	pages = {837--848},
	title = {Invariant {Probabilities} for {Certain} {Markov} {Processes}},
	volume = {37},
	year = {1966}}

@article{barnsley_recurrent_1989,
	author = {Barnsley, Michael F. and Elton, John H. and Hardin, Douglas P.},
	journal = {Constructive Approximation},
	number = {1},
	pages = {3--31},
	title = {Recurrent iterated function systems},
	volume = {5},
	year = {1989}}

@article{barnsley_iterated_1985,
	author = {Barnsley, M. F. and Demko, S.},
	journal = {Proceedings of the Royal Society of London. Series A, Mathematical and Physical Sciences},
	number = {1817},
	pages = {243--275},
	title = {Iterated {Function} {Systems} and the {Global} {Construction} of {Fractals}},
	volume = {399},
	year = {1985}}

@article{wu_asymptotic_2011,
	author = {Wu, Wei Biao},
	journal = {Statistics and Its Interface},
	number = {2},
	pages = {207--226},
	title = {Asymptotic theory for stationary processes},
	volume = {4},
    options = {uniquename=false},
	year = {2011}}

@article{Basrak2009regularly,
  title={Regularly varying multivariate time series},
  author={Basrak, Bojan and Segers, Johan},
  journal={Stochastic processes and their applications},
  volume={119},
  number={4},
  pages={1055--1080},
  year={2009},
  publisher={Elsevier}
}

@article{Matsui2023self,
	author = {Muneya Matsui and Thomas Mikosch and Olivier Wintenberger},
	journal = {Stochastic Processes and their Applications},
	pages = {104729},
	title = {Self-normalized partial sums of heavy-tailed time series},
	volume = {190},
	year = {2025}}

@inproceedings{lei_fine-grained_2020,
	author = {Lei, Yunwen and Ying, Yiming},
	booktitle = {Proceedings of the 37th {International} {Conference} on {Machine} {Learning}},
	pages = {5809--5819},
	publisher = {JMLR.org},
	series = {{ICML}'20},
	title = {Fine-grained analysis of stability and generalization for stochastic gradient descent},
	volume = {119},
    options = {uniquename=false},
	year = {2020}}

@inproceedings{park_generalization_2022,
	author = {Park, Sejun and Simsekli, Umut and Erdogdu, Murat A.},
	booktitle = {Advances in {Neural} {Information} {Processing} {Systems}},
	pages = {2790--2802},
	title = {Generalization {Bounds} for {Stochastic} {Gradient} {Descent} via {Localized} {$\varepsilon$}-{Covers}},
	volume = {35},
    options = {uniquename=false},
	year = {2022}}

@inproceedings{sachs_generalization_2023,
	author = {Sachs, Sarah and Erven, Tim van and Hodgkinson, Liam and Khanna, Rajiv and {\c S}im{\c s}ekli, Umut},
	booktitle = {Proceedings of {Thirty} {Sixth} {Conference} on {Learning} {Theory}},
	pages = {4863--4880},
	publisher = {PMLR},
	title = {Generalization {Guarantees} via {Algorithm}-dependent {Rademacher} {Complexity}},
    options = {uniquename=false},
	year = {2023}}

@inproceedings{vural_mirror_2022,
	author = {Vural, Nuri Mert and Yu, Lu and Balasubramanian, Krishna and Volgushev, Stanislav and Erdogdu, Murat A.},
	booktitle = {Proceedings of {Thirty} {Fifth} {Conference} on {Learning} {Theory}},
	pages = {65--102},
	publisher = {PMLR},
	title = {Mirror {Descent} {Strikes} {Again}: {Optimal} {Stochastic} {Convex} {Optimization} under {Infinite} {Noise} {Variance}},
    options = {uniquename=false},
	year = {2022}}

@inproceedings{zhang_why_2019,
	author = {Zhang, Jingzhao and He, Tianxing and Sra, Suvrit and Jadbabaie, Ali},
	title = {Why {Gradient} {Clipping} {Accelerates} {Training}: {A} {Theoretical} {Justification} for {Adaptivity}},
    booktitle = {International Conference on Learning Representations},
    options = {uniquename=false},
	year = {2019}}

@article{buraczewski_asymptotics_2012,
	author = {Buraczewski, Dariusz and Damek, Ewa and Mirek, Mariusz},
	journal = {Stochastic Processes and their Applications},
	number = {1},
	pages = {42--67},
	title = {Asymptotics of stationary solutions of multivariate stochastic recursions with heavy tailed inputs and related limit theorems},
	volume = {122},
    options = {uniquename=false},
	year = {2012}}

@inproceedings{wang_convergence_2021,
	address = {Red Hook, NY, USA},
	author = {Wang, Hongjian and G{\"u}rb{\"u}zbalaban, Mert and Zhu, Lingjiong and {\c S}im{\c s}ekli, Umut and Erdogdu, Murat A.},
	booktitle = {Proceedings of the 35th {International} {Conference} on {Neural} {Information} {Processing} {Systems}},
	pages = {18866--18877},
	publisher = {Curran Associates Inc.},
	series = {{NIPS} '21},
	title = {Convergence rates of stochastic gradient descent under infinite noise variance},
	year = {2021}}

@article{cuny_martingale_2014,
	author = {Cuny, Christophe and Merlev{\`e}de, Florence},
	journal = {The Annals of Probability},
	number = {2},
	pages = {760--793},
	title = {On {Martingale} {Approximations} and the {Quenched} {Weak} {Invariance} {Principle}},
	volume = {42},
    options = {uniquename=false},
	year = {2014}}

@article{krasulina_stochastic_1969,
	author = {Krasulina, T. P.},
	journal = {Theory of Probability \& Its Applications},
	number = {3},
	pages = {522--526},
	title = {On {Stochastic} {Approximation} {Processes} with {Infinite} {Variance}},
	volume = {14},
    options = {uniquename=false},
	year = {1969}}

@article{wu_additive_2003,
	author = {Wu, Wei Biao},
	journal = {Statistica Sinica},
	number = {4},
	pages = {1259--1267},
	title = {Additive {Functionals} of {Infinite}-{Variance} {Moving} {Averages}},
	volume = {13},
    options = {uniquename=false},
	year = {2003}}

@inproceedings{gurbuzbalaban_heavy-tail_2021,
	author = {Gurbuzbalaban, Mert and Simsekli, Umut and Zhu, Lingjiong},
	booktitle = {Proceedings of the 38th {International} {Conference} on {Machine} {Learning}},
	pages = {3964--3975},
	publisher = {PMLR},
	title = {The {Heavy}-{Tail} {Phenomenon} in {SGD}},
    options = {uniquename=false},
	year = {2021}}

@article{gupta_convergence_2021,
	author = {Gupta, Abhishek and Haskell, William B.},
	journal = {SIAM Journal on Mathematics of Data Science},
	number = {4},
	pages = {1141--1167},
	title = {Convergence of {Recursive} {Stochastic} {Algorithms} {Using} {Wasserstein} {Divergence}},
	volume = {3},
    options = {uniquename=false},
	year = {2021}}

@article{mirek_heavy_2011,
	author = {Mirek, Mariusz},
	journal = {Probability Theory and Related Fields},
	number = {3},
	pages = {705--734},
	title = {Heavy tail phenomenon and convergence to stable laws for iterated {Lipschitz} maps},
	volume = {151},
    options = {uniquename=false},
	year = {2011}}

@article{gupta_probabilistic_2024,
	author = {Gupta, Abhishek and Jain, Rahul and Glynn, Peter},
	journal = {IEEE Transactions on Automatic Control},
	number = {9},
	pages = {5947--5962},
	title = {Probabilistic {Contraction} {Analysis} of {Iterated} {Random} {Operators}},
	volume = {69},
    options = {uniquename=false},
	year = {2024}}

@article{merad_convergence_2023,
	author = {Merad, Ibrahim and Gaïffas, St{\'e}phane},
	title = {Convergence and concentration properties of constant step-size {SGD} through {Markov} chains},
    note = {arXiv:2306.11497},
	journal = {arXiv preprint},
    options = {uniquename=false},
	year = {2023}}

@article{wu_strong_2007,
	author = {Wu, Wei Biao},
	journal = {The Annals of Probability},
	number = {6},
	pages = {2294--2320},
	title = {Strong {Invariance} {Principles} for {Dependent} {Random} {Variables}},
	volume = {35},
    options = {uniquename=false},
	year = {2007}}

@article{wang_stochastic_2010,
	author = {Wang, Xin and Gao, Na},
	journal = {IEEE Transactions on Information Theory},
	number = {5},
	pages = {2382--2391},
	title = {Stochastic {Resource} {Allocation} {Over} {Fading} {Multiple} {Access} and {Broadcast} {Channels}},
	volume = {56},
    options = {uniquename=false},
	year = {2010}}

@article{gandikota_vqsgd_2022,
	author = {Gandikota, Venkata and Kane, Daniel and Maity, Raj Kumar and Mazumdar, Arya},
	journal = {IEEE Transactions on Information Theory},
	number = {7},
	pages = {4573--4587},
	title = {{vqSGD}: {Vector} {Quantized} {Stochastic} {Gradient} {Descent}},
	volume = {68},
    options = {uniquename=false},
	year = {2022}}

@article{rio_moment_2009,
	author = {Rio, Emmanuel},
	journal = {Journal of Theoretical Probability},
	number = {1},
	pages = {146--163},
	title = {Moment {Inequalities} for {Sums} of {Dependent} {Random} {Variables} under {Projective} {Conditions}},
	volume = {22},
    options = {uniquename=false},
	year = {2009}}

@article {pinelis_best_2015,
    AUTHOR = {Pinelis, Iosif},
     TITLE = {Best possible bounds of the von {B}ahr--{E}sseen type},
  JOURNAL = {Annals of Functional Analysis},
    VOLUME = {6},
      YEAR = {2015},
    NUMBER = {4},
     PAGES = {1--29},
}

@article{chen_statistical_2020,
	author = {Chen, Xi and Lee, Jason D. and Tong, Xin T. and Zhang, Yichen},
	journal = {The Annals of Statistics},
	number = {1},
	pages = {251--273},
	title = {Statistical inference for model parameters in stochastic gradient descent},
	volume = {48},
    options = {uniquename=false},
	year = {2020}}

@article{bach_adaptivity_2014,
	author = {Bach, Francis},
	journal = {Journal of Machine Learning Research},
	number = {19},
	pages = {595--627},
	title = {Adaptivity of {Averaged} {Stochastic} {Gradient} {Descent} to {Local} {Strong} {Convexity} for {Logistic} {Regression}},
	volume = {15},
    options = {uniquename=false},
	year = {2014}}

@article{zhong_online_2023,
	author = {Zhong, Yanjie and Kuffner, Todd and Lahiri, Soumendra},
	note = {arXiv:2306.02205},
	journal = {arXiv preprint},
	title = {Online {Bootstrap} {Inference} with {Nonconvex} {Stochastic} {Gradient} {Descent} {Estimator}},
    options = {uniquename=false},
	year = {2023}}

@article{su_higrad_2023,
	author = {Su, Weijie and Zhu, Yuancheng},
	title = {{HiGrad}: {Uncertainty} {Quantification} for {Online} {Learning} and {Stochastic} {Approximation}},
    journal = {Journal of Machine Learning Research},
    volume = {24},
    year = {2023},
    options = {uniquename=false},
    pages = {1--53}}

@article{liang_statistical_2019,
	author = {Liang, Tengyuan and Su, Weijie},
	journal = {Journal of the Royal Statistical Society Series B: Statistical Methodology},
	number = {2},
	pages = {431--456},
	title = {Statistical {Inference} for the {Population} {Landscape} via {Moment}-{Adjusted} {Stochastic} {Gradients}},
	volume = {81},
    options = {uniquename=false},
	year = {2019}}

@article{fang_scalable_2019,
	author = {Fang, Yixin},
	journal = {Scandinavian Journal of Statistics},
	number = {4},
	pages = {987--1002},
	title = {Scalable statistical inference for averaged implicit stochastic gradient descent},
	volume = {46},
    options = {uniquename=false},
	year = {2019}}

@article{fang_online_2018,
	author = {Fang, Yixin and Xu, Jinfeng and Yang, Lei},
	title = {Online {Bootstrap} {Confidence} {Intervals} for the {Stochastic} {Gradient} {Descent} {Estimator}},
    journal = {Journal of Machine Learning Research},
    volume = {19},
    year = {2018},
    options = {uniquename=false},
    pages = {1--21}}

@article{chao_generalization_2019,
	author = {Chao, Shih-Kang and Cheng, Guang},
    note = {arXiv:1909.10072},
	journal = {arXiv preprint},
	title = {A generalization of regularized dual averaging and its dynamics},
    options = {uniquename=false},
	year = {2019}}

@article{komlos_approximation_1976,
	author = {Koml{\'o}s, Jx and Major, P. and Tusn{\'a}dy, G.},
	journal = {Zeitschrift f{\"u}r Wahrscheinlichkeitstheorie und Verwandte Gebiete},
	number = {1},
	pages = {33--58},
	title = {An approximation of partial sums of independent {RV}'s, and the sample {DF}. {II}},
	volume = {34},
    options = {uniquename=false},
	year = {1976}}

@article{komlos_approximation_1975,
	author = {Koml{\'o}s, J. and Major, P. and Tusn{\'a}dy, G.},
	journal = {Zeitschrift f{\"u}r Wahrscheinlichkeitstheorie und Verwandte Gebiete},
	number = {1},
	pages = {111--131},
	title = {An approximation of partial sums of independent {RV}'-s, and the sample {DF}. {I}},
	volume = {32},
    options = {uniquename=false},
	year = {1975}}

@book{stoer_intro_2002,
    author = {Stoer, J and Bulirsch, R},
    title = {Introduction to {Numerical} {Analysis}},
    edtion = {3rd},
    publisher = {Springer, New York},
    options = {uniquename=false},
    year = {2002}
}

@inproceedings{moulines_non-asymptotic_2011,
	author = {Moulines, Eric and Bach, Francis},
	booktitle = {Proceedings of the 23rd {International} {Conference} on {Neural} {Information} {Processing} {Systems}},
	title = {Non-{Asymptotic} {Analysis} of {Stochastic} {Approximation} {Algorithms} for {Machine} {Learning}},
	pages = {856--864},
    options = {uniquename=false},
	year = {2011}}

@article{lou_beyond_2022,
	author = {Lou, Zhipeng and Zhu, Wanrong and Wu, Wei Biao},
	journal = {Journal of Machine Learning Research},
	number = {46},
	title = {Beyond {Sub}-{Gaussian} {Noises}: {Sharp} {Concentration} {Analysis} for {Stochastic} {Gradient} {Descent}},
	volume = {23},
    pages = {1--22},
    options = {uniquename=false},
	year = {2022}}

@inproceedings{lakshminarayanan_linear_2018,
	author = {Lakshminarayanan, Chandrashekar and Szepesvari, Csaba},
	booktitle = {Proceedings of the {Twenty}-{First} {International} {Conference} on {Artificial} {Intelligence} and {Statistics}},
	pages = {1347--1355},
	title = {Linear {Stochastic} {Approximation}: {How} {Far} {Does} {Constant} {Step}-{Size} and {Iterate} {Averaging} {Go}?},
    options = {uniquename=false},
	year = {2018}}

@article{jain_parallelizing_2018,
	author = {Jain, Prateek and Kakade, Sham M. and Kidambi, Rahul and Netrapalli, Praneeth and Sidford, Aaron},
	journal = {Journal of Machine Learning Research},
	number = {223},
	pages = {1--42},
	title = {Parallelizing {Stochastic} {Gradient} {Descent} for {Least} {Squares} {Regression}: {Mini}-batching, {Averaging}, and {Model} {Misspecification}},
	volume = {18},
    options = {uniquename=false},
	year = {2018}}

@article{gyorfi_averaged_1996,
	author = {Gy{\"o}rfi, L{\'a}szl{\'o} and Walk, Harro},
	journal = {SIAM Journal on Control and Optimization},
	number = {1},
	pages = {31--61},
	title = {On the {Averaged} {Stochastic} {Approximation} for {Linear} {Regression}},
	volume = {34},
    options = {uniquename=false},
	year = {1996}}

@article{dieuleveut_nonparametric_2016,
	author = {Dieuleveut, Aymeric and Bach, Francis},
	journal = {The Annals of Statistics},
	number = {4},
	pages = {1363--1399},
	title = {Nonparametric stochastic approximation with large step-sizes},
	volume = {44},
    options = {uniquename=false},
	year = {2016}}

@article{dieuleveut_harder_2017,
	author = {Dieuleveut, Aymeric and Flammarion, Nicolas and Bach, Francis},
	journal = {The Journal of Machine Learning Research},
	number = {1},
	pages = {3520--3570},
	title = {Harder, better, faster, stronger convergence rates for least-squares regression},
	volume = {18},
    options = {uniquename=false},
	year = {2017}}

@inproceedings{defossez_averaged_2015,
	author = {Defossez, Alexandre and Bach, Francis},
	booktitle = {Proceedings of the 18th {International} {Conference} on {Artificial} {Intelligence} and {Statistics}},
	pages = {205--213},
	title = {Averaged {Least}-{Mean}-{Squares}: {Bias}-{Variance} {Trade}-offs and {Optimal} {Sampling} {Distributions}},
    options = {uniquename=false},
	year = {2015}}

@inproceedings{vaswani_fast_2019,
	author = {Vaswani, Sharan and Bach, Francis and Schmidt, Mark},
	booktitle = {Proceedings of the {Twenty}-{Second} {International} {Conference} on {Artificial} {Intelligence} and {Statistics}},
	title = {Fast and {Faster} {Convergence} of {SGD} for {Over}-{Parameterized} {Models} and an {Accelerated} {Perceptron}},
    options = {uniquename=false},
	year = {2019}}

@article{schmidt_fast_2013,
	author = {Schmidt, Mark and Roux, Nicolas Le},
	note = {arXiv:1308.6370},
	journal = {arXiv prerpint},
	title = {Fast {Convergence} of {Stochastic} {Gradient} {Descent} under a {Strong} {Growth} {Condition}},
    options = {uniquename=false},
	year = {2013}}

@inproceedings{needell_stochastic_2014,
	author = {Needell, Deanna and Ward, Rachel and Srebro, Nati},
	booktitle = {Proceedings of the 27th {International} {Conference} on {Neural} {Information} {Processing} {Systems}},
	pages = {1017--1025},
	title = {Stochastic {Gradient} {Descent}, {Weighted} {Sampling}, and the {Randomized} {Kaczmarz} algorithm},
    options = {uniquename=false},
	year = {2014}}

@inproceedings{ma_power_2018,
	author = {Ma, Siyuan and Bassily, Raef and Belkin, Mikhail},
	booktitle = {Proceedings of the 35th {International} {Conference} on {Machine} {Learning}},
	pages = {3325--3334},
	title = {The {Power} of {Interpolation}: {Understanding} the {Effectiveness} of {SGD} in {Modern} {Over}-parametrized {Learning}},
    options = {uniquename=false},
	year = {2018}}

@inproceedings{yu_analysis_2021,
	author = {Yu, Lu and Balasubramanian, Krishna and Volgushev, Stanislav and Erdogdu, Murat A.},
	title = {An {Analysis} of {Constant} {Step} {Size} {SGD} in the {Non}-convex {Regime}: {Asymptotic} {Normality} and {Bias}},
    booktitle = {{Proceedings of the 35th {International} {Conference} on {Neural} {Information} {Processing} {Systems}},},
    pages = {4234--4248},
    options = {uniquename=false},
	year = {2021}}

@article{pflug_stochastic_1986,
	author = {Pflug, Georg Ch.},
	journal = {SIAM Journal on Control and Optimization},
	number = {4},
	pages = {655--666},
	title = {Stochastic {Minimization} with {Constant} {Step}-{Size}: {Asymptotic} {Laws}},
	volume = {24},
    options = {uniquename=false},
	year = {1986}}

@inproceedings{bach_non-strongly-convex_2013,
	author = {Bach, Francis and Moulines, Eric},
	booktitle = {Proceedings of the 26th {International} {Conference} on {Neural} {Information} {Processing} {Systems}},
	title = {Non-strongly-convex smooth stochastic approximation with convergence rate {O}(1/n)},
    pages = {773--781},
    options = {uniquename=false},
	year = {2013}}

@article{sacks_asymptotic_1958,
	author = {Sacks, Jerome},
	journal = {The Annals of Mathematical Statistics},
	number = {2},
	pages = {373--405},
	title = {Asymptotic {Distribution} of {Stochastic} {Approximation} {Procedures}},
	volume = {29},
    options = {uniquename=false},
	year = {1958}}

@incollection{robbins_convergence_1971,
	author = {Robbins, H. and Siegmund, D.},
	booktitle = {Optimizing {Methods} in {Statistics}},
	pages = {233--257},
	publisher = {Academic Press},
	title = {A {Convergence} {Theorem} for {Non} {Negative} {Almost} {Supermartingales} and {Some} {Applications}},
    options = {uniquename=false},
	year = {1971}}

@article{ljung_analysis_1977,
	author = {Ljung, L.},
	journal = {IEEE Transactions on Automatic Control},
	number = {4},
	pages = {551--575},
	title = {Analysis of recursive stochastic algorithms},
	volume = {22},
    options = {uniquename=false},
	year = {1977}}

@article{lai_stochastic_2003,
	author = {Lai, Tze Leung},
	journal = {The Annals of Statistics},
	number = {2},
	pages = {391--406},
	title = {Stochastic approximation: invited paper},
	volume = {31},
    options = {uniquename=false},
	year = {2003}}

@article{fabian_asymptotic_1968,
	author = {Fabian, Vaclav},
	journal = {The Annals of Mathematical Statistics},
	number = {4},
	pages = {1327--1332},
	title = {On {Asymptotic} {Normality} in {Stochastic} {Approximation}},
	volume = {39},
    options = {uniquename=false},
	year = {1968}}

@article{dvoretzky_stochastic_1956,
	author = {Dvoretzky, Aryeh},
	pages = {39--56},
	journal = {Proceedings of the Third Berkeley Symposium on Mathematical Statistics and Probability},
	title = {On {Stochastic} {Approximation}},
	volume = {3},
    number = {1},
    options = {uniquename=false},
	year = {1956}}

@article{blum_approximation_1954,
	author = {Blum, Julius R.},
	journal = {The Annals of Mathematical Statistics},
	number = {2},
	pages = {382--386},
	title = {Approximation {Methods} which {Converge} with {Probability} one},
	volume = {25},
    options = {uniquename=false},
	year = {1954}}

@incollection{ruppert_efficient_1988,
	author = {Ruppert, David},
	title = {Efficient {Estimations} from a {Slowly} {Convergent} {Robbins}-{Monro} {Process}},
    booktitle = {Technical report},
    number = {781},
    publisher = {Cornell University Operations Research and Industrial Engineering},
    options = {uniquename=false},
	year = {1988}}

@article{polyak_acceleration_1992,
	author = {Polyak, B. T. and Juditsky, A. B.},
	journal = {SIAM Journal on Control and Optimization},
	number = {4},
	pages = {838--855},
	title = {Acceleration of {Stochastic} {Approximation} by {Averaging}},
	volume = {30},
    options = {uniquename=false},
	year = {1992}}

@article{karmakar_optimal_2020,
	author = {Karmakar, Sayar and Wu, Wei Biao},
	journal = {Statistica Sinica},
	number = {3},
	pages = {1399--1417},
	title = {Optimal {Gaussian} {Approximation} for {Multiple} {Time} {Series}},
	volume = {30},
    options = {uniquename=false},
	year = {2020}}

@article{diaconis_freedman_1999,
	author = {Diaconis, Persi and Freedman, David},
	journal = {SIAM Review},
	number = {1},
	pages = {45--76},
	volume = {41},
	title = {Iterated {Random} {Functions}},
    options = {uniquename=false},
	year = {1999}}

@article{dahlhaus_bernoulli_2019,
	author = {Dahlhaus, Rainer and Richter, Stefan and Wu, Wei Biao},
	journal = {Bernoulli},
	number = {2},
	title = {Towards a general theory for nonlinear locally stationary processes},
	volume = {25},
	year = {2019},
    options = {uniquename=false},
    pages = {1013--1044}}

@article{dahlhaus_rao_aos_2006,
	author = {Dahlhaus, Rainer and Rao, Suhasini Subba},
	journal = {The Annals of Statistics},
	number = {3},
	pages = {1075--1114},
	title = {Statistical inference for time-varying {ARCH} processes},
	volume = {34},
    options = {uniquename=false},
	year = {2006}}

@article{Huo2023_linear_stochastic_approx,
    author = {Huo, Dongyan and Chen, Yudong and Xie, Qiaomin},
    title = {Bias and Extrapolation in Markovian Linear Stochastic Approximation with Constant Stepsizes},
    year = {2023},
    issue_date = {June 2023},
    volume = {51},
    number = {1},
    journal = {Proceedings of the 2023 ACM SIGMETRICS International Conference on Measurement and Modeling of Computer Systems},
    pages = {81–-82},
    options = {uniquename=false}
}

@article {Hardle1986,
    AUTHOR = {H\"{a}rdle, Wolfgang},
     TITLE = {A note on jackknifing kernel regression function estimators},
    JOURNAL = {IEEE Transactions on Information Theory},
    VOLUME = {32},
      YEAR = {1986},
    NUMBER = {2},
     PAGES = {298--300},
    options = {uniquename=false}
}

@article{propp1996exact,
  title={Exact sampling with coupled Markov chains and applications to statistical mechanics},
  author={Propp, James Gary and Wilson, David Bruce},
  journal={Random Structures \& Algorithms},
  volume={9},
  number={1-2},
  pages={223--252},
  year={1996},
  publisher={Wiley Online Library}
}

@article{Zhu2023,
  title={Online covariance matrix estimation in stochastic gradient descent},
  author={Zhu, Wanrong and Chen, Xi and Wu, Wei Biao},
  journal={Journal of the American Statistical Association},
  year={2023},
  volume={118},
  number={541},
  pages={393--404},
  publisher={Taylor \& Francis},
  options = {uniquename=false}
}

@article {Wu2009,
    AUTHOR = {Wu, Wei Biao},
     TITLE = {Recursive estimation of time-average variance constants},
    JOURNAL = {The Annals of Applied Probability},
    VOLUME = {19},
      YEAR = {2009},
    NUMBER = {4},
    options = {uniquename=false},
    PAGES = {1529--1552} 
}

@article {Xiao2011,
    AUTHOR = {Xiao, Han and Wu, Wei Biao},
     TITLE = {A single-pass algorithm for spectrum estimation with fast
              convergence},
    JOURNAL = {IEEE Transactions on Information Theory},
    VOLUME = {57},
      YEAR = {2011},
    NUMBER = {7},
     PAGES = {4720--4731},
    options = {uniquename=false}
}

@article{berkes_liu_wu_2014,
	author = {Berkes, Istv{\'a}n and Liu, Weidong and Wu, Wei Biao},
	journal = {The Annals of Probability},
	number = {2},
	pages = {794--817},
	title = {Koml{\'o}s--{Major}--{Tusn{\'a}dy} approximation under dependence},
	volume = {42},
    options = {uniquename=false},
	year = {2014}}

@article{wu_nonlinear_2005,
	author = {Wu, W B},
	journal = {Proceedings of the National Academy of Sciences},
	number = {40},
	pages = {14150--14154},
	pmid = {16179388},
	shorttitle = {Nonlinear system theory},
	title = {Nonlinear system theory: {Another} look at dependence},
	volume = {102},
    options = {uniquename=false},
	year = {2005}}

@article {wu_limit_2004,
    AUTHOR = {Wu, W B and Shao, Xiaofeng},
     TITLE = {Limit theorems for iterated random functions},
  JOURNAL = {Journal of Applied Probability},
    VOLUME = {41},
      YEAR = {2004},
    NUMBER = {2},
     PAGES = {425--436},
    options = {uniquename=false}
}

@article{robbins_stochastic_1951,
	author = {Robbins, Herbert and Monro, Sutton},
	journal = {The Annals of Mathematical Statistics},
	number = {3},
	pages = {400--407},
	title = {A {Stochastic} {Approximation} {Method}},
	volume = {22},
    options = {uniquename=false},
	year = {1951}}

@article{kiefer_stochastic_1952,
	author = {Kiefer, J. and Wolfowitz, J.},
	journal = {The Annals of Mathematical Statistics},
	number = {3},
	pages = {462--466},
	title = {Stochastic {Estimation} of the {Maximum} of a {Regression} {Function}},
	volume = {23},
    options = {uniquename=false},
	year = {1952}}

@article{bach_bridging_2020,
	author = {Dieuleveut, Aymeric and Durmus, Alain and Bach, Francis},
	journal = {The Annals of Statistics},
	number = {3},
	pages = {1348--1382},
	title = {Bridging the gap between constant step size stochastic gradient descent and {Markov} chains},
	volume = {48},
    options = {uniquename=false},
	year = {2020}}

\end{appendix}

\end{document}